\RequirePackage[l2tabu, orthodox]{nag}

\documentclass[12pt,phd,a4paper,oneside]{ucl_thesis}

\usepackage{blindtext}

\usepackage{emptypage}

\usepackage{graphicx}

\usepackage{float}

\usepackage{amsmath}

\usepackage{gensymb}
\usepackage{textcomp}

\usepackage{setspace}
\usepackage{multirow}

\usepackage{bibentry} 

\usepackage[format=hang,font=small,labelfont=bf]{caption}

\usepackage{etoolbox}

\usepackage{bibentry}
\makeatletter\let\saved@bibitem\@bibitem\makeatother
\usepackage[hidelinks]{hyperref}
\makeatletter\let\@bibitem\saved@bibitem\makeatother
\makeatletter
\AtBeginDocument{
    \hypersetup{
        pdfsubject={Thesis Subject},
        pdfkeywords={Thesis Keywords},
        pdfauthor={Author},
        pdftitle={Title},
    }
}
\makeatother

\usepackage{bm}
\usepackage{mathrsfs}

\usepackage{graphicx}
\usepackage{amsmath}
\usepackage{amsthm}
\usepackage{booktabs}
\usepackage{algorithmic}

\usepackage{mathrsfs}
\usepackage{multirow}
\usepackage[ruled,vlined,linesnumbered]{algorithm2e}
\usepackage{colortbl}
\usepackage{booktabs}

\usepackage{bm}
\usepackage{amssymb}
\usepackage{color}
\usepackage{subfigure}
\usepackage{mathrsfs}
\usepackage{multirow}
\usepackage{natbib}
\usepackage{fancyhdr}
\usepackage[utf8]{inputenc}

\newcommand{\bA}        {{\mathbf{A}}}
\newcommand{\ba}        {{\mathbf{a}}}

\newcommand{\bc}        {{\mathbf{c}}}

\newcommand{\bE}        {{\mathbf{E}}}
\newcommand{\be}        {{\mathbf{e}}}

\newcommand{\bff}        {{\mathbf{f}}}
\newcommand{\bG}        {{\mathbf{G}}}
\newcommand{\bg}        {{\mathbf{g}}}

\newcommand{\bh}        {{\mathbf{h}}}

\newcommand{\bi}        {{\mathbf{i}}}

\newcommand{\bj}        {{\mathbf{j}}}

\newcommand{\bo}        {{\mathbf{o}}}

\newcommand{\bq}        {{\mathbf{q}}}

\newcommand{\br}        {{\mathbf{r}}}

\newcommand{\bW}        {{\mathbf{W}}}

\newcommand{\bX}        {{\mathbf{X}}}
\newcommand{\bx}        {{\mathbf{x}}}

\newcommand{\bz}        {{\mathbf{z}}}

\newcommand{\kETAL}    {{\em et al. }}

\begin{document}

\nobibliography*



\title{Blood Glucose Level Prediction: A Graph-based Explainable Method with Federated Learning}
\author{Chengzhe Piao, Ken Li}
\department{Institute of Health Informatics}

\maketitle
\makedeclaration

\begin{abstract} 
Around 400 thousand people in the UK have type 1 diabetes (T1D). 
They cannot maintain their normal blood glucose (BG) levels without the help of insulin delivery because their pancreas cannot produce enough insulin. 
Some appropriate precautions, e.g., insulin delivery and carbohydrate intake, can help them avoid complications caused by unstable BG levels.
Blood glucose level prediction (BGLP) based on the BG time series collected by continuous glucose monitoring (CGM) enables them to take suitable precautions, where CGM tracks the BG levels every 5 minutes.

Numerous researchers focus on modeling complex and dynamic temporal features by proposing novel sequential models based on historical BG time series, and they also consider extra attributes, e.g., carbohydrate intake, insulin delivery, and time, as the input.
These methods achieved excellent performance in BGLP; some are explainable from a temporal perspective.
However, they cannot provide clinicians with clear correlations among the considered attributes, presenting how much importance each attribute has in BGLP. 
Besides, some of these methods try to learn population patterns by merging the training data of all participants, which is an invasion of privacy.

Based on the limitations, we proposed a graph attentive memory (GAM), consisting of a graph attention network (GAT) and a gated recurrent unit (GRU). 
GAT utilizes graph attention to model the correlations of multiple attributes, and the dynamic graph structure can present transparent relationships among these attributes. The attention weights can be used to dynamically measure the importance of the attributes at different time slots.
Moreover, federated learning (FL) is introduced to learn population patterns safely, protecting participants' privacy.

Our proposed method is evaluated by OhioT1DM'18 \citep{DBLP:conf/ijcai/MarlingB18} and OhioT1DM'20 \citep{DBLP:conf/ecai/MarlingB20}, containing 12 participants in total, where the data includes multiple attributes.
We designed experiments to select 6 valuable attributes, and we proved the stability of our model by exploring the impact of hyperparameters.
Finally, our model got the root mean squared error (RMSE) as 18.753 mg/dL or 31.275 mg/dL when the prediction horizon (PH) is 30 or 60 minutes, and it got the mean absolute relative difference (MARD) as 8.893\% or 15.903\% when PH is 30 or 60 minutes.
Compared with the methods of the BGLP 2020 Challenge, our model achieved third place when training without FL.
We also use two examples to visualize our GAM, presenting the dynamic correlations among the active attributes in different time slots.

\end{abstract}


\setcounter{tocdepth}{2} 

\tableofcontents

\chapter{Introduction}
\label{chap:intro}

\section{Background}
\label{sec:bg}

T1D\footnote{https://www.nhs.uk/conditions/type-1-diabetes/}, whose ICD-10 is E10, happens when the pancreas cannot produce enough insulin, leading to high BG levels ($> 180$ mg/dL, hyperglycemia).
People with T1D need insulin to maintain their everyday life by insulin pumps and bolus stabilizing their BG levels in a normal range (70-180 mg/dL).
Otherwise, they are highly likely to get hypoglycemia ($< 70$ mg/dL) or hyperglycemia ($> 180$ mg/dL), contributing to some serious complications\footnote{https://www.nhs.uk/conditions/type-1-diabetes/living-with-type-1-diabetes/avoiding-complications/}, e.g., kidney problems, sight problems, blindness, and so forth.
However, there are many precautions to avoid these complications, and one effective prevention is to monitor and manage BG levels actively.

Currently, people with T1D often use CGM to monitor their BG levels, where CGM can track the variations of BG levels per 5 minutes.
Based on the nearly continuous BG time series, a lot of researchers successfully proposed all kinds of methods to do the BGLP for people with T1D.
Even though these approaches achieve excellent performance, they still have some non-ignorable limitations (see section \ref{sec:related}).
This paper is aiming at solving these problems by proposing novel methods and algorithms (see section \ref{sec:obj}), where the proposed methods are evaluated in OhioT1DM'18 \citep{DBLP:conf/ijcai/MarlingB18} and OhioT1DM'20 \citep{DBLP:conf/ecai/MarlingB20}, and the contributions are briefly mentioned in section \ref{sec:contribution}.

\section{Related work}
\label{sec:related}
\subsection{Methods for Blood Glucose Level Prediction (BGLP)}
\label{sec:bglp1}
In BGLP, the historical data can be organized as multivariate time series (MTS) when considering multiple attributes. 
Since 2020, methods based on neural networks (NNs) have achieved outstanding performance, exceeding non-NN approaches, especially in the BGLP 2020 Challenge \citep{DBLP:conf/ecai/2020kdh}.

\cite{RubinFalconeFW20} introduced N-BEATS \citep{DBLP:conf/iclr/OreshkinCCB20} to model MTS, where N-BEATS is the state-of-the-art method dealing with time series around 2020, and the proposed method based on N-BEATS achieved first place in the BGLP 2020 Challenge. N-BEATS utilizes residual structure and forecast and backcast mechanisms to make the sequential model very deep, modeling complex temporal dependencies.
Besides, it has trend stacks and seasonal stacks, making the whole structure temporally explainable to some degree.
However, it is far from enough because this approach cannot explain the importance of attributes, i.e., feature importance, during the prediction.
For example, Rubin-Falcone \kETAL considered glucose levels, bolus, finger sticks, etc., as the input of their NNs. However, they cannot explain each attribute's importance during the prediction. 
Nevertheless, many clinicians care more about how a model makes predictions rather than only getting an estimation from a black box.

Similarly, recent approaches in BGLP primarily focus on modeling temporal dependencies based on RNN-like structures, e.g., long short-term memory (LSTM, \cite{DBLP:conf/ecai/YangWTWMZYL20}), BiLSTM \citep{650093} and gated recurrent unit (GRU, \cite{DBLP:conf/emnlp/ChoMGBBSB14}) rather than explicitly modeling the correlations of attributes in an explainable way.
Hence, we first discuss how these works focus on modeling temporal features.
%
\cite{DBLP:conf/med/IaconoMT22} utilized the original LSTM to model the historical CGM data. There are fewer innovations in the methods in their work.
Comparably, some researchers try to introduce other algorithms to make the recurrent model more powerful in learning complex temporal correlations.
%
\cite{DBLP:conf/ecai/BhimireddySOGP20} proposed a BiLSTM-based model trained with a sequence-to-sequence framework \citep{DBLP:conf/nips/SutskeverVL14}, because they believe this framework can predict continuous future time series of CGM data, enabling users to have a complete changing curve of the future BG levels.
\cite{9691857} proposed an ensemble deep learning model, where it uses some deep sequential models, e.g., LSTM, BiLSTM, etc., as weak learners, and the output of the weak learner is collected by a meta-learner which generates final estimations by fusing the collected predictions.

The above methods achieved satisfying performance in BGLP, but they are not explainable.
Meanwhile, many researchers focus on on revising the recurrent structure by adding attention mechanisms to make the model explainable from a temporal view.
\cite{DBLP:conf/ecai/BevanC20} proposed a model based on temporal attention and LSTM, only using the historical BG levels collected by CGM as the input. It achieved good performance in OhioT1DM.
\cite{9681840, 9813400} proposed an attention-based sequential model which leverages GRU layers to model MTS, including historical BG levels, meal intake, insulin delivery, etc., followed by temporal attention to integrate the output of GRU layers.
%
Instead of using recurrent structure, \cite{9474665} introduced self-attention \citep{DBLP:conf/nips/VaswaniSPUJGKP17} to model temporal dependencies and achieved excellent performance in OhioT1DM. One of the advantages of using self-attention is that the temporal dependencies are explainable.
However, the computation resources are higher than RNN-based structures, which need more time to converge during the training process.

Nevertheless, even though the above methods contribute to modeling temporal dependencies, they ignore explicitly modeling correlations of various attributes.
Some researchers tend to add special modules to extract features from different attributes. 
\cite{DBLP:conf/ecai/YangWTWMZYL20} proposed a multi-scale and multi-lag LSTM to model different attributes explicitly. The multi-scale LSTM models the CGM data with different sampling granularity, and each attribute has a special LSTM to model its temporal variations.
\cite{DBLP:journals/titb/LiDLHG20} proposed a convolution recurrent neural network (CRNN), where CNN layers are used to extract features from MTS, and a modified LSTM is utilized to model temporal correlations, where the LSTM is revised in a residual way, adding the input to the hidden state.
\cite{DBLP:conf/ecai/ZhuYLHG20} proposed a particular generative adversarial network (GAN), using a gated recurrent unit (GRU, \cite{DBLP:conf/emnlp/ChoMGBBSB14}) as a generator and convolution neural networks (CNN) as a discriminator, where the generator generates the future estimations of BG levels. The discriminator distinguishes the actual data from the predicted data, and the whole structure is trained in an adversarial algorithm \citep{DBLP:conf/nips/GoodfellowPMXWOCB14}, making it possible to utilize complex features from different attributes.

Then, also as for the feature extraction, \cite{DBLP:journals/asc/YangYMWL22} proposed a channel deep learning framework based on LSTM. They hold the view that different attributes have various time dependencies. For example, BG levels change for several hours after a meal, while BG levels might be affected for the whole day after receiving some long-term insulin. Hence, they leverage special channel-wise gates to control the feature extraction from different attributes, where only conducive features are leveraged. They use LSTM to model temporal dependencies of the MTS.
\cite{DBLP:journals/midm/RabbyTHLMH21} proposed a stacked LSTM aided by a Kalman smoothing approach leveraged to deal with the inaccuracy of CGM data. They believe the stacked LSTM can further improve prediction accuracy. Furthermore, CGM data is processed by a Kalman smoother filter and other attributes, i.e., meal, insulin, and so forth, are processed by a feature extraction module. The output of these two modules is fused and acts as the input of the stacked LSTM.

Even if the above methods can model complex temporal dependencies and can also model correlations of attributes, the correlations are not explainable. 
The NNs can achieve excellent performance in BGLP, but they are still black boxes.

\subsection{Multi-Variate Time Series and Graph Neural Networks}
\label{sec:mvts}

Recently, there have been some general models for MTS, but they have not been used in BGLP.
\cite{DBLP:conf/icml/SchirmerELR22} proposed a continuous recurrent unit (CRU), given that the observation times of samples are irregular in MTS in some scenarios. The model is able to deal directly with MTS via a continuous-discrete Kalman filter. The model interpolates MTS according to neural ordinary differential equations, achieving better performance than other methods.
Given that different time series in MTS have various temporal granularity, \cite{DBLP:conf/www/Hou0LLLC022} proposed a multi-granularity residual learning framework (MRLF), where the multi-granularity can extract features from different temporal perspectives, and residual learning framework make the NNs deeper. Hence, this structure can extract more complex temporal features.

Similar to the methods in the previous subsection, even though these two methods can extract complex temporal features from MTS, they do not model the correlations among the attributes in MTS.
Some researchers leverage graph neural networks (GNNs) or graph convolutional networks (GCNs) to model the relationships among the attributes in MTS because graph-based methods can clearly describe the relationships.
In order to explicitly and flexibly model the correlations among the attributes in MTS in an explainable manner, we need graph-based methods.

Recently, GNNs are very popular in all kinds of applications, and GNNs are the combination of graph and neural networks.
In MTS, specific attributes may collect valid samples at each time slot. 
After embedding these samples separately, the generated tensors can be treated as graph node tensors, where each tensor relates to a node in a graph.
Then, we assume that each active node connects to all nodes, including itself, and we want to introduce a GNN method to model the correlations of the graph node tensors dynamically.
This is because graphs can clearly represent the correlations among these attributes.

A classic GNN is proposed by \cite{DBLP:conf/iclr/KipfW17}, where the concept of GCN is similar to CNN.
GCN is the convolution for graph node tensors, and the convolution is along the edges, where the weights of the edges coming to a node can be seen as a filter.
After a GCN layer, each node tensor will be replaced by the weighted summation of all the adjacent node tensors.
This process can be seen as the neural message passing and aggregation, and the adjacent node tensors are neural messages in GCN.
In this case, nodes' messages are merged based on the graph's structure, i.e., edges and weights of the edges. 
The size of a weight of a coming edge decides how crucial the adjacent node tensor is. The higher the weight, the more critical the adjacent node tensor.
Hence, the size of the weights can be used to calculate the ranking of attribute importance.
Then, simplifying graph convolutional networks (SGC) is proposed by \cite{DBLP:conf/icml/WuSZFYW19}.
SGC is a simplified version of GCN so as to reduce complexity and computation by removing unnecessary activation functions.
Besides, as for modeling complicated graph relationships, \cite{DBLP:conf/aaai/0001RFHLRG19} proposed a k-dimensional GNN (k-GNN) containing higher-order graph structures at various scales.

However, these GNNs are based on static graphs, where the structure, including the edges and the weights of the edges, are unchangeable.
Such graphs are not suitable in our scenario, as they cause unnecessary computations when most attributes do not have valid samples at specific time slots.

In terms of GNNs based on dynamic graphs, where the weights of the edges are learnable, \cite{DBLP:conf/iclr/VelickovicCCRLB18} proposed graph attention networks (GAT), where each weight of an edge is based on the node tensors that are adjacent, and the weights are learnable.
Then, \cite{DBLP:conf/iclr/Brody0Y22} proposed a GATv2 which is more expressive using a more dynamic attention mechanism.

In this paper, our proposed model is based on GAT.
This is because we can build dynamic graphs for active nodes at each time slot with the help of changing and learnable weights of edges.
Note that we do not use GATv2, because we did not get more benefits when using it.

On the other hand, many researchers also proposed GNN-based sequential models.
\cite{DBLP:conf/icml/ChenTCDDZ22} proposed a variational graph convectional recurrent network (VGCRN) for MTS, consisting of GCNs and LSTM. Then, VGCRN is extended to a deep variational network, and it is more potent in modeling complex correlations of the attributes in MTS, especially when facing noisy MTS. 
However, GCNs is based on static graphs, which is unsuitable for this paper, as in specific time slots, only part of the nodes in the graph are activated (the values of attributes are not none).
Using static graphs in this scenario is a waste of computation resources. 

When modeling complex correlations among the attributes in MTS, multi-graph-based NNs are also considered. \cite{DBLP:conf/ijcai/ShaoJWKXMZZS22} proposed a multi-graph neural network (MGNN) so as to extract complex graph features from long-term MTS. They built multiple graph models to respectively learn the contextual correlations of the attributes in MTS, followed by a dynamic multi-graph fusion module to aggregate the extracted features.
The actual number of conducive attributes in this scenario is limited, and there are not very complicated correlations among the attributes.
This view has also been confirmed in our experiments, so we do not need such complex multi-graph-based structures.

\cite{DBLP:conf/iclr/ZhangZTZ22} proposed a graph-based sequential NN, consisting of dynamic GNN layers and self-attention layers, where the structure and the weights of the graph are learnable, and the self-attention models the temporal dependencies. This is a precisely perfect model that we really need, as the graph is explainable, and the weights of edges in the graphs are also changeable, and the temporal correlations are also explainable by self-attention. However, the official code\footnote{https://github.com/mims-harvard/Raindrop/blob/main/code/Ob\_propagation.py}, where the link of the code is also given in their paper, is not consistent with the methods proposed in their paper. There is a super massive gap between their imagined model and their actually implemented model. Furthermore, we carefully debug their published code and test the actual performance, confirming that their imagination is totally different from their implementation.

Therefore, the latest methods for MTS in top conferences cannot satisfy our expectations, i.e., dynamically building explainable correlations for the attributes.
This is why we propose novel methods instead of introducing popular MTS models.
\subsection{Training and Deployment of the BGLP Methods}
\label{sec:bglp2}

In terms of the training and deployment of these NNs,  \cite{DBLP:conf/ecai/DanielsHG20, 9497711} introduced multi-task learning to train personalized models, where it does not need to mix the personalized data to pretrain a population model, followed by fine-tuning in personalized data separately. This is because multi-task learning enables concurrently leveraging population patterns and keeping personal specialties, generating multiple personalized models at a time without too much fine-tuning.
\cite{9401083} applied a well-trained LSTM-based model in a microcontroller unit to predict future BG levels for people with T1D.
Then, \cite{9681840} successfully utilized an Internet of Medical Things (IoMT) enabled wearable device to implement a temporal attention-based GRU model, where the device contains a low-cost chip to run the trained model, and desktops and cloud platform are leveraged to backup data and fine-tuning models.
\cite{d2022prediction} also successfully leveraged the API tools of TensorFlow and Kerase to implement their well-trained LSTM-based model on edge, sacrificing some prediction accuracy, which is tolerable.

However, the above methods neglect to protect the participants' privacy when mixing all the data to learn population patterns, which restricts the number of participants, and the amount of available training data is also limited.

\subsection{Federated Learning (FL)}
\label{sec:re_fl}
Nowadays, most people have smartphones, and they regard these smart devices as an indispensable part of their lives.
With the rapidly developing of technology, smart devices have more and more powerful computing hardware, supporting more sophisticated computations.

Then, FL appears, and the target of FL is to fully leverage the computing resources of these intelligent devices to do the training tasks of artificial intelligence, using the data generated by the smart devices, and keeping the data safely in the place where the data is generated.
Meanwhile, the learnable parameters from personal devices are sent to a central server periodically, and a population model is generated by averaging the learnable parameters from these personal devices.
When the population model is updated, smart devices update the local learnable parameters by copying the learnable parameters of the population model in the server.
This is the classic FegAvg algorithm proposed by \cite{DBLP:conf/aistats/McMahanMRHA17}.

Recently, FL has been developing towards being decentralized, personalized and asynchronous.
\cite{jiang2020decentralised} proposed a decentralized FL, fully leveraging node-to-node bandwidth and removing the center server.
\cite{t2020personalized} proposed a personalized FL called ``pFedMe''. It utilizes Moreau envelopes as regularized loss in smart devices, decoupling the local models of smart devices from the population model in the server.
\cite{DBLP:conf/icml/ShamsianNFC21} proposed a personalized federated hypernetworks, called ``pFedHN''. These hypernetworks make the learnable parameters be shared effectively across smart devices.
\cite{DBLP:conf/icml/MarfoqNVK22} also proposed a personalized FL based on local memorization, fully utilizing local images and text to generate compact representation tensors.
\cite{DBLP:conf/icml/YangZK022} proposed an anarchic federated learning (AFL). It enables the smart devices flexibly join or leave the training, and the smart devices asynchronously communicate with the server.

However, in this work, we only use the classic FL, i.e., FedAvg, and the advanced FL will be considered in our future work.

\section{Objectives}
\label{sec:obj}
Based on the limitations of the recent work in sections \ref{sec:bglp1}-\ref{sec:mvts}, we have the following targets in this paper.

Recent Models generally focus on modeling complex temporal dependencies and extracting features from different attributes, e.g., BG time series collected by CGM, carbohydrate and insulin intake.
Some of these methods explain temporal dependencies by leveraging temporal attention or self-attention mechanisms.
However, they do not explain how important each attribute is in BGLP. 
Hence, we aim to leverage the GAT (see section \ref{sec:mvts}) to propose a novel sequential model in order to dynamically present clear explanations for the correlations and the importance of these attributes.

Furthermore, recent work does not consider privacy protection when training population models.
Given that FL can well protect the participants' privacy when learning population patterns, we try to introduce FL (see section \ref{sec:re_fl}) to train the proposed model, in order to figure out whether FL can be leveraged in BGLP.

\section{Contributions}
\label{sec:contribution}

First of all, we proposed a graph attentive memory, called ``GAM'', based on GAT and GRU. 
The model is an explainable sequential NN for dealing with MTS, clearly presenting the correlations of the attributes by graph attention. 
In GAM, the graph structure, i.e., edges and weights of the edges, is dynamically changing based on which attributes are activated (the collected values are not none), where each attribute is a node of the graph, and the node can generate and send neural messages to the adjacent nodes.
The graph features extracted by GAT are aggregated by GRU sequentially, generating a compact tensor for predicting BG levels.
Besides, we introduced FL to train our proposed model (see chapter \ref{chap:method}).

Secondly, we designed four experiments in OhioT1DM'18 and OhioT1DM'20. The first experiment proved the conducive attributes are ``glucose\_level'', ``meal'', ``bolus'', ``finger\_stick'', ``sleep'' and ``exercise''. The second experiment showed the impact of different hyperparameters of our proposed model, presenting the stability in different settings. The third experiment showed the influence of FL, confirming that FL decreased some prediction accuracy when being used with our model. The final experiment proved that our model cannot be used with a time-aware attention because it cannot further improve the performance (see chapter \ref{chap:exp_res}).

Finally, we compared our proposed model (training without FL) with the top 8 methods in BGLP 2020 Challenge, showing a satisfying performance of our model. 
Our model got the root mean squared error (RMSE) as 18.753 mg/dL or 31.275 mg/dL when the prediction horizon (PH) is 30 or 60 minutes, and it got the mean absolute relative difference (MARD) as 8.893\% or 15.903\% when PH is 30 or 60 minutes.
We visualized our proposed model in order to present the explainable graph-based correlations of attributes in BGLP (see chapter \ref{chap:dis}).  
\chapter{Method}
\label{chap:method}

In this chapter, the problem is defined in section \ref{sec:problem_def}, and the setting and description of the dataset are in section \ref{sec:setting_dataset}.
Then, ethical concerns are mentioned in section \ref{sec:ethics}.
Next, data preprocessing and data analysis are respectively in sections \ref{sec:data_preprocessing} and \ref{sec:data_analysis}.
Finally, the proposed model is depicted in section \ref{sec:proposed_method} in great detail, starting with an overview and followed by details of each module.

\section{Problem Definition}
\label{sec:problem_def}
\textbf{Blood glucose level prediction based on regular multivariate time series (BGLP-RMTS)}: given $N$ attributes from historical $T$ time slots, denoted as $\bX=[\bx_1 ...\ \bx_t ...\ \bx_T ]\in \mathbb{R}^{N\times T}$, where $\bx_t\in\mathbb{R}^N$ contains $N$ attributes, predict the BG level $y_{T+W}$ in $30$ minutes ($W=6$) or in $60$ minutes ($W=12$). As for $\mathcal{T}\triangleq\{1, ..., t, ..., T\}$, the adjacent pair equals 5 minutes, which is the same as the sampling frequency of CGM since the BG levels are recorded every 5 minutes by CGM.

Usually, the attributes are historical BG levels, carbohydrate intake, insulin delivery, etc., and these attributes are self-reported or collected by all kinds of sensors.
Initially, these attributes are organized as irregular multivariate time series (IMTS).
Specifically, the values of each attribute may be collected or reported with various frequencies, and even the adjacent values of the same attributes may have different time intervals.
Besides, the values may be missing for many reasons, e.g., forgetting to report carbohydrate intake after a meal. 
Therefore, IMTS is padded with zero in this work.
This is because the current sequential neural network, such as LSTM, multi-head attention, etc., are good at dealing with RMTS rather than IMTS.

\section{Setting and Dataset}
\label{sec:setting_dataset}

OhioT1DM'18 \citep{DBLP:conf/ijcai/MarlingB18} and OhioT1DM'20 \citep{DBLP:conf/ecai/MarlingB20} are the two datasets considered in this work.
They consist of 12 participants (6 of each), and all these 12 participants have T1D.
The statistics of these two datasets can be seen in Table \ref{tab:statistics}.
We have the following observations. 
\begin{itemize}
    \item There are 5 female and 7 male participants, meaning that the data is nearly balanced in gender.
    \item Most participants are middle-aged, as 8 participants belong to 40-60 in terms of age.
    \item For most participants, the proportion between the number of training and testing examples is around 5:1.
\end{itemize}
\begin{table}[tb]

	\caption{Basic information of the OhioT1DM'18 and OhioT1DM'20.}
	\label{tab:statistics}
	\centering
	\resizebox{0.95\textwidth}{!}{
		\begin{tabular}{@{}cccccc@{}}
			\toprule
			Patient ID (PID) & BGLP challenge & Gender & Age   & \#Training\_examples & \#Test\_examples \\ \midrule
			559        & 2018           & female & 40–60 & 10796             & 2514          \\
			563        & 2018           & male   & 40–60 & 12124             & 2570          \\
			570        & 2018           & male   & 40–60 & 10982             & 2745          \\
			575        & 2018           & female & 40–60 & 11866             & 2590           \\
			588        & 2018           & female & 40–60 & 12640             & 2791          \\
			591        & 2018           & female & 40–60 & 10847             & 2760          \\
			540        & 2020           & male   & 20–40 & 11947             & 2884          \\
			544        & 2020           & male   & 40–60 & 10623             & 2704          \\
			552        & 2020           & male   & 20–40 & 9080              & 2352          \\
			567        & 2020           & female & 20–40 & 10858             & 2377          \\
			584        & 2020           & male   & 40–60 & 12150             & 2653          \\
			596        & 2020           & male   & 60–80 & 10877             & 2731          \\ \bottomrule
	\end{tabular}}
	\vspace{-0.1cm}
\end{table}

\section{Ethics}
\label{sec:ethics}
According to \cite{DBLP:conf/ecai/MarlingB20}, there are more than 50 anonymous contributors, and they provid their daily events and data.
These people signed consent documents, enabling these data to be available to researchers for only research usage after de-identification. Hence,  the data were totally de-identified based on the Safe Harbor method.
On the other hand, researchers must sign a data use agreement (DUA) before they get access to the data.
Hence, OhioT1DM'18 and OhioT1DM'20 are public as long as signing the DUA, so there are no ethical problems with the datasets, and ethical approval is not needed.

\section{Data Preprocessing}
\label{sec:data_preprocessing}
The data is preprocessed based on the following steps.
\begin{itemize}
    \item \textbf{Loading data}. The original data is saved individually in ``{PID}-ws-training.xml'' and ``{PID}-ws-testing.xml'' files, e.g., ``540-ws-training.xml'' and ``540-ws-testing.xml files''. Then, the data is extracted from these .xml files and separately saved by the pandas data frame.
    \item \textbf{Removing attributes}. ``stressors'', ``illness'' and ``hypo\_event'' are removed from all attributes, as they are incredibly sparse. For example, in terms of ``illness'', each person only has around 1 samples, so it is removed because the quantity is far from enough.
    \item \textbf{Reorganising attributes}. Some attributes that can last for specific periods are discretized. For example, if ``sleep'' happens over the period between ts\_begin and ts\_end, it is re-sampled every 5 minutes for discretization. The attributes generating values multiple times within 5 minutes are averaged by 5 minutes, e.g., ``basis\_heart\_rate''. More details of processing these attributes will be explained later in this section, and the statistics of these attributes can be seen in Tables \ref{tab:statistics_attri_1} and \ref{tab:statistics_attri_2}.
    \item \textbf{Splitting, normalizing and padding}. The training data extracted from ``{PID}-ws-training.xml'' files are split into training data ($80\%$) and validation data ($20\%$) for early stopping and avoiding overfitting. Testing data from ``{PID}-ws-testing.xml'' are all leverage to evaluation of the well trained methods. All training data, validation data and testing data are normalized with the mean and standard derivation of the attributes in the training data through standard normalization, Then, all attributes are collected as RMTS by padding with zeros, where time interval between the adjacent tenors in RMTS is 5 minutes.
    \item \textbf{Generating training samples}. For the participant $p\in\mathcal{P}$, training samples extracted from training data are organized as $(\bX^p, y_{T+W}^p)\in\mathcal{D}^{p}_{Training}, \forall p$ which is obtained by a sliding window ($W=6$ or $W=12$), where $\mathcal{P}\triangleq\{1, ..., p, ..., P\}$ is the participants. Note that $y_{T+W}^p$ is not the padding value, i.e., zero.
    \item \textbf{Generating validation and testing samples}. Similarly, for the participant $p\in\mathcal{P}$, validation samples $\mathcal{D}^{p}_{Validation}$ and testing samples $\mathcal{D}^{p}_{Testing}$ are respectively extracted from the validation data and testing data. The general process is also similar to generating training samples. In order to ensure the number of testing samples is the same as the one in Table \ref{tab:statistics}, when generating testing samples, several data from the last validation data are leveraged.
\end{itemize}

Note that neither interpolation nor extrapolation is introduced in the data processing, considering both \cite{RubinFalconeFW20} and \cite{DBLP:conf/ecai/BevanC20} find that using zero padding can lead to more stable methods and more accurate predictions. 
This is because padding IMTS with zeros after standard normalization is equivalent to padding IMTS with the mean value.

\begin{table}[tb]

	\caption{Statistics of self-reported attributes in OhioT1DM'18 and OhioT1DM'20.}
	\label{tab:statistics_attri_1}
	\centering
	\resizebox{0.8\textwidth}{!}{
	    \begin{tabular}{ccccccc}
        \hline
        PID & finger\_stick (mg/dL)     & meal (g)             & bolus (U)          & sleep         & work          & exercise       \\ \hline
        559 & 185.70$\pm$100.95 & 35.93$\pm$15.99  & 3.66$\pm$1.83  & 2.25$\pm$0.56 & 4.48$\pm$0.79 & 5.17$\pm$0.99  \\
        563 & 165.39$\pm$64.00  & 29.90$\pm$16.31  & 7.96$\pm$4.10  & 2.60$\pm$0.49 & 3.11$\pm$0.56 & 4.52$\pm$1.23  \\
        570 & 199.21$\pm$68.74  & 105.25$\pm$42.16 & 7.58$\pm$3.75  & 2.13$\pm$0.51 & 1.85$\pm$0.55 & 5.02$\pm$0.74  \\
        575 & 157.99$\pm$82.20  & 40.57$\pm$22.96  & 4.83$\pm$2.56  & 2.71$\pm$0.53 & 4.27$\pm$0.99 & 5.53$\pm$1.05  \\
        588 & 153.66$\pm$54.96  & 32.52$\pm$31.00  & 4.27$\pm$2.31  & 2.78$\pm$0.41 & 4.82$\pm$0.53 & 5.22$\pm$0.46  \\
        591 & 163.10$\pm$68.40  & 31.43$\pm$14.01  & 3.16$\pm$1.76  & 2.71$\pm$0.49 & None          & 4.96$\pm$1.48  \\
        540 & 155.22$\pm$72.45  & 55.50$\pm$29.03  & 3.93$\pm$2.47  & None          & None          & None           \\
        544 & 140.70$\pm$56.89  & 70.94$\pm$36.87  & 12.19$\pm$5.21 & 1.94$\pm$0.45 & 3.43$\pm$0.85 & None           \\
        552 & 152.42$\pm$60.72  & 54.00$\pm$32.15  & 3.82$\pm$3.22  & 2.64$\pm$0.65 & 2.03$\pm$0.67 & 7.52$\pm$1.02  \\
        567 & 185.79$\pm$65.34  & 75.44$\pm$20.40  & 11.93$\pm$5.66 & 2.01$\pm$0.77 & None          & None           \\
        584 & 205.34$\pm$63.76  & 54.79$\pm$12.38  & 7.19$\pm$3.08  & 2.97$\pm$0.18 & 5.05$\pm$0.22 & 10.00$\pm$0.00 \\
        596 & 140.58$\pm$42.15  & 25.15$\pm$13.29  & 3.01$\pm$1.44  & 2.82$\pm$0.46 & None          & 5.04$\pm$1.08  \\ \hline
        \end{tabular}
	}
	\vspace{-0.1cm}
\end{table}

\begin{table}[tb]

	\caption{Statistics of attributes collected by sensors in OhioT1DM'18 and OhioT1DM'20.}
	\label{tab:statistics_attri_2}
	\centering
	\resizebox{1.0\textwidth}{!}{
        \begin{tabular}{cccccccccc}
        \hline
        PID & glucose\_level (mg/dL)   & basal (U/h)         & basis\_gsr ($\mu S$)    & basis\_skin\_temperature ($\degree F$) & acceleration ($m/s^2$)  & basis\_sleep  & basis\_air\_temperature ($\degree F$) & basis\_steps   & basis\_heart\_rate (BPM) \\ \hline
        559 & 167.23$\pm$69.91 & 0.96$\pm$0.32 & 0.40$\pm$2.04 & 87.65$\pm$3.43           & None          & 1.00$\pm$0.00 & 84.27$\pm$4.37          & 3.65$\pm$15.05 & 73.99$\pm$16.04    \\
        563 & 149.80$\pm$49.75 & 0.83$\pm$0.24 & 0.41$\pm$1.68 & 87.94$\pm$2.71           & None          & 1.00$\pm$0.00 & 84.02$\pm$3.60          & 4.85$\pm$11.33 & 96.45$\pm$13.89    \\
        570 & 192.95$\pm$64.10 & 0.93$\pm$0.19 & 0.55$\pm$2.61 & 86.42$\pm$3.66           & None          & 1.00$\pm$0.00 & 82.52$\pm$4.13          & 2.73$\pm$8.30  & 83.49$\pm$11.88    \\
        575 & 143.34$\pm$60.40 & 0.74$\pm$0.12 & 0.29$\pm$1.53 & 87.31$\pm$3.10           & None          & 1.00$\pm$0.00 & 83.76$\pm$3.87          & 2.81$\pm$7.07  & 81.12$\pm$11.51    \\
        588 & 166.82$\pm$50.33 & 1.20$\pm$0.20 & 0.06$\pm$0.79 & 90.18$\pm$4.08           & None          & 1.00$\pm$0.00 & 87.76$\pm$5.33          & 2.23$\pm$10.74 & 76.89$\pm$14.28    \\
        591 & 153.75$\pm$56.94 & 0.96$\pm$0.17 & 2.27$\pm$5.07 & 87.86$\pm$2.99           & None          & 1.00$\pm$0.00 & 84.97$\pm$3.78          & 3.15$\pm$10.15 & 66.86$\pm$11.69    \\
        540 & 141.35$\pm$58.21 & 0.48$\pm$0.33 & 1.16$\pm$5.22 & 86.95$\pm$4.17           & 1.02$\pm$0.07 & 1.00$\pm$0.00 & None                    & None           & None               \\
        544 & 163.44$\pm$58.87 & 1.46$\pm$0.25 & 1.70$\pm$4.58 & 87.91$\pm$3.10           & 1.73$\pm$3.18 & 1.00$\pm$0.00 & None                    & None           & None               \\
        552 & 145.47$\pm$54.23 & 1.22$\pm$0.24 & 0.29$\pm$0.90 & 84.15$\pm$3.21           & 1.03$\pm$0.08 & 1.00$\pm$0.00 & None                    & None           & None               \\
        567 & 152.46$\pm$59.97 & 1.07$\pm$0.28 & 1.14$\pm$3.23 & 89.61$\pm$3.40           & 1.00$\pm$0.06 & 1.00$\pm$0.00 & None                    & None           & None               \\
        584 & 188.50$\pm$65.19 & 1.66$\pm$0.22 & 4.03$\pm$8.56 & 84.66$\pm$3.09           & 0.97$\pm$0.06 & 1.00$\pm$0.00 & None                    & None           & None               \\
        596 & 147.33$\pm$49.52 & 0.47$\pm$0.07 & 0.25$\pm$1.44 & 87.12$\pm$2.73           & 1.04$\pm$0.03 & 1.00$\pm$0.00 & None                    & None           & None               \\ \hline
        \end{tabular}
	}
	\vspace{-0.1cm}
\end{table}

The brief introduction of selected attributes and the preprocessing details are as follows.
\begin{itemize}
    \item glucose\_level: the BG levels (mg/dL) are collected by CGM per 5 minutes. In this paper, after preprocessing, by padding missing values with zeros, the time interval between the adjacent points is 5 minutes.
    \item finger\_stick: the BG levels (mg/dL) are obtained by finger stick, a self-monitoring method.
    \item basal: the rate (U/h) of insulin delivery of pump. The rate is fixed from a particular timestamp ``ts'', and it is changed to another value at the next timestamp. ``temp\_basal'' is the temporary rate of insulin delivery between ``ts\_begin'' and ``ts\_end'', when the normal basal is suspended. Hence, ``basal'' and ``temp\_basal'' are merged as ``basal''. Then, the basal is re-sampled every 5 minutes.
    \item bolus: it is a kind of insulin dose (U), including two types. The first type is to deliver all insulin at a time. The second type is to extend insulin delivery within a period from ``ts\_begin'' to ``ts\_end''. Then, the second type of bolus is re-sampled every 5 minutes.
    \item meal: it is the carbohydrate intake (g) of a meal self-reported by the participants. The type of meal, i.e., breakfast, lunch, dinner, snacks, etc., is also recorded but ignored in this paper. This is because the type of meal can be regarded as a piece of redundant information when considering temporal features, i.e., time of a day, in the input.
    \item sleep: a participant falls asleep between ``ts\_begin'' and ``ts\_end'', and ``sleep'' is a self-reported attribute. The participant also estimates the sleep quality from 1 to 3, where 3 means good sleep, 1 is poor sleep, and 2 is in the middle.
    \item work: a participant works between ``ts\_begin'' and ``ts\_end'', and ``work'' is a self-reported attribute. The intensity of the work is estimated by the participant from 1 to 10, where 10 means working in extremely high intensity.
    \item exercise: a participant exercises between ``ts\_begin'' and ``ts\_end'', and ``exercises'' is a self-reported attribute. The intensity of the exercise is estimated by the participant from 1 to 10, where 10 means exercising at extremely high intensity.
    \item basis\_heart\_rate: it is the heart rate (BPM) of the participants, and it is collected by Basis Peak per 1 minute. Then, it is aggregated per 5 minutes by averaging all values within the 5 minutes.
    \item basis\_gsr: it is the galvanic skin response ($\mu$S) of the participants, and it is collected by Basis Peak per 5 minutes or Empatica Embrace per 1 minute. Then, the values collected per 1 minute are aggregated per 5 minutes by averaging all values with in the 5 minutes.
    \item basis\_skin\_temperature: it is the skin temperature ($\degree$F) of the participants, and it is collected by Basis Peak per 5 minutes or Empatica Embrace per 1 minute. Then, the values collected per 1 minute are aggregated per 5 minutes by averaging all values within the 5 minutes.
    \item basis\_air\_temperature: it is the air temperature ($\degree$F) of the participants, and it is collected by Basis Peak per 1 minute. Then, it is aggregated per 5 minutes by averaging all values within the 5 minutes.
    \item basis\_steps: it is the step count of the participants every 5 minutes, and it is collected by Basis Peak.
    \item acceleration: it is the magnitude of acceleration ($m/s^2$) of the participants, and it is collected by Empatica Embrace per 1 minute. Then, it is aggregated per 5 minutes by averaging all values within the 5 minutes.

\end{itemize}

In terms of the way of collecting values, the attributes can be divided into two categories: 1) attributes that are self-reported (see Table \ref{tab:statistics_attri_1}); 2) attributes that are collected by sensors (see Table \ref{tab:statistics_attri_2}).
From Tables \ref{tab:statistics_attri_1} and \ref{tab:statistics_attri_2}, we can have the following observations.
\begin{itemize}
    \item For each attribute, the distributions are different among various participants. For example, patient 584 tends to have worse BG management, as the mean ``glucose\_level'' is hyperglycemia ($>$180 mg/dL). Patient 544 has the best BG management since the mean ``glucose\_level'' is the lowest and within a normal range (70-180 mg/dL).
    \item Participants (559-591) from OhioT1DM'18 do not have ``acceleration'', while participants (540-596) from OhioT1DM'20 do not have ``basis\_air\_temporature'', ``basis\_steps'', ``basis\_heart\_rate''.
\end{itemize}

\section{Data Analysis}
\label{sec:data_analysis}
\subsection{Impact of ``meal'' and ``bolus''}
\begin{figure}[tb]
	\centering
	\includegraphics[width = 1.0\columnwidth]{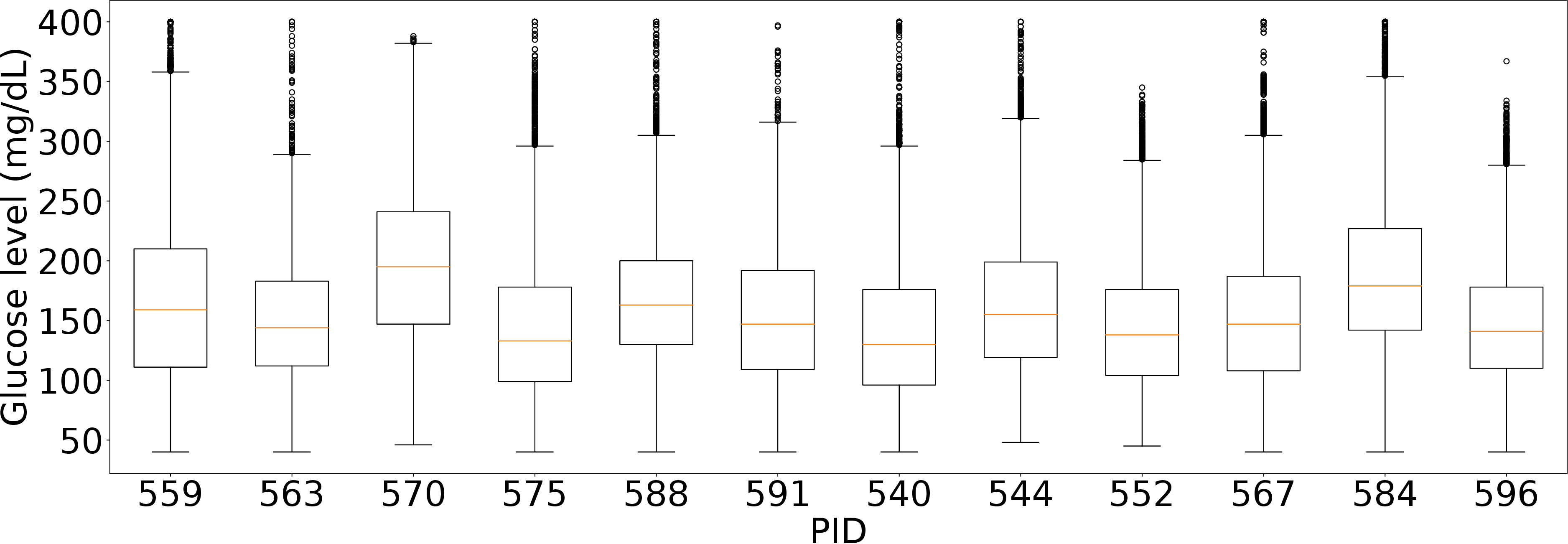}
	\caption{Distributions of ``glucose\_level'' in terms of all participants.}
	\label{fig:glucose_level}
\end{figure}
\begin{figure}[tb]
	\centering
	\includegraphics[width = 1.0\columnwidth]{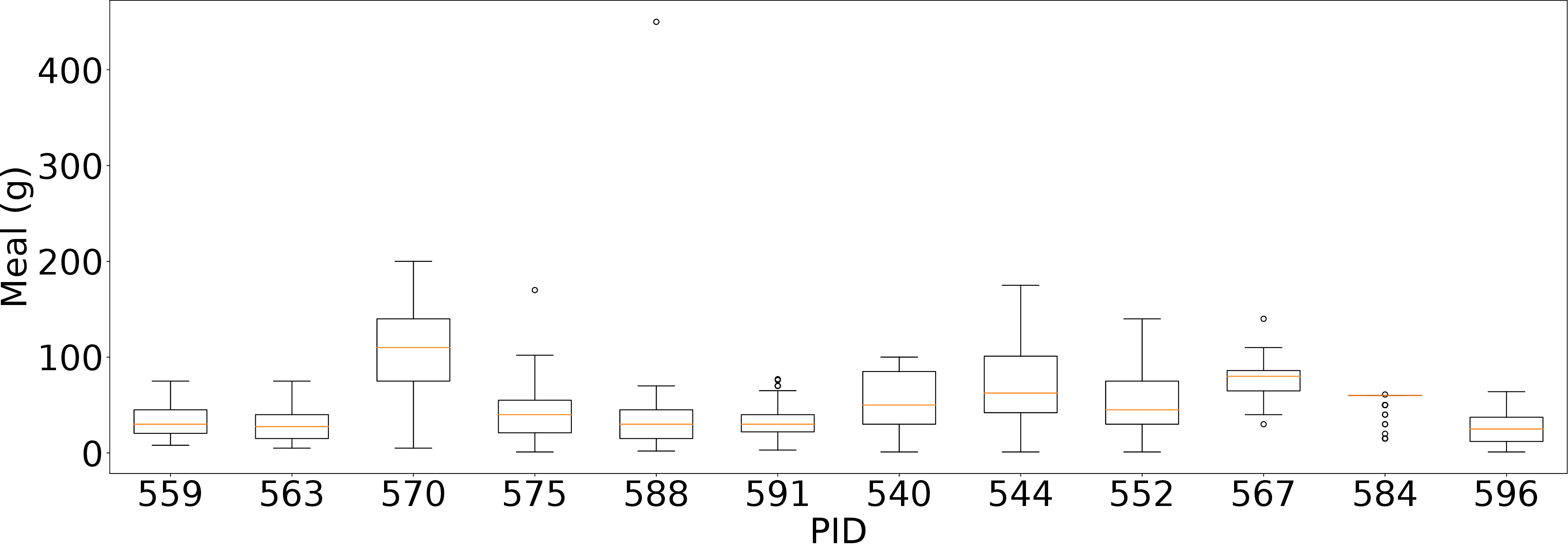}
	\caption{Distributions of ``meal'' in terms of all participants.}
	\label{fig:meal}
\end{figure}
\begin{figure}[tb]
	\centering
	\includegraphics[width = 1.0\columnwidth]{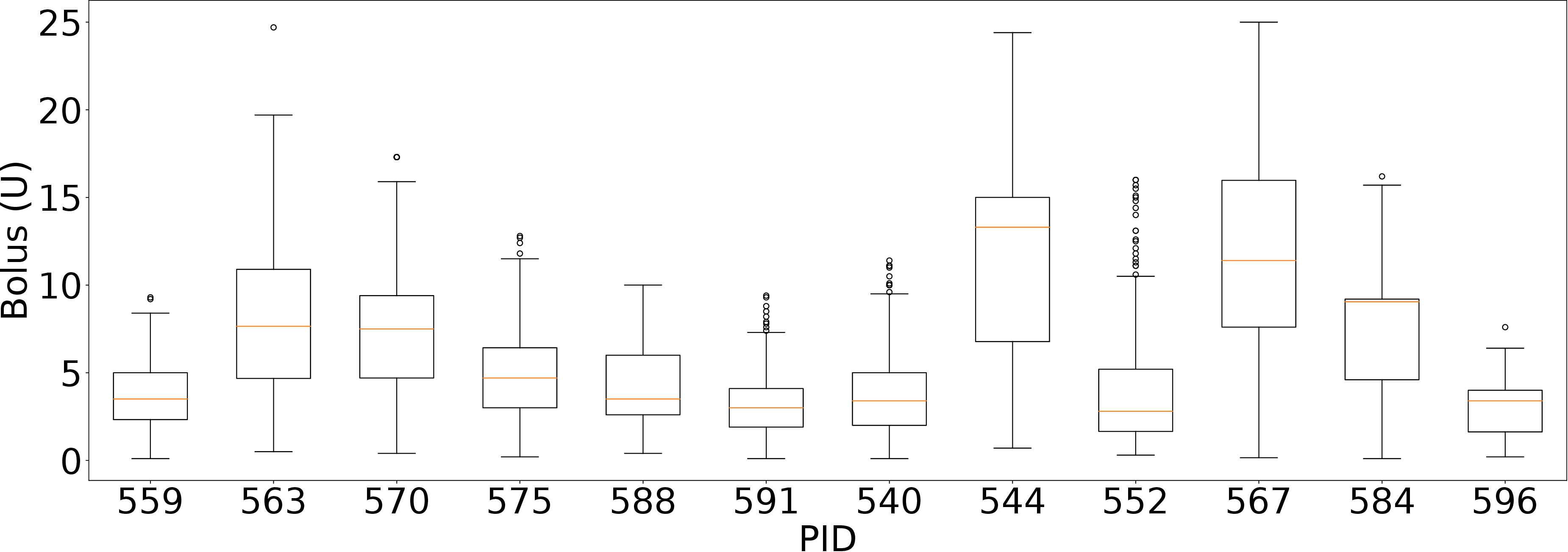}
	\caption{Distributions of ``bolus'' in terms of all participants.}
	\label{fig:bolus}
\end{figure}

\begin{table}[tb]

	\caption{Proportions of Hypoglycemia, Normal and Hyperglycemia for each participant.}
	\label{tab:hypo_hper}
	\centering
	\resizebox{0.5\textwidth}{!}{
        \begin{tabular}{cccc}
        \hline
        PID & Hypoglycemia & Normal  & Hyperglycemia \\ \hline
        559 & 3.61\%       & 56.71\% & 39.68\%       \\
        563 & 1.99\%       & 71.79\% & 26.21\%       \\
        570 & 1.49\%       & 40.49\% & 58.02\%       \\
        575 & 7.60\%       & 68.21\% & 24.18\%       \\
        588 & 0.80\%       & 61.98\% & 37.22\%       \\
        591 & 3.93\%       & 64.89\% & 31.18\%       \\
        540 & 6.02\%       & 71.00\% & 22.98\%       \\
        544 & 1.30\%       & 64.73\% & 33.97\%       \\
        552 & 3.28\%       & 73.56\% & 23.16\%       \\
        567 & 6.67\%       & 64.72\% & 28.60\%       \\
        584 & 0.90\%       & 50.04\% & 49.06\%       \\
        596 & 1.99\%       & 74.32\% & 23.69\%       \\ \hline
        \end{tabular}
	}
	\vspace{-0.1cm}
\end{table}
\begin{figure}[tb]
	\centering
	\includegraphics[width = 1.0\columnwidth]{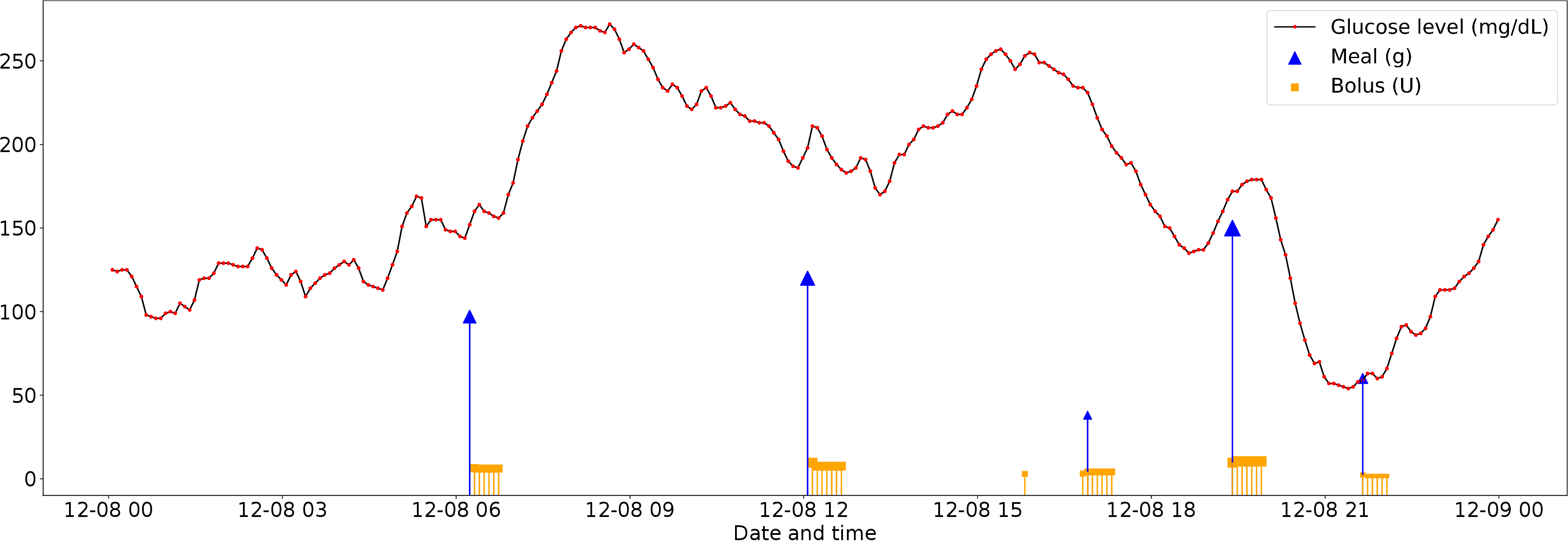}
	\caption{Visualizing ``glucose\_level'', ``bolus'' and ``meal'' of a day for participant 570.}
	\label{fig:show_570_day}
\end{figure}
In this section, we first visualize the distribution of ``glucose\_level'' (Figure \ref{fig:glucose_level}), ``meal'' (Figure \ref{fig:meal}) and ``bolus'' (Figure \ref{fig:bolus}).

Overall, we can observe that some participants have severe hyperglycemia. For example, 58.02\% of the samples of  participant 570 are hyperglycemia (see Table \ref{tab:hypo_hper}).
This is because participant 570 has the highest carbohydrate intake (see Figure \ref{fig:meal}) but relatively lower bolus delivery compared with other participants.

Comparably, we can also find that even though participants 544 and 567 have high carbohydrate intake (see Figure \ref{fig:meal}), their BG levels can maintain relatively normal (see Figure \ref{fig:glucose_level}). Only 33.97\% and 28.60\% of the samples of participant 544 and 567 are hyperglycemia, respectively (see Table \ref{tab:hypo_hper}).
This is because they have high insulin delivery (see Figure \ref{fig:bolus}), enabling them to reduce their BG levels even if they have high carbohydrate intake.

Furthermore, for participant 570, we visualize the variations of ``glucose\_level'', ``meal'' and ``bolus'' of a day (see Figure \ref{fig:show_570_day}). 
We can observe that, in most cases, insulin is delivered after having a meal. Besides, BG levels tend to increase in the next few hours after a meal.
With the help of insulin, there is always a downtrend following the uptrend of the BG levels.

In summary, both ``meal'' and ``insulin'' have a significant influence on the ``glucose\_level''. Hence, we can guess that maybe ``meal'' and ``insulin'' are helpful for the prediction of future BG levels.

\subsection{Impact of Temporal Features}
\label{sec:imp_of_tf}
\begin{figure}[tb]
	\centering
	\includegraphics[width = 1.\columnwidth]{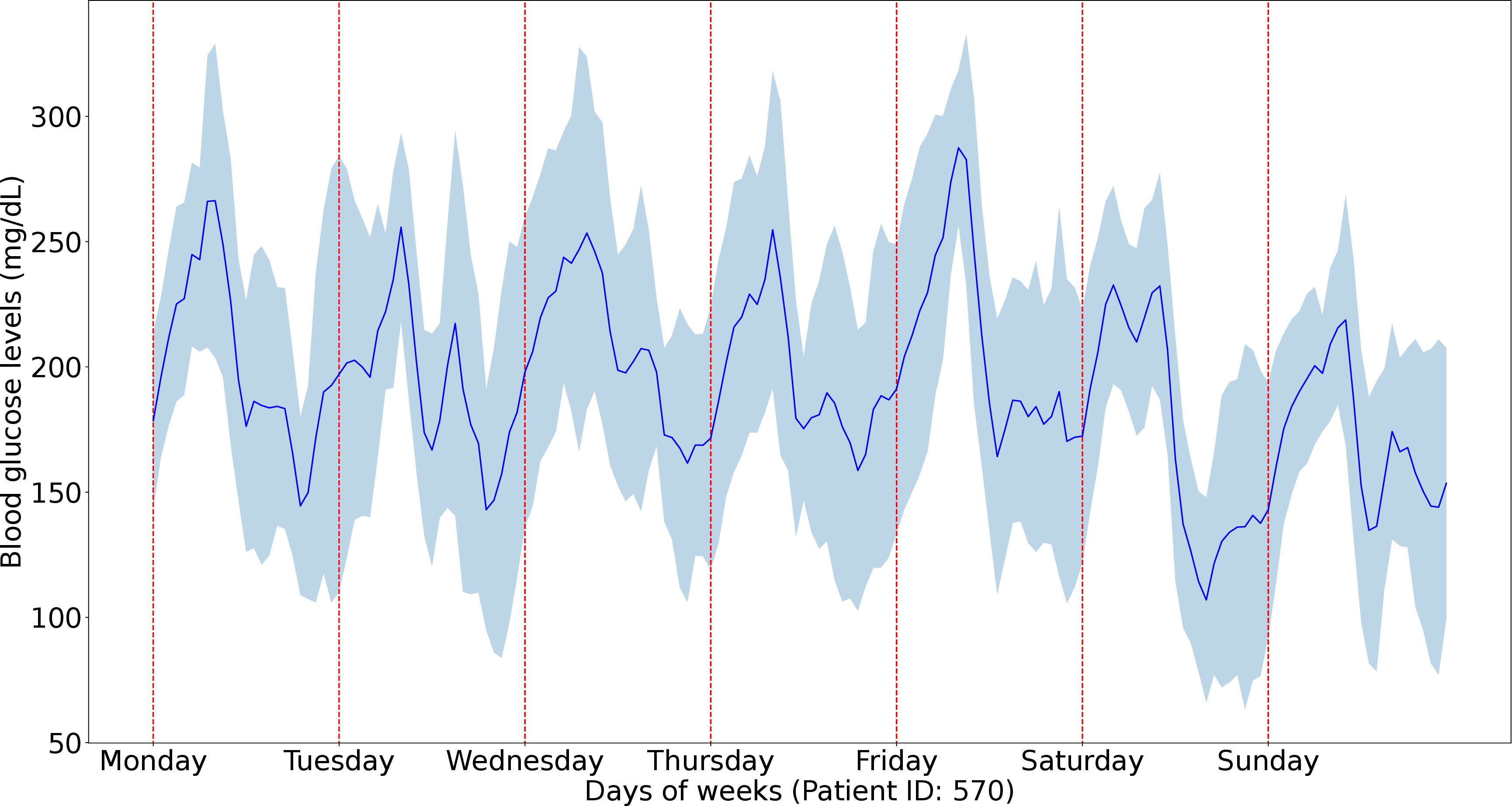}
	\caption{Visualizing ``glucose\_level'' for participant 570 in terms of days of weeks.}

	\label{fig:plot_570_days}
\end{figure}

\begin{figure}[tb]
	\centering
	\includegraphics[width = 1.\columnwidth]{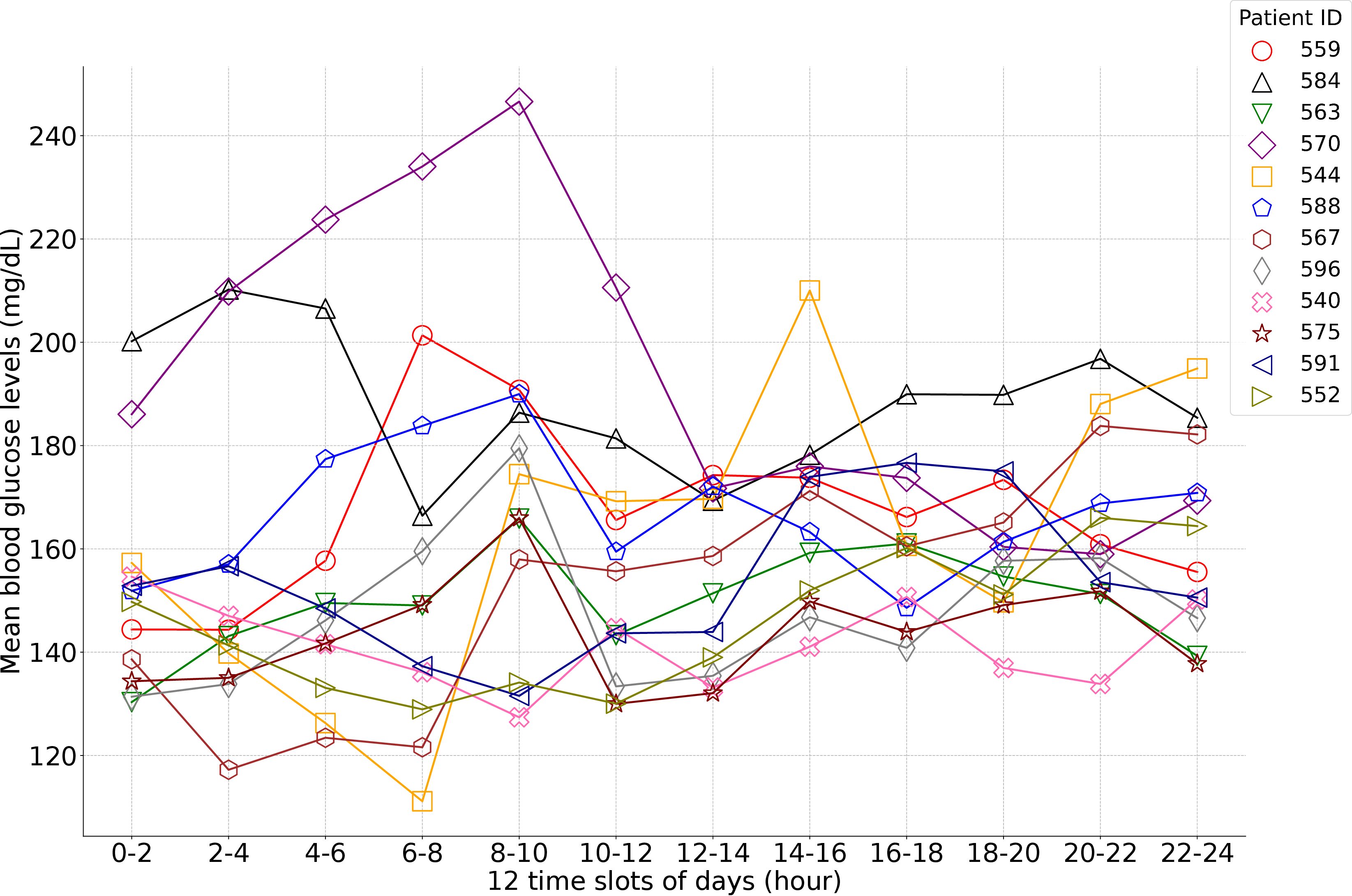}
	\caption{Visualizing ``glucose\_level'' for all participant in terms of time slots of days.}
	\label{fig:plot_day}
\end{figure}

\begin{figure}[tb]
	\centering
	\includegraphics[width = 1.\columnwidth]{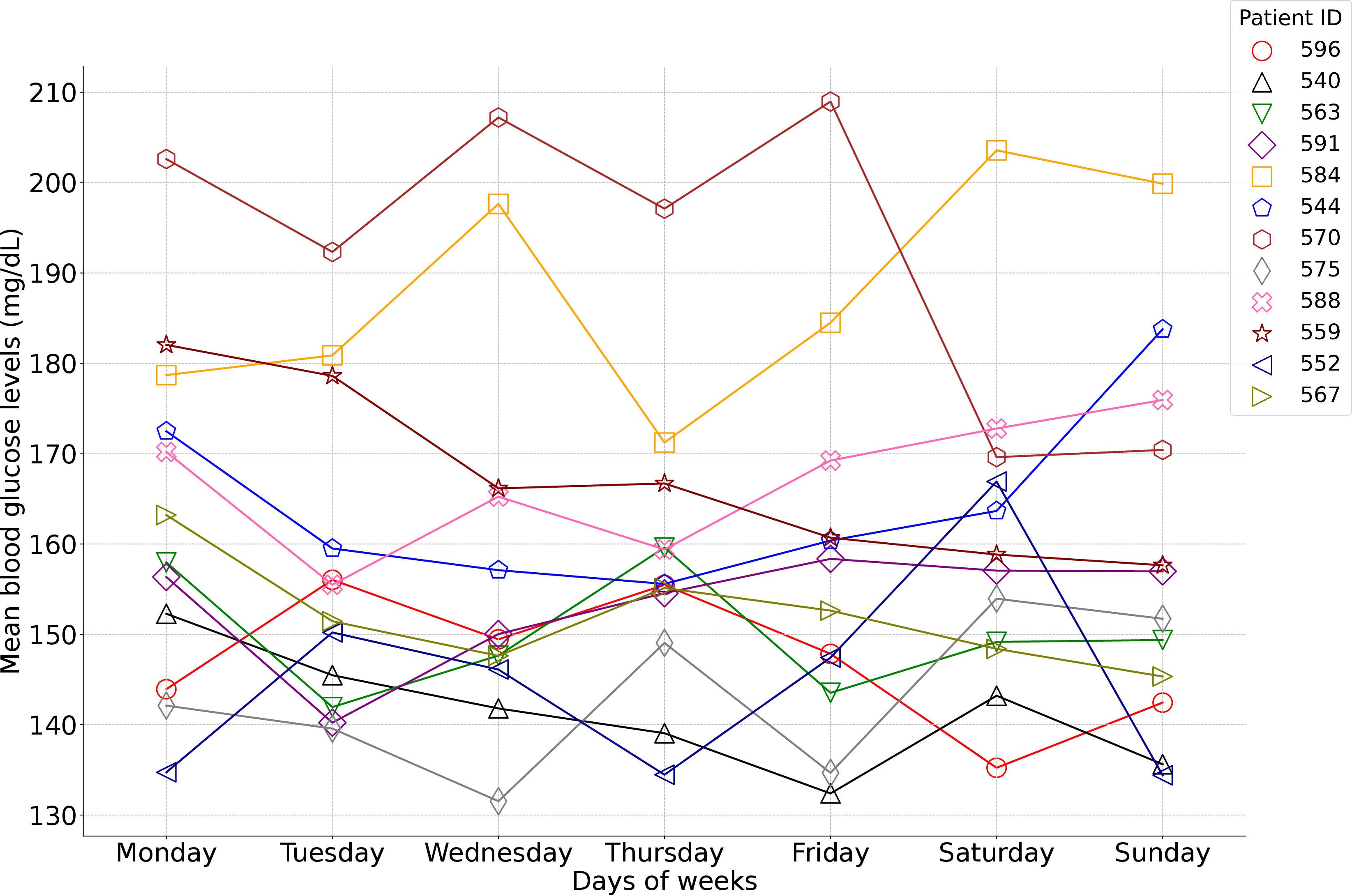}
	\caption{Visualizing ``glucose\_level'' for all participant in terms of days of weeks.}

	\label{fig:weekplot}
\end{figure}
In order to figure out the temporal features of ``glucose\_levels'', we visualize the variations of ``glucose\_level'' from different temporal granularity. We find that most participants have interesting and special temporal features.
For example, Figure \ref{fig:plot_570_days} contains all ``glucose\_level'' samples of patient 570, organized sequentially by time.
The darker blue line is the mean value of ``glucose\_level''.

We can find that for each day, the BG levels tend to be high in the first half and low in the second half.
It is obviously a daily pattern.
On the other hand, we can find that the participant tends to have higher BG levels during weekdays compared with weekends.
It is a weekly pattern.

Moreover, we visualize the variations in the ``glucose\_level'' of all participants in terms of time slots of days (see Figure \ref{fig:plot_day}) and days of weeks (see Figure \ref{fig:weekplot}).
People have daily and weekly patterns, which different lifestyles and body conditions may cause.

Finally, according to the observations above, temporal features, i.e., times of days and days of weeks, are also considered in the following experiments. More relevant experiments will show whether temporal features help predict future BG levels.
Given that each person has his or her BG changing styles, personalized model will be designed in the following sections. 

\section{Proposed method}
\label{sec:proposed_method}
\subsection{Overview}
\begin{figure}[tb]
	\centering
	\includegraphics[width = 1.\columnwidth]{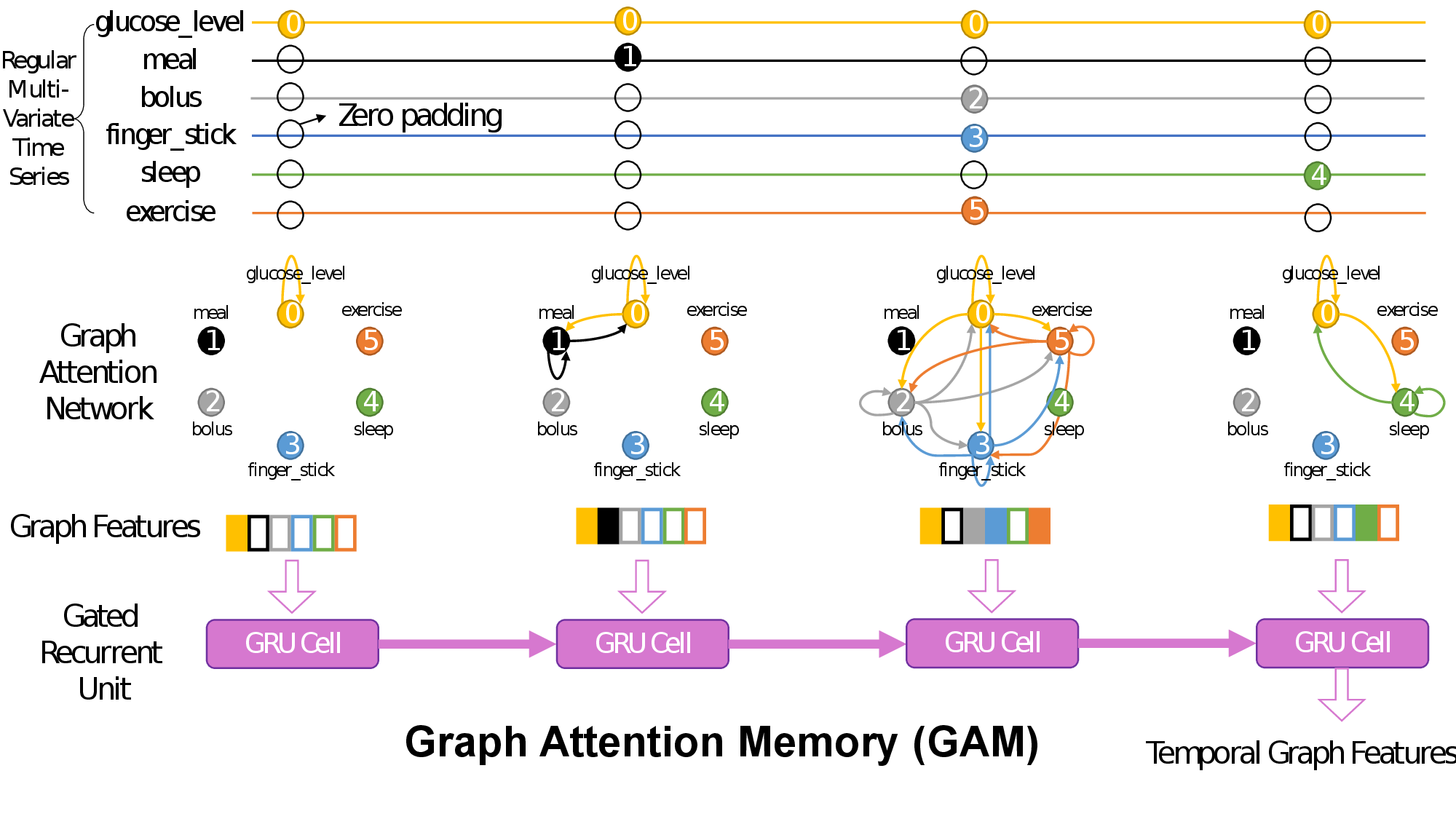}
	\caption{Graph attention memory (GAM).}

	\label{fig:gam}
\end{figure}

In this paper, we proposed a graph attentive memory neural network, called ``GAM'' and trained by FL.
It can be seen in Figure \ref{fig:gam}, consisting of a graph attention network (GAT, \cite{DBLP:conf/iclr/VelickovicCCRLB18}) and a gated recurrent unit (GRU, \cite{DBLP:conf/emnlp/ChoMGBBSB14}).
GAT is leveraged to explicitly model correlations among various attributes in an RMTS generated from IMTS padded with zeros.
Each valid value of an attribute, excluding zeros, can be seen as a node of a dynamic graph, and the value is mapped to a tensor by fully connected layers, denoted as graph embeddings.
Then, at each time slot, graph features are gotten from GAT via graph convolutions on graph embeddings given an adjacent matrix.
These graph features are sequentially aggregated by GRU, generating a compact vector containing temporal graph features.
The temporal graph features will be used for the estimation of BG levels.

On the other hand, the overview of $\hat{y}_{T+W} = {\rm GAM}(\bX; \mathbf{\theta})$ are as follows ($\forall t\in\mathcal{T}, \forall n\in\mathcal{N}$):
\begin{subequations}
\label{Eqn:model}
\begin{align}
    \label{eqn:model_1}
    &\be_t^n= {\rm f}_n(x_t^n), \\
    \label{eqn:model_2}
    &\bE_t= [\be_t^1 \ ... \ \be_t^n \ ...\ \be_t^N],   \\
    \label{eqn:model_3}
    &\bG_t = {\rm GATs}(\bE_t, \bA), \\
    \label{eqn:model_4}
    &\bh_t = {\rm GRU}(\phi(\bG_t), \bh_{t-1}), \\
    \label{eqn:model_5}
    &\hat{y}_{T+W} = {\rm f}_{out}(\bh_T), 
\end{align}
\end{subequations}
where $\mathcal{T} \triangleq \{1, ..., t, ..., T\}$ and $\mathcal{N} \triangleq \{1, ..., n, ..., N\} $; $x_t^n$ is the item in the $n$-th row $t$-th column of $\bX\in\mathbb{R}^{N\times T}$ which is the RMTS defined in Section \ref{sec:problem_def}; ${\rm f}_n(\cdot)$ and ${\rm f}_{out}(\cdot)$ are fully connected layers; $\bA\in\mathbb{R}^{N\times N}$ is the adjacent matrix of a graph; $\bh_t$ is the hidden state of GRU, and $\bh_0$ is a zero tensor; $\hat{y}_{T+W}$ is the estimation of ${y}_{T+W}$.

\begin{figure}[tb]
	\centering
	\includegraphics[width = 1.\columnwidth]{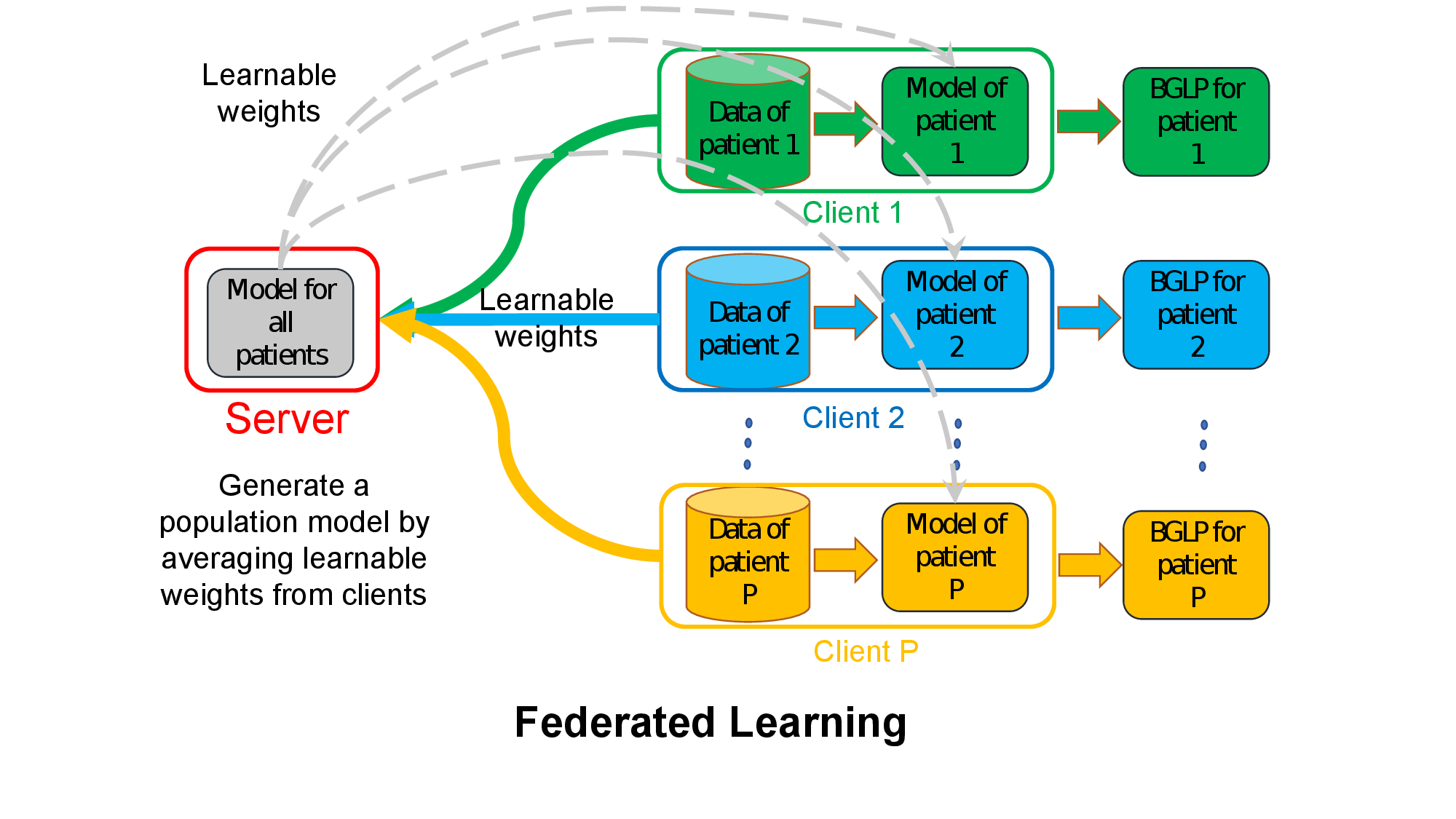}
	\caption{Federated learning (FL).}

	\label{fig:fl}
\end{figure}

Firstly, in Equation (\ref{eqn:model_1}), each item $x_t^n$, excluding zeros, of RMTS $\bX$ is individually embedded by fully connected layers ${\rm f}_n(\cdot)$, where each attribute $n\in\mathcal{N}$ has its own ${\rm f}_n(\cdot)$. 
The embedding $\be_t^n\in\mathbb{R}^{E}$ is a tensor representation of $x_t^n$, and it is collected by $\bE_t\in\mathbb{R}^{E\times N}$ in Equation (\ref{eqn:model_2}).
Then, for each time slot $t$, in Equation (\ref{eqn:model_3}), GATs are leveraged to do the graph convolution for $\bE_t$ when given the adjacent matrix $\bA$.
More details are explained in section \ref{sec:gat}.
After graph convolutions, $\bG_t\in\mathbb{R}^{E^\prime \times N}$ is generated, containing correlations among all the attributes.
Next, $\bG_t$ is flattened to a tensor by $\phi(\cdot)$, as $\phi(\bG_t)\in\mathbb{R}^{E^\prime N\times 1}$. Then, $\{\phi(\bG_1), ..., \phi(\bG_t), ..., \phi(\bG_T) \}$  aggregated by GRU in Equation (\ref{sec:gru}), and more details are explained in section \ref{sec:gru}, generating a compact tensor $\bh_T$ which contains temporal information.
Finally, the estimated glucose level $\hat{y}_{T+W}$ is generated from $\bh_T$ through fully connected layers ${\rm f}_{out}(\cdot)$.

The above structure has the following advantages.
\begin{itemize}
    \item Graph convolutions by GAT can model complex correlations among attributes, focusing on more valuable attributes and ignoring noises.
    \item GRU-based memory can further aggregate temporal information, reducing irrelevant information by a self-gated mechanism. 
    \item GAM fully leverages attribute and temporal information by combining GAT and GRU in the BGLP. 
\end{itemize}

Mean square error (MSE) loss is leveraged in the following training process, as:
\begin{equation}
    \label{Eqn:loss}
    {\rm L}(y_{T+W, b}, \hat{y}_{T+W, b}; \theta) = \frac{1}{B}\sum_b(y_{T+W, b} - \hat{y}_{T+W, b})^2,
\end{equation}
where $B$ is the batch size, $\theta$ is the learnable weights of the proposed model.
Then, Adam \citep{DBLP:journals/corr/KingmaB14} optimizer is introduced to update learnable weights.

The training of the GAM is explained in section \ref{sec:train} and section \ref{sec:train_fl}.
Especially federated learning (FL) is generally depicted in Figure \ref{fig:fl}.
In FL, personal data can only be accessed by the related personalized model in the clients.
After training for certain epochs in the clients, the learnable weights of the personal model are sent to a server, and a population model is calculated by averaging the latest personal learnable weights.
Then, personalized models of clients copy the latest learnable weights of the population model, and each client replaces the personalized learnable weights with the population learnable weights. 
The above process is repeated until the convergence.
In this case, the participant privacy is well protected by the clients, as the personal data is only kept by the clients separately.
Meanwhile, population patterns are also learned in the server without invasion of privacy.
Note that we leverage multi-process to simulate FL in this paper, where each client or the server is an independent process.

\subsection{Graph Attention Networks (GATs)}
\label{sec:gat}
For each time slot $t$, node embedding $\bE_t$ can be gotten from Equation (\ref{eqn:model_2}).
Then, each attribute $n\in\mathcal{N}$ has its own embedding $\be_t^n$, and all attribute embedding tensors can be treated as a tensor graph $<\bE_t, \bA>$, where $\bA$ is the adjacent matrix.
In this case, each valid attribute connects all other valid attributes.
For example, when $i,j \in \mathcal{N}$, if $\be_t^i$ and $\be_t^j$ are not null, both $\bA[i,j]$ and $\bA[j,i]$ are equal to 1.
That is to say. There are directed edges, respectively, from node $j$ to node $i$ and from $i$ to node $j$.
Otherwise, when any tensor of $\be_t^i$ and $\be_t^i$ is null, both $\bA[i,j]$ and $\bA[j,i]$ are equal to 0.
It means that there are no directed edges between node $i$ and node $j$.
Note that all valid attributes connect to themselves. In other words, $\bA[i,i] = 1$ when $\be_t^i$ is not null.

Then, GAT is introduced to this work, and it can be represented as follows ($\forall t\in\mathcal{T}, \forall i\in\mathcal{N}$):
\begin{subequations}
\label{Eqn:gat}
\begin{align}
    \label{eqn:gat_1}
    &\alpha_t^{i,j}= \frac{\exp\left({\rm LeakyReLU}\left(\ba^{\mathsf{T}}[\bW\be_t^i; \bW\be_t^j]\right)\right)}{\sum_{k\in\mathcal{N}_i}\exp\left({\rm LeakyReLU}\left(\ba^{\mathsf{T}}[\bW\be_t^i; \bW\be_t^k]\right)\right)},  \\
    \label{eqn:gat_2}
    &\bg_t^i= \sigma\left(\sum_{j\in\mathcal{N}_i}\alpha^{i,j}_t\bW \be_t^j\right),  \\
    \label{eqn:gat_3}
    &\bG_t = [\bg_t^1 \ ... \ \bg_t^i\ ...\ \bg_t^N], 
\end{align}
\end{subequations}
where $\bW\in\mathbb{R}^{E^\prime \times E}$ and $\ba\in\mathbb{R}^{2E^\prime}$ contains learnable parameters; $[;]$ meas the concatenation of two tensors; $\mathcal{N}_i$ contains all adjacent nodes of the node $i$.

Each node tensor $\be_t^j$ can generate a neural message $\bW\be_t^j$ which can be passed along the edges to the adjacent nodes.
For example, when $\bA[i,j] = 1$, neural message $\bW\be_t^j$ can be passed to node $i$.
Meanwhile, if $\bA[i,j] = 0$, neural message $\bW\be_t^j$ cannot be passed to node $i$.
Only when the node tensor $\be_t^j$ is not null can the node tensor be transformed into a neural message by $\bW$.

Hence, at the end of time slot $t$, each node might receive several neural messages from the adjacent nodes.
For example, in Equation (\ref{eqn:gat_2}), the adjacent nodes of node $i$ are denoted as $\mathcal{N}_i$, then the neural messages $\bW \be_t^j$ from the adjacent nodes ($j\in\mathcal{N}_i$) are weighted and summarized by the attention weights $\{\alpha_t^{i,j}| j\in\mathcal{N}_i\}$ which are generated in Equation (\ref{eqn:gat_1}).

Specifically, in Equation (\ref{eqn:gat_1}), a similarity score between the neural messages $\bW\be_t^i$ and $\bW\be_t^j$ are calculated by ${\rm LeakyReLU}\left(\ba^{\mathsf{T}}[\bW\be_t^i; \bW\be_t^j]\right)$.
The edge weight from node $j$ to node $i$, i.e., attention weight $\alpha^{i,j}_t$, is calculated by normalizing the similarity score through ``softmax''.

Finally, after Equation (\ref{eqn:gat_1}) and Equation (\ref{eqn:gat_2}), each node $i$ has a new tensor $\bg_t^i$ which is a substitute for $\be_t^i$, and the new tensor contains all information of the previous tensor $\be_t^i$ and the adjacent node tensors $\{\be_t^j|j\in\mathcal{N}_i\}$.
Then, $\{\bg_t^i|i\in\mathcal{N}\}$ are collected and reorganised as new graph node embedding $\bG_t$ as Equation (\ref{eqn:gat_3}).
When there are multiple GATs layers ($\{1, ..., l , ..., L\}$), the input $\bE_t^l$ of the current GAT layer $l$ is the output $\bG_t^{l-1}$ of the previous layer $l-1$, where $\bE_t^1=\bE_t,\ \bE_t^l=\bG_t^{l-1},\ l>1$.


On the other hand, for the neural message generating, multi-head attention is also introduced as follows: 
\begin{equation}
    \label{Eqn:multi_head}
    \bg_t^i= \sigma\left(\frac{1}{M}\sum_m\bg_t^{i, m}\right) = \sigma\left(\frac{1}{M}\sum_m\sum_{j\in\mathcal{N}_i}\alpha^{i,j,m}_t\bW^m \be_t^j\right). 
\end{equation}
Each node $j$ can generate multiple neural messages $\{\bW^1 \be_t^j, ..., \bW^m \be_t^j, ..., \bW^M \be_t^j\}$, where $M$ is the number of heads.
Each head $m$ has its own new neural tensor $\bg_{t}^{i,m}$, and the final new neural tenor $\bg_{t}^{i}$ is calculated by averaging $\bg_{t}^{i,m}$ along $m$.

GAT has the following advantages.
\begin{itemize}
    \item The attention weights in the graph edges are flexible and change according to the comparison between the neural messages. It is able to focus/ignore different attributes in different time slots dynamically. Compared with GCN, whose edge weights are fixed, GAT-based method are more robust and flexible.
    \item Similar to CNN, when controlling the number of GAT layers $L$, the granularity of features can be selected. When being given larger $L$, features with more details are extracted. On the other hand, when $L$ is small, general features are captured by GATs.
    \item By leveraging graph-based structure, the method becomes explainable, as it is obvious to observe how different attributes affect each other and how much the impact is through visualizing attention weights and graph structures. Besides, only the valid attributes (not null) can generate and pass neural messages to the valid adjacent tensor nodes (not null), so this method effectively deals with RMTS. 
\end{itemize}

\subsection{Gated Recurrent Units (GRU) based Memory}
\label{sec:gru}
After passing through GATs in all historical time slots $\mathcal{T}$, a set of new graph node tensors are generated, as $\{\bG_t|t\in\mathcal{T}\}$.
The graph tensors $\bG_t\in\mathbb{R}^{E^\prime \times T}$ are reshaped to $\phi(\bG_t)\in\mathbb{R}^{E^\prime T \times 1}$ by flattening $\phi(\cdot)$.
GRU-based memorization is introduced to deal with $\{\phi(\bG_t)|t\in\mathcal{T}\}$, as:
\begin{subequations}
\label{Eqn:gru}
\begin{align}
    \label{eqn:gru_1}
    &\br_t = \sigma(\bW_1\phi(\bG_t) + b_1 + \bW_2\bh_{t-1} + b_2),  \\
    \label{eqn:gru_2}
    &\bz_t = \sigma(\bW_3\phi(\bG_t) + b_3 + \bW_4\bh_{t-1} + b_4),  \\
    \label{eqn:gru_3}
    &\bq_t = \tanh(\bW_5\phi(\bG_t) + b_5 + \br_t * (\bW_6\bh_{t-1}+b_6)), \\
    \label{eqn:gru_4}
    &\bh_{t} = (1-\bz_t)*\bq_t + \bz_t * \bh_{t-1},
\end{align}
\end{subequations}
where $\bW$ and $b$ are learnable parameters; $*$ is Hadamard product; $\bh_t\in\mathbb{R}^{H\times 1}$ is a hidden state, where $\bh_0$ is initialized with $\mathbf{0}$, and $H$ is the hidden size.

GRU is one of the self-gated recurrent neural networks, and it is designed to avoid vanishing gradient problems, especially when dealing with long-term series.
It contains a reset gate $\br_t$ and an update gate $\bz_t$, and they are separately generated from the combination of the current input $\phi(\bG_t)$ and previous history $\bh_{t-1}$, as Equation (\ref{eqn:gru_1}) and Equation (\ref{eqn:gru_2}).
Then, in Equation (\ref{eqn:gru_3}), reset gate $\br_t$ can decide how much previous history information from $\bh_{t-1}$ will be forgotten.
When $\br_t$ is close to $\mathbf{1}$, nearly all historical information is kept.
Otherwise, when $\br_t$ is close to $\mathbf{0}$, almost all history is forgotten. 
Similarly, in Equation (\ref{eqn:gru_4}), the update gate can decide how much history from $\bh_{t-1}$ can be passed to the output $\bh_{t}$.

The above structure has the following advantages.
\begin{itemize}
    \item GRU-based memory can generate a compact tensor from $\{\phi(\bG_t)|t\in\mathcal{T}\}$, remaining useful information but forgetting irrelevant information and noises. It makes the whole model deal with long-term RMTS.
    \item Compared with other self-gated methods (LSTM), GRU only leverages one memory state ($\bh_t$), while LSTM leverages two memory states ($\bh_t$ and $\bc_t$). Hence, GRU needs fewer computation resources and is more effective. 
\end{itemize}

\subsection{Training without FL}
\label{sec:train}

\begin{algorithm}[tb]
	\caption{Training without FL}\label{Alg:training_no_fl}
	\textbf{Input}: Training datasets $\{\mathcal{D}_{Training}^p|p\in\mathcal{P}\}$ and validation datasets $\{\mathcal{D}_{Validation}^p|p\in\mathcal{P}\}$ of all participants $\mathcal{P}$, initialized parameters $\theta^{global}_0$ \;
	\BlankLine
	 
	Build a global function ${\rm GAM}(\bX; \theta_0^{global})$ based on Equation (\ref{Eqn:model})\;
	\BlankLine
	\# Step 1: global traning
	
	$\mathcal{D}_{Training} = \mathcal{D}_{Training}^1\cup...\cup\mathcal{D}_{Training}^p\cup...\cup\mathcal{D}_{Training}^P$ \;
	$\tau \leftarrow 0$\;
	\
	\While{$\tau < T^{global}$}{ 

	    Randomly select a batch of samples from mixed training data $\mathcal{D}_{Training}$\;
	    Use Equation (\ref{Eqn:loss}) to calculate loss with a batch of samples\;
	    Use Adam optimizer to update $\theta_{\tau}^{global}$ as $\theta_{\tau+1}^{global}$\;
	    $\tau \leftarrow \tau+1$\;
	    \If{$\tau \% T^{eval1} == 0$}{
	        \For{ $p$ in $\mathcal{P}$}{
    	        Use validation data $\mathcal{D}_{Validation}^p$ to evaluate ${\rm GAM}(\bX; \theta_{\tau}^{global})$\;
	        }
            \If{performance improved in the validation results}{
                $\theta^{global\_best}\gets \theta_{\tau}^{global}$ \;
            }
	    }
	}
	
	\BlankLine
	\# Step 2: personalized fine tuning
	
    Turn down learning rate\;
	\For{$p$ in $\mathcal{P}$ }{
	    Build a personal model ${\rm GAM}(\bX; \theta_0^{p})$ based on Equation (\ref{Eqn:model}) \;
	    $\theta_0^{p}\gets\theta_{\tau_1}^{global\_best}$\;
	    $\tau \leftarrow 0$\;
    	\While{$\tau<T^{person}$}{ 
    	    Randomly select a batch of samples from the personal training dataset $\mathcal{D}_{Training}^p$\;
	        Use Equation (\ref{Eqn:loss}) to calculate loss with a batch of samples\;
	        Use Adam optimizer to update $\theta_{\tau}^{p}$ as $\theta_{\tau+1}^{p}$\;
	        $\tau \leftarrow \tau+1$\;
	        \If{$\tau \% T^{eval2} == 0$}{
    	        \For{ $p$ in $\mathcal{P}$}{
    	            Use validation data $\mathcal{D}_{Validation}^p$ to evaluate ${\rm GAM}(\bX; \theta_{\tau}^{p})$\;
    	        }
    
                \If{performance improved in the validation results}{
                    $\theta^{best, p}\gets \theta_{\tau}^{p}$ \;
                }
            }
    	}
    	Save $\theta^{best, p}$ \;
	}
	
\end{algorithm}

A two-step training algorithm (Algorithm \ref{Alg:training_no_fl}) is introduced from \cite{DBLP:journals/jhir/ZhuLCHG20} since it can not only leverage population patterns across different personal data but also keep personal characteristics via individual fine-tuning based on personal data.

Specifically, in Algorithm \ref{Alg:training_no_fl}, the input of the algorithm contains training datasets $\{\mathcal{D}_{Training}^p|p\in\mathcal{P}\}$ and validation datasets $\{\mathcal{D}_{Validation}^p|p\in\mathcal{P}\}$ of all participants $\mathcal{P}$ and initialized parameters $\theta^{global}_0$ (line 1).
Firstly, according to Equation (\ref{Eqn:model}), a population model ${\rm GAM}(\bX; \theta_0^{global})$ is built with $\theta_0^{global}$ (line 2).
The first step of this algorithm is global training.
It mixes the personal training data from all participants (line 4).
Then, initialize the training epoch $\tau$ with $0$ (line 5), and the global training is started (line 6).
In the training loop, a batch of samples is randomly selected from the mixing training dataset $\mathcal{D}_{Training}$ (line 7).
Equation (\ref{Eqn:loss}) is utilized to calculate gradients, and learnable weights $\theta_{\tau}^{global}$ are updated by the Adam optimizer (lines 8-10).
In certain training epochs (line 11), GAM is evaluated by the validation datasets $\{\mathcal{D}^p_{Validation}|p\in\mathcal{P}\}$ (lines 12-13).
Based on the validation results, if the performance is improved, the best learnable global parameters are saved as $\theta^{global\_best}$ (lines 14-15).

The second training step, personalized fine-tuning, begins with turning down the learning rate (line 17), followed by the loop of participants (line 18).
For each participant $p$, a personalized model is built based on Equation (\ref{Eqn:model}), and it is initialized by the best global parameters $\theta^{global\_best}$ (lines 19-20).
Then, the training epoch $\tau$ is initialized with $0$ (line 21).
In the loop of fine-tuning for the personalized model (line 22), a batch of samples are ramdomly selected from the personal training dataset $\mathcal{D}_{Training}^{p}$ (line 23).
Equation (\ref{Eqn:loss}) is also utilized to calculate gradients, and learnable weights $\theta_{\tau}^{global}$ are updated by the Adam optimizer (lines 24-26).
GAM is evaluated by the validation datasets $\{\mathcal{D}^p_{Validation}|p\in\mathcal{P}\}$ in certain training epochs (lines 27-29).
Similarly, if the performance is improved, the best personal learnable personal parameters are saved as $\theta^{best, p}$ (lines 30-32).

\subsection{Training with FL}
\label{sec:train_fl}

\begin{algorithm}[tb]
	\caption{Training with FL}\label{Alg:training_with_fl}
	\textbf{Input}: Training datasets $\{\mathcal{D}_{Training}^p|p\in\mathcal{P}\}$ and validation datasets $\{\mathcal{D}_{Validation}^p|p\in\mathcal{P}\}$ of all participants $\mathcal{P}$, initialized parameters $\theta^{global}_0$ \;
	\BlankLine
	 
	\BlankLine
	\# Step 1: federated learning

	$\tau_1 \leftarrow 0$\;
	\While{$\tau_1<T^{total}$}{
	    \For{$p$ in $\mathcal{P}$}{ 
	        \# each participant is in a client
	        
	        $\tau_2 \leftarrow 0$\;
            Build a personal model ${\rm GAM}(\bX; \theta_{\tau_2}^{p})$ based on (\ref{Eqn:model}) \;
            
            \# send population parameters from the server to the client
            
    	    $\theta_{\tau_2}^{p}\gets\theta_{\tau_1}^{global}$ \; 
        	\While{$\tau_2<T^{client}$}{ 
        	    Randomly select a batch of samples from the personal training dataset $\mathcal{D}_{Training}^p$\;
    	        Use Equation (\ref{Eqn:loss}) to calculate loss with a batch of samples\;
    	        Use Adam optimizer to update $\theta_{\tau_2}^{p}$ as $\theta_{\tau_2+1}^{p}$\;
    	        $\tau_2 \leftarrow \tau_2+1$\;
        	}
        }
        
        \# collect personal parameters from all clients
        
        $\theta_{\tau_1 + 1}^{global} \gets \frac{1}{P}\sum_p\theta_{\tau_2}^{p}$\;
        $\tau_1 \leftarrow \tau_1+1$\;
        \If{$\tau_1\%T^{eval1}==0$}{
            \For{ $p$ in $\mathcal{P}$}{
                Use validation data $\mathcal{D}_{Validation}^p$ to evaluate ${\rm GAM}(\bX; \theta_{\tau_1}^{global})$\;
            }
            \If{performance improved in the validation results}{
                $\theta^{best\_global}\gets \theta_{\tau_1}^{global}$ \;
            }
        }

	}

	
	\BlankLine
	\# Step 2: personalized fine tuning
	
    Turn down learning rate\;
	\For{$p$ in $\mathcal{P}$ }{
	    Build a personal model ${\rm GAM}(\bX; \theta_0^{p})$ based on (\ref{Eqn:model}) \;
	    $\theta_0^{p}\gets\theta^{global\_best}$\;
	    $\tau \leftarrow 0$\;
    	\While{$\tau<T^{person}$}{ 
    	    Randomly select a batch of samples from the personal training dataset $\mathcal{D}_{Training}^p$\;
	        Use Equation (\ref{Eqn:loss}) to calculate loss with a batch of samples\;
	        Use Adam optimizer to update $\theta_{\tau}^{p}$ as $\theta_{\tau+1}^{p}$\;
	        $\tau \leftarrow \tau+1$\;
	        \If{$\tau\%T^{eval2}==0$}{
    	        \For{ $p$ in $\mathcal{P}$}{
    	            Use validation data $\mathcal{D}_{Validation}^p$ to evaluate ${\rm GAM}(\bX; \theta_{\tau}^{p})$\;
    	        }
    
                \If{performance improved in the validation results}{
                    $\theta^{best, p}\gets \theta_{\tau}^{p}$ \;
                }
            }
    	} 
    	Save $\theta^{best, p}$ \;
	}
	
\end{algorithm}

As we mentioned in the previous subsection, two-step training can currently keep population patterns and personal characteristics.
Nevertheless, it leaks participant privacy by mixing personal data.
Hence, if it is possible to sacrifice some performance, privacy may be well protected by introducing FL \citep{DBLP:conf/aistats/McMahanMRHA17}.
In FL, personal data is only stored in the place where it is generated rather than being sent to a center and mixed with other data.

The details of FL are in Algorithm \ref{Alg:training_with_fl}.
The input of the algorithm includes training datasets $\{\mathcal{D}_{Training}^p|p\in\mathcal{P}\}$ and validation datasets $\{\mathcal{D}_{Validation}^p|p\in\mathcal{P}\}$ of all participants $\mathcal{P}$ and initialized parameters $\theta^{global}_0$ (line 1).
There are $T^{total}$ of training epochs for FL (line 4).
In each epoch $\tau_1$, each client has a participant $p$, and a personalized model is built based on Equation (\ref{Eqn:model}) (lines 5-8).
The server sends the population parameters $\theta^{global}_{\tau_1}$ to the client, and the personalized model is initialized by the population parameters (lines 9-10).
Then, the client trains the personalized model by batch learning using Equation (\ref{Eqn:loss}) and Adam optimizer for $T^{client}$ epochs (lines 11-15).
When all clients finish the personalized training, the server collects parameters from all clients and generates population parameters $\theta_{\tau_1}^{global}$ by averaging (lines 16-18).
Next, in certain epochs (line 19), the personalized model is evaluated by personal validation data in each client (lines 20-21).
If the total performance is improved, the current population parameters will be saved as the best population parameters $\theta^{best\_global}$ (lines 22-23). 

The second step of Algorithm \ref{Alg:training_with_fl}, personalized fine-tuning, is the same as the one in Algorithm \ref{Alg:training_no_fl}.
The only difference between Algorithm \ref{Alg:training_no_fl} and Algorithm \ref{Alg:training_with_fl} is in the first training step since Algorithm \ref{Alg:training_no_fl} mixes all training data to generate a population model, while Algorithm \ref{Alg:training_with_fl} utilizes FL to generate a population model, keeping the personal data in privacy.

\subsection{Metrics}
\label{sec:metrics}
Root mean square error (RMSE) is introduced to evaluate the performance of models comprehensively, and it is utilized to select the best learnable parameters and the best combination of attributes.
Mean absolute relative difference (MARD) is sensitive to small values, so it is used to evaluate the prediction of small values.
Besides, mean absolute error (MAE) is also introduced to conveniently compare the proposed model with other methods, as all methods in Ohio 2020 challenge \citep{DBLP:conf/ecai/2020kdh} use MAE as one of the metrics.
The above metrics are defined as follows:
\begin{equation}
    RMSE = \sqrt{\frac{1}{I} \sum_i({y}_{T+W, i} - \hat{y}_{T+W, i})^2};
\end{equation}
\begin{equation}
    MARD = \frac{1}{I} \sum_i\frac{|{y}_{T+W, i} - \hat{y}_{T+W, i}|}{{y}_{T+W, i}} \times 100\%;
\end{equation}
\begin{equation}
    MAE = \frac{1}{I} \sum_i |{y}_{T+W, i} - \hat{y}_{T+W, i}|;
\end{equation}
where $I$ is the number of test samples of a participant.
In the selection of best parameters during validation, only RMSE is referred to, as it can be seen as overall performance.

\chapter{Results}
\label{chap:exp_res}

In this chapter, we discuss the experiment settings and result analysis of the impact of attributes (see section \ref{sec:imp_attri}), the impact of hyperparameters (see section \ref{sec:imp_hyper}), the impact of FL (see section \ref{sec:imp_fl}) and the impact of time-aware attention (see section \ref{sec:imp_ta}).
Then, we answer the research questions raised in the research proposal in section \ref{sec:ans} according to the observations and conclusions of these experiments.

\section{Impact of Attributes}
\label{sec:imp_attri}

\subsection{Experiment Settings}
\label{sec:attri_setting}

In order to well exploit different attributes in OhioT1DM'20 and OhioT1DM'18, considering the experimental consumption, we tried limited combinations of attributes.
Surprisingly, we still find that using 6 attributes, i.e., ``glucose\_level'', ``meal'', ``bolus'', ``finger\_stick'', ``sleep'' and ``exercise'' are pretty helpful and bring about explicit improvements in most cases.

Hence, the first experiment is to show whether each of these 6 attributes positively impacts BGLP-RMTS.
In the experiment, each time, we remove one attribute, excluding ``glucose\_level'', from the 6 attributes.
We also run GAM with all these 6 attributes and GRU with only ``glucose\_level'' for comparisons.
If GAM with all these 6 attributes performs the best, we can come to a conclusion that each attributes can positively affect BGLP-RMTS. 
Meanwhile, if GRU with only ``glucose\_level'' performs the worst, we can draw a conclusion that the combination of these 6 attributes is successful.
The relevant results are in Tables \ref{table:6_rmse_attri_wo}-\ref{table:12_mae_attri_wo}.

On the other hand, we also want to show why we did not consider other attributes.
Therefore, we incorporate one more attribute at a time when given the 6 attributes.
We also use the GAM with all 6 attributes for comparisons. 
If the new considered attribute cannot surpass the GAM with all the 6 attributes in most cases, we will not take the new attribute into account.
The relevant results are in Tables \ref{table:6_rmse_attri}-\ref{table:12_mae_attri}.

\textbf{Hyperparameter settings}: prediction window is $W=6$ or $W=12$; historical sequence length $T=12$; GRU hidden size is $H=256$; no. of heads $M=1$; no. of GAT layers $L=1$. The training algorithm is based on Algorithm \ref{Alg:training_no_fl}, and no. of global training epochs in the first or second training step is $T^{global}=10,000$ or $T^{person}=800$; evaluation interval in the first or second training step is $T^{eval1}=1,000$ or $T^{eval2}=160$; batch size is $B=128$; learning rate in the first or second training step is $0.001$ or $0.00001$.

\subsection{Result Analysis}
\label{sec:attri_res}

\begin{table}[tb]
\centering
\caption{In terms of RMSE (mg/dL), the impact of selecting attributes ($W=6$), where 6 attributes are ``glucose\_level'', ``meal'', ``bolus'', ``finger\_stick'', ``sleep'' and ``exercise''.}
\resizebox{1.05\textwidth}{!}{
    \begin{tabular}{lccccccccccccc}
    \hline
    \multicolumn{1}{c}{Method}         & 559             & 563             & 570             & 575             & 588             & 591             & 540             & 544             & 552             & 567             & 584             & 596             & Average         \\ \hline
    GAM 6 attributes w/o meal          & 19.505          & 17.842          & 16.940          & 22.521          & 17.315          & 21.044          & 21.201          & 16.255          & 16.264          & 21.570          & 22.226          & 16.821          & 19.125          \\
    GAM 6 attributes w/o bolus         & 19.737          & 17.996          & 16.737          & 22.397          & 17.084          & 20.886          & 21.703          & \textbf{15.620} & 16.703          & 21.981          & 23.133          & 16.403          & 19.198          \\
    GAM 6 attributes w/o finger\_stick & 19.951          & 17.824          & 16.996          & 22.461          & 16.943          & 21.457          & 20.988          & 16.221          & 16.405          & 21.978          & 22.626          & 16.097          & 19.162          \\
    GAM 6 attributes w/o sleep         & \textbf{19.009} & 17.905          & \textbf{16.119} & 22.501          & 17.172          & 20.826          & 21.061          & 16.124          & 16.468          & \textbf{21.523} & 22.282          & 16.273          & 18.939          \\
    GAM 6 attributes w/o exercise      & 19.106          & 17.977          & 16.988          & 22.418          & 17.302          & 21.002          & \textbf{20.886} & 16.269          & \textbf{16.186} & 21.544          & \textbf{21.997} & 16.017          & 18.974          \\
    GAM 6 attributes                   & 19.415          & \textbf{17.596} & 16.662          & \textbf{22.279} & \textbf{17.057} & \textbf{20.717} & 21.008          & 15.954          & 16.502          & 21.582          & 22.220          & \textbf{16.012} & \textbf{18.917} \\
    GRU glucose\_level                 & 19.439          & 18.291          & 18.604          & 23.620          & 18.485          & 21.832          & 22.771          & 18.002          & 17.367          & 22.919          & 23.587          & 17.203          & 20.177          \\ \hline
    \end{tabular}
}
\label{table:6_rmse_attri_wo}
\end{table}

\begin{table}[tb]
\centering
\caption{In terms of MARD (\%), the impact of selecting attributes ($W=6$), where 6 attributes are ``glucose\_level'', ``meal'', ``bolus'', ``finger\_stick'', ``sleep'' and ``exercise''.}
\resizebox{1.05\textwidth}{!}{
    \begin{tabular}{lccccccccccccc}
    \hline
    \multicolumn{1}{c}{Method}         & 559            & 563            & 570            & 575            & 588            & 591             & 540             & 544            & 552            & 567             & 584             & 596            & Average        \\ \hline
    GAM 6 attributes w/o meal          & 8.828          & 7.753          & 5.375          & 9.738          & 7.548          & 12.173          & 10.867          & 7.740          & 9.198          & 10.911          & 10.913          & 8.778          & 9.152          \\
    GAM 6 attributes w/o bolus         & 8.889          & 7.754          & 5.374          & 9.554          & 7.493          & 12.033          & 11.091          & \textbf{7.489} & 9.358          & 11.108          & 11.470          & 8.592          & 9.184          \\
    GAM 6 attributes w/o finger\_stick & 8.965          & 7.725          & 5.382          & 9.495          & \textbf{7.365} & 12.253          & 10.883          & 7.715          & 9.214          & 11.136          & 11.198          & 8.545          & 9.156          \\
    GAM 6 attributes w/o sleep         & 8.711          & 7.734          & \textbf{5.205} & 9.598          & 7.445          & \textbf{11.845} & 10.778          & 7.620          & 9.235          & \textbf{10.646} & 10.932          & 8.481          & \textbf{9.019} \\
    GAM 6 attributes w/o exercise      & \textbf{8.615} & 7.708          & 5.364          & 9.483          & 7.500          & 11.939          & \textbf{10.748} & 7.750          & \textbf{9.063} & 10.850          & \textbf{10.791} & 8.491          & 9.025          \\
    GAM 6 attributes                   & 8.790          & \textbf{7.648} & 5.259          & \textbf{9.443} & 7.416          & 11.965          & 10.779          & 7.606          & 9.146          & 10.866          & 11.029          & \textbf{8.430} & 9.031          \\
    GRU glucose\_level                 & 8.992          & 7.978          & 5.742          & 10.268         & 8.028          & 12.352          & 11.583          & 8.688          & 9.662          & 11.235          & 11.255          & 8.984          & 9.564          \\ \hline
    \end{tabular}
}
\label{table:6_mard_attri_wo}
\end{table}

\begin{table}[tb]
\centering
\caption{In terms of MAE (mg/dL), the impact of selecting attributes ($W=6$), where 6 attributes are ``glucose\_level'', ``meal'', ``bolus'', ``finger\_stick'', ``sleep'' and ``exercise''.}
\resizebox{1.05\textwidth}{!}{
    \begin{tabular}{lccccccccccccc}
    \hline
    \multicolumn{1}{c}{Method}         & 559             & 563             & 570             & 575             & 588             & 591             & 540             & 544             & 552             & 567             & 584             & 596             & Average         \\ \hline
    GAM 6 attributes w/o meal          & 13.226          & 12.445          & 10.891          & 14.236          & 12.667          & 15.108          & 15.558          & 11.676          & 11.945          & 14.664          & 15.955          & 11.828          & 13.350          \\
    GAM 6 attributes w/o bolus         & 13.335          & 12.490          & 10.818          & 14.055          & 12.522          & 14.885          & 15.922          & \textbf{11.204} & 12.267          & 14.945          & 16.677          & 11.539          & 13.388          \\
    GAM 6 attributes w/o finger\_stick & 13.340          & 12.422          & 10.942          & 13.962          & \textbf{12.387} & 15.197          & 15.544          & 11.636          & 12.101          & 14.977          & 16.268          & 11.404          & 13.348          \\
    GAM 6 attributes w/o sleep         & 12.897          & 12.475          & \textbf{10.503} & 13.968          & 12.535          & \textbf{14.812} & 15.430          & 11.497          & 12.052          & \textbf{14.537} & 15.859          & 11.390          & 13.163          \\
    GAM 6 attributes w/o exercise      & \textbf{12.683} & 12.419          & 10.967          & 13.924          & 12.630          & 14.937          & \textbf{15.370} & 11.683          & \textbf{11.871} & 14.676          & \textbf{15.817} & 11.335          & 13.192          \\
    GAM 6 attributes                   & 12.980          & \textbf{12.268} & 10.659          & \textbf{13.852} & 12.471          & 14.872          & 15.389          & 11.489          & 12.017          & 14.628          & 16.005          & \textbf{11.271} & \textbf{13.158} \\
    GRU glucose\_level                 & 13.292          & 12.784          & 11.660          & 14.927          & 13.365          & 15.454          & 16.702          & 12.909          & 12.683          & 15.361          & 16.519          & 12.083          & 13.978          \\ \hline
    \end{tabular}
}
\label{table:6_mae_attri_wo}
\end{table}

\begin{table}[tb]
\centering
\caption{In terms of RMSE (mg/dL), the impact of selecting attributes ($W=12$), where 6 attributes are ``glucose\_level'', ``meal'', ``bolus'', ``finger\_stick'', ``sleep'' and ``exercise''.}
\resizebox{1.05\textwidth}{!}{
    \begin{tabular}{lccccccccccccc}
    \hline
    \multicolumn{1}{c}{Method}           & 559             & 563             & 570             & 575             & 588             & 591             & 540             & 544             & 552             & 567             & 584             & 596             & Average         \\ \hline
    GAM 6 attributes w/o meal            & 32.818          & 28.968          & 28.834          & 35.662          & 29.971          & \textbf{32.453} & 38.750          & 27.333          & 28.918          & 36.931          & 36.309          & 27.158          & 32.009          \\
    GAM 6 attributes w/o bolus           & 32.478          & \textbf{28.876} & 27.626          & \textbf{33.702} & 29.275          & 33.206          & 39.380          & 26.211          & 29.623          & 37.782          & 38.082          & 26.364          & 31.884          \\
    GAM 6 attributes w/o   finger\_stick & 31.454          & 29.769          & 28.630          & 35.333          & \textbf{29.070} & 33.104          & 38.265          & 26.619          & 28.865          & 36.954          & 36.287          & \textbf{25.949} & 31.691          \\
    GAM 6 attributes w/o sleep           & 32.155          & 29.707          & 27.831          & 35.126          & 29.509          & 33.032          & 38.540          & 27.721          & \textbf{28.497} & \textbf{36.589} & \textbf{36.027} & 27.293          & 31.835          \\
    GAM 6 attributes w/o exercise        & \textbf{31.241} & 29.552          & 28.289          & 34.993          & 29.148          & 33.043          & \textbf{38.261} & \textbf{26.027} & 28.762          & 37.081          & 36.307          & 26.014          & 31.560          \\
    GAM 6 attributes                     & 31.282          & 29.641          & \textbf{27.024} & 35.095          & 29.666          & 32.885          & 38.445          & 26.348          & 28.650          & 36.861          & 36.267          & 26.314          & \textbf{31.540} \\
    GRU glucose\_level                   & 32.819          & 29.892          & 30.052          & 35.516          & 30.792          & 33.810          & 39.894          & 31.045          & 29.569          & 38.032          & 38.267          & 28.524          & 33.184          \\ \hline
    \end{tabular}
}
\label{table:12_rmse_attri_wo}
\end{table}

\begin{table}[tb]
\centering
\caption{In terms of MARD (\%), the impact of selecting attributes ($W=12$), where 6 attributes are ``glucose\_level'', ``meal'', ``bolus'', ``finger\_stick'', ``sleep'' and ``exercise''.}
\resizebox{1.05\textwidth}{!}{
    \begin{tabular}{lccccccccccccc}
    \hline
    \multicolumn{1}{c}{Method}           & 559             & 563             & 570            & 575             & 588             & 591             & 540             & 544             & 552             & 567             & 584             & 596             & Average         \\ \hline
    GAM 6 attributes w/o meal            & 15.767          & 12.758          & 9.846          & 17.414          & 13.015          & 20.804          & 20.226          & 13.506          & 17.553          & 21.280          & 18.911          & 14.759          & 16.320          \\
    GAM 6 attributes w/o bolus           & 16.099          & \textbf{12.646} & 9.692          & \textbf{16.467} & 12.593          & 21.117          & 20.477          & 12.659          & 17.255          & 21.545          & 19.636          & 14.240          & 16.202          \\
    GAM 6 attributes w/o   finger\_stick & 15.609          & 12.996          & 9.821          & 16.968          & \textbf{12.582} & 21.123          & \textbf{19.879} & 12.741          & \textbf{16.876} & 21.145          & 19.007          & \textbf{14.181} & 16.077          \\
    GAM 6 attributes w/o sleep           & 16.203          & 12.887          & 9.744          & 17.351          & 12.738          & \textbf{20.566} & 19.963          & 13.249          & 17.270          & \textbf{20.263} & \textbf{18.502} & 14.725          & 16.122          \\
    GAM 6 attributes w/o exercise        & 15.588          & 12.790          & 9.479          & 16.965          & 12.770          & 20.952          & 20.006          & \textbf{12.565} & 17.140          & 21.074          & 19.228          & 14.356          & 16.076          \\
    GAM 6 attributes                     & \textbf{15.517} & 12.851          & \textbf{9.421} & 16.857          & 12.762          & 20.721          & 20.003          & 12.655          & 17.173          & 20.955          & 19.039          & 14.262          & \textbf{16.018} \\
    GRU glucose\_level                   & 16.836          & 13.282          & 10.730         & 17.990          & 13.275          & 21.225          & 20.801          & 16.462          & 17.874          & 21.111          & 19.700          & 15.683          & 17.081          \\ \hline
    \end{tabular}
}
\label{table:12_mard_attri_wo}
\end{table}

\begin{table}[tb]
\centering
\caption{In terms of MAE (mg/dL), the impact of selecting attributes ($W=12$), where 6 attributes are ``glucose\_level'', ``meal'', ``bolus'', ``finger\_stick'', ``sleep'' and ``exercise''.}
\resizebox{1.05\textwidth}{!}{
    \begin{tabular}{lccccccccccccc}
    \hline
    \multicolumn{1}{c}{Method}           & 559             & 563             & 570             & 575             & 588             & 591             & 540             & 544             & 552             & 567             & 584             & 596             & Average         \\ \hline
    GAM 6 attributes w/o meal            & 23.367          & \textbf{20.481} & 19.685          & 25.094          & 22.017          & 25.115          & 29.372          & 20.064          & 22.257          & 27.719          & 27.541          & 19.870          & 23.549          \\
    GAM 6 attributes w/o bolus           & 23.514          & 20.601          & 19.284          & \textbf{23.857} & 21.336          & 25.233          & 29.869          & \textbf{18.965} & 22.345          & 28.000          & 28.710          & 19.049          & 23.397          \\
    GAM 6 attributes w/o   finger\_stick & \textbf{22.570} & 20.965          & 19.727          & 24.478          & \textbf{21.289} & 25.467          & \textbf{28.949} & 19.337          & 21.963          & 27.538          & 27.572          & \textbf{18.803} & 23.222          \\
    GAM 6 attributes w/o sleep           & 23.150          & 21.038          & 19.433          & 24.550          & 21.593          & \textbf{25.085} & 29.182          & 20.024          & 22.038          & \textbf{26.942} & \textbf{26.947} & 19.632          & 23.301          \\
    GAM 6 attributes w/o exercise        & 22.745          & 20.820          & 19.531          & 24.259          & 21.515          & 25.353          & 29.016          & 19.025          & 22.024          & 27.667          & 27.719          & 18.895          & 23.214          \\
    GAM 6 attributes                     & 22.653          & 20.901          & \textbf{19.028} & 24.383          & 21.678          & 25.094          & 29.114          & 19.300          & \textbf{21.960} & 27.540          & 27.488          & 19.235          & \textbf{23.198} \\
    GRU glucose\_level                   & 24.299          & 21.713          & 20.978          & 25.520          & 22.206          & 25.652          & 30.234          & 23.903          & 22.771          & 28.026          & 28.483          & 20.929          & 24.560          \\ \hline
    \end{tabular}
}
\label{table:12_mae_attri_wo}
\end{table}

From Tables \ref{table:6_rmse_attri_wo}-\ref{table:12_mae_attri_wo}, we have the following observations and conclusions.

Overall, when prediction window $W=6$ or $W=12$, in terms of RMSE, MARD and MAE, ``GAM 6 attributes'' performs the best in most cases.
For example, when $W=6$, the average RMSE of ``GAM 6 attributes'' is 18.917 mg/dL, while RMSE of others are higher than this value (see Table \ref{table:6_rmse_attri_wo}).
Only when $W=6$, in terms of MARD, ``GAM 6 attributes'' is not the best.
In this case, the average MARD of ``GAM 6 attributes'' is 9.031\%, while the average MARD of ``GAM 6 attributes w/o sleep'' is $9.019$\% (see Table \ref{table:6_mard_attri_wo}). 
We can also see that ``GRU glucose\_level'' consistently performs the worst in all cases.
For example, when $W=6$, the average RMSE of ``GRU glucose\_level'' is 20.177 mg/dL, while the average RMSE of ``GAM 6 attributes'' is only 18.917 mg/dL (see Table \ref{table:6_rmse_attri_wo}).
Then, we can draw a conclusion that each one of the 6 attributes are helpful to the BGLP-RMTS, and the combination and utilization of these 6 attributes lead to excellent improvements.

On the other hand, in terms of all these metrics, we can observe that when removing ``meal'', ``insulin'' or ``finger\_stick'', the performance of GAM decreases more compared with the one removing ``sleep'' or ``exercise''.
Therefore, ``meal'', ``insulin'' and ``finger\_stick'' are more critical compared with ``sleep'' and ``exercise'' in  BGLP-RMTS.
Hence, when this approach is about to be used in a scenario with limited computational resources, ``sleep'' and ``exercise'' can be neglected if the prediction accuracy is relatively not very important.  

Besides, we can also find some general observations in these tables, which always appears in the rest of the experiments in this paper, so we only mention these observations once in this part.
When the prediction window is larger, i.e., $W=12$, the BGLP-RMTS gets harder, compared with the prediction window $W=6$, since all metrics are much larger when $W=12$.
For example, in Tables \ref{table:6_rmse_attri_wo} and \ref{table:12_rmse_attri_wo}, the average RMSE is consistently less than 21 mg/dL in the previous table, while the average RMSE in the latter table are always higher than 30 mg/dL.  
The difficulties of BGLP-RMTS are various when facing different participants.
For example, in Table \ref{table:6_rmse_attri_wo}, the RMSE of participant 540 is much larger than the RMSE of participant 544 in all different methods, so modeling patterns for participant 540 is harder compared with participant 544.
This observation can further show that using a personalized model is quite essential, since the personal data varies a lot across different participants.

\begin{table}[tb]
\centering
\caption{In terms of RMSE (mg/dL), the impact of selecting attributes ($W=6$), where 6 attributes are ``glucose\_level'', ``meal'', ``bolus'', ``finger\_stick'', ``sleep'' and ``exercise''.}
\resizebox{1.05\textwidth}{!}{
    \begin{tabular}{lccccccccccccc}
    \hline
    \multicolumn{1}{c}{Method}                    & 559             & 563             & 570             & 575             & 588             & 591             & 540             & 544             & 552             & 567             & 584             & 596             & Average         \\ \hline
    GAM 6 attributes + basal                      & 19.232          & 19.756          & 19.415          & 22.340          & 17.556          & 20.919          & \textbf{20.685} & 16.058          & 16.857          & 21.839          & 25.066          & 16.006          & 19.644          \\
    GAM 6 attributes +   basis\_skin\_temperature & 19.270          & 17.948          & \textbf{16.214} & 22.706          & 16.992          & 21.125          & 20.960          & 16.420          & 16.493          & 21.742          & 22.607          & 16.013          & 19.041          \\
    GAM 6 attributes + basis\_gsr                 & 19.397          & 18.027          & 16.599          & 22.448          & 17.089          & 21.224          & 20.774          & 16.339          & 16.634          & 21.717          & 22.676          & 16.199          & 19.094          \\
    GAM 6 attributes +   basis\_sleep             & 19.061          & 17.779          & 16.538          & \textbf{22.074} & 17.226          & 20.825          & 21.275          & 15.821          & 16.436          & \textbf{21.448} & 21.997          & 15.930          & \textbf{18.867} \\
    GAM 6 attributes +   acceleration             & 19.307          & 17.933          & 16.419          & 22.242          & 17.207          & 20.709          & 21.080          & 15.835          & 16.685          & 21.816          & 22.154          & \textbf{15.894} & 18.940          \\
    GAM 6 attributes + work                       & 19.389          & 17.780          & 16.797          & 22.099          & 16.944          & 20.825          & 21.102          & 16.061          & 16.536          & 21.496          & \textbf{21.754} & 15.910          & 18.891          \\
    GAM 6 attributes +   basis\_steps             & 19.048          & 17.678          & 16.432          & 22.382          & 17.262          & 20.769          & 21.166          & 15.853          & 16.358          & 21.715          & 22.181          & 16.036          & 18.907          \\
    GAM 6 attributes +   basis\_heart\_rate       & 19.669          & \textbf{17.436} & 16.708          & 22.573          & 16.966          & \textbf{20.616} & 21.132          & 16.104          & 16.524          & 21.651          & 22.044          & 16.045          & 18.956          \\
    GAM 6 attributes +   basis\_air\_temperature  & 19.306          & 17.675          & 16.227          & 22.319          & 17.181          & 20.722          & 21.028          & 15.975          & \textbf{16.156} & 21.660          & 22.062          & 16.133          & 18.870          \\
    GAM 6 attributes +   day\_of\_week            & \textbf{18.851} & 17.677          & 16.743          & 22.692          & 16.981          & 20.725          & 21.079          & \textbf{15.822} & 16.229          & 21.904          & 22.486          & 16.315          & 18.959          \\
    GAM 6 attributes +   total\_seconds           & 19.238          & 17.871          & 16.322          & 22.328          & \textbf{16.900} & 21.174          & 20.899          & 16.074          & 16.526          & 21.881          & 22.498          & 16.091          & 18.984          \\
    GAM 6 attributes                              & 19.415          & 17.596          & 16.662          & 22.279          & 17.057          & 20.717          & 21.008          & 15.954          & 16.502          & 21.582          & 22.220          & 16.012          & 18.917          \\ \hline
    \end{tabular}
}
\label{table:6_rmse_attri}
\end{table}

\begin{table}[tb]
\centering
\caption{In terms of MARD (\%), the impact of selecting attributes ($W=6$), where 6 attributes are ``glucose\_level'', ``meal'', ``bolus'', ``finger\_stick'', ``sleep'' and ``exercise''.}
\resizebox{1.05\textwidth}{!}{
    \begin{tabular}{lccccccccccccc}
    \hline
    \multicolumn{1}{c}{Method}                    & 559            & 563            & 570            & 575            & 588            & 591             & 540             & 544            & 552            & 567             & 584             & 596            & Average        \\ \hline
    GAM 6 attributes + basal                      & 8.748          & 8.509          & 7.151          & 9.476          & 7.712          & 12.045          & \textbf{10.699} & 7.638          & 9.399          & 10.993          & 13.944          & \textbf{8.361} & 9.556          \\
    GAM 6 attributes +   basis\_skin\_temperature & 8.670          & 7.681          & 5.287          & 9.813          & 7.382          & 12.050          & 10.835          & 7.764          & 9.377          & 10.737          & 11.099          & 8.472          & 9.097          \\
    GAM 6 attributes + basis\_gsr                 & 8.876          & 7.762          & 5.270          & 9.853          & 7.509          & 12.000          & 10.758          & 7.651          & 9.460          & 10.858          & 11.158          & 8.522          & 9.140          \\
    GAM 6 attributes +   basis\_sleep             & 8.467          & 7.706          & 5.277          & \textbf{9.359} & 7.458          & 11.795          & 10.840          & 7.589          & \textbf{9.110} & \textbf{10.552} & 10.961          & 8.401          & \textbf{8.960} \\
    GAM 6 attributes +   acceleration             & 8.664          & 7.686          & 5.224          & 9.488          & 7.418          & 11.920          & 10.714          & 7.547          & 9.305          & 11.033          & 10.946          & 8.433          & 9.031          \\
    GAM 6 attributes + work                       & 8.725          & 7.706          & 5.311          & 9.470          & \textbf{7.352} & 11.926          & 10.771          & 7.573          & 9.283          & 10.833          & \textbf{10.742} & 8.395          & 9.007          \\
    GAM 6 attributes +   basis\_steps             & 8.644          & 7.629          & 5.266          & 9.492          & 7.443          & 11.853          & 10.768          & \textbf{7.501} & 9.161          & 10.871          & 10.918          & 8.457          & 9.000          \\
    GAM 6 attributes +   basis\_heart\_rate       & 8.775          & 7.666          & 5.298          & 9.680          & 7.418          & \textbf{11.768} & 10.794          & 7.562          & 9.210          & 10.830          & 10.887          & 8.499          & 9.032          \\
    GAM 6 attributes +   basis\_air\_temperature  & 8.611          & 7.678          & 5.419          & 9.561          & 7.431          & 11.865          & 10.733          & 7.565          & 9.123          & 10.907          & 10.834          & 8.535          & 9.022          \\
    GAM 6 attributes +   day\_of\_week            & \textbf{8.591} & 7.689          & 5.323          & 9.538          & 7.370          & 11.962          & 10.833          & 7.529          & 9.139          & 10.994          & 11.170          & 8.526          & 9.055          \\
    GAM 6 attributes +   total\_seconds           & 8.781          & 7.703          & \textbf{5.203} & 9.722          & 7.401          & 12.068          & 10.706          & 7.567          & 9.256          & 10.975          & 11.072          & 8.447          & 9.075          \\
    GAM 6 attributes                              & 8.790          & \textbf{7.648} & 5.259          & 9.443          & 7.416          & 11.965          & 10.779          & 7.606          & 9.146          & 10.866          & 11.029          & 8.430          & 9.031          \\ \hline
    \end{tabular}
}
\label{table:6_mard_attri}
\end{table}

\begin{table}[tb]
\centering
\caption{In terms of MAE (mg/dL), the impact of selecting attributes ($W=6$), where 6 attributes are ``glucose\_level'', ``meal'', ``bolus'', ``finger\_stick'', ``sleep'' and ``exercise''.}
\resizebox{1.05\textwidth}{!}{
    \begin{tabular}{lccccccccccccc}
    \hline
    \multicolumn{1}{c}{Method}                    & 559             & 563             & 570             & 575             & 588             & 591             & 540             & 544             & 552             & 567             & 584             & 596             & Average         \\ \hline
    GAM 6 attributes + basal                      & 12.970          & 13.845          & 13.557          & 13.841          & 12.905          & 14.954          & \textbf{15.266} & 11.497          & 12.282          & 14.833          & 19.037          & 11.280          & 13.856          \\
    GAM 6 attributes +   basis\_skin\_temperature & 12.838          & 12.410          & 10.628          & 14.316          & 12.409          & 15.229          & 15.534          & 11.729          & 12.179          & 14.660          & 16.250          & 11.401          & 13.299          \\
    GAM 6 attributes + basis\_gsr                 & 13.069          & 12.496          & 10.717          & 14.243          & 12.561          & 15.146          & 15.359          & 11.627          & 12.284          & 14.781          & 16.378          & 11.412          & 13.339          \\
    GAM 6 attributes +   basis\_sleep             & \textbf{12.691} & 12.438          & 10.701          & \textbf{13.796} & 12.577          & 14.802          & 15.654          & 11.404          & 11.977          & \textbf{14.343} & 15.848          & 11.231          & \textbf{13.122} \\
    GAM 6 attributes +   acceleration             & 12.853          & 12.401          & 10.599          & 13.874          & 12.565          & 14.855          & 15.436          & 11.358          & 12.142          & 14.868          & 15.957          & 11.274          & 13.182          \\
    GAM 6 attributes + work                       & 13.057          & 12.440          & 10.761          & 13.830          & \textbf{12.377} & 14.806          & 15.481          & 11.437          & 12.142          & 14.574          & \textbf{15.640} & \textbf{11.204} & 13.146          \\
    GAM 6 attributes +   basis\_steps             & 12.911          & 12.330          & 10.598          & 13.906          & 12.548          & 14.766          & 15.494          & \textbf{11.312} & 11.954          & 14.648          & 15.847          & 11.351          & 13.139          \\
    GAM 6 attributes +   basis\_heart\_rate       & 13.368          & 12.289          & 10.754          & 14.185          & 12.481          & \textbf{14.734} & 15.495          & 11.473          & 12.059          & 14.612          & 15.899          & 11.410          & 13.230          \\
    GAM 6 attributes +   basis\_air\_temperature  & 12.976          & 12.399          & 10.879          & 14.035          & 12.560          & 14.836          & 15.447          & 11.418          & \textbf{11.883} & 14.708          & 15.828          & 11.496          & 13.206          \\
    GAM 6 attributes +   day\_of\_week            & 12.705          & 12.369          & 10.691          & 13.996          & 12.403          & 14.901          & 15.460          & 11.337          & 11.925          & 14.766          & 16.111          & 11.476          & 13.178          \\
    GAM 6 attributes +   total\_seconds           & 12.988          & 12.450          & \textbf{10.533} & 14.152          & 12.389          & 15.106          & 15.375          & 11.413          & 12.129          & 14.873          & 16.232          & 11.277          & 13.243          \\
    GAM 6 attributes                              & 12.980          & \textbf{12.268} & 10.659          & 13.852          & 12.471          & 14.872          & 15.389          & 11.489          & 12.017          & 14.628          & 16.005          & 11.271          & 13.158          \\ \hline
    \end{tabular}
}
\label{table:6_mae_attri}
\end{table}

\begin{table}[tb]
\centering
\caption{In terms of RMSE (mg/dL), the impact of selecting attributes ($W=12$), where 6 attributes are ``glucose\_level'', ``meal'', ``bolus'', ``finger\_stick'', ``sleep'' and ``exercise''.}
\resizebox{1.05\textwidth}{!}{
    \begin{tabular}{lccccccccccccc}
    \hline
    \multicolumn{1}{c}{Method}                    & 559             & 563             & 570             & 575             & 588             & 591             & 540             & 544             & 552             & 567             & 584             & 596             & Average         \\ \hline
    GAM 6 attributes + basal                      & 31.749          & 29.598          & 29.113          & \textbf{34.662} & 29.477          & 32.908          & 38.907          & 26.425          & 28.771          & 36.682          & 36.980          & 27.210          & 31.874          \\
    GAM 6 attributes +   basis\_skin\_temperature & \textbf{30.994} & 29.421          & 27.942          & 34.996          & 29.750          & 33.113          & \textbf{38.042} & \textbf{25.926} & 28.865          & 36.262          & 35.862          & 26.007          & 31.432          \\
    GAM 6 attributes + basis\_gsr                 & 31.578          & 29.702          & 27.457          & 35.327          & 29.183          & 33.516          & 38.115          & 26.981          & 29.408          & 36.938          & 35.957          & 26.027          & 31.682          \\
    GAM 6 attributes +   basis\_sleep             & 31.617          & 29.758          & 27.975          & 35.327          & 29.115          & \textbf{32.444} & 39.046          & 26.634          & \textbf{28.505} & 36.661          & 36.199          & 26.028          & 31.609          \\
    GAM 6 attributes +   acceleration             & 31.806          & 30.187          & 28.829          & 35.954          & 29.277          & 32.635          & 38.573          & 26.543          & 29.169          & 37.410          & 35.972          & 25.776          & 31.844          \\
    GAM 6 attributes + work                       & 32.879          & 29.994          & 28.376          & 35.563          & 28.723          & 32.786          & 38.535          & 26.123          & 29.173          & 36.920          & \textbf{35.553} & 26.183          & 31.734          \\
    GAM 6 attributes +   basis\_steps             & 31.791          & 29.614          & 27.865          & 35.730          & 30.153          & 32.948          & 39.049          & 26.413          & 28.934          & 37.020          & 36.332          & 26.192          & 31.837          \\
    GAM 6 attributes +   basis\_heart\_rate       & 31.369          & 29.637          & 27.707          & 35.127          & 30.243          & 32.515          & 38.708          & 26.402          & 28.720          & 37.030          & 35.708          & 26.351          & 31.626          \\
    GAM 6 attributes +   basis\_air\_temperature  & 31.416          & 29.758          & 28.197          & 36.478          & 30.747          & 32.890          & 38.837          & 26.219          & 28.529          & 37.075          & 36.295          & 26.387          & 31.902          \\
    GAM 6 attributes +   day\_of\_week            & 31.819          & 29.536          & 28.148          & 35.315          & 29.253          & 32.843          & 38.727          & 26.291          & 28.676          & 37.437          & 36.418          & 25.834          & 31.692          \\
    GAM 6 attributes +   total\_seconds           & 31.967          & \textbf{29.315} & 27.563          & 35.132          & \textbf{28.677} & 32.827          & 38.088          & 25.997          & 28.867          & \textbf{36.016} & 36.195          & \textbf{25.537} & \textbf{31.348} \\
    GAM 6 attributes                              & 31.282          & 29.641          & \textbf{27.024} & 35.095          & 29.666          & 32.885          & 38.445          & 26.348          & 28.650          & 36.861          & 36.267          & 26.314          & 31.540          \\ \hline
    \end{tabular}
}
\label{table:12_rmse_attri}
\end{table}

\begin{table}[tb]
\centering
\caption{In terms of MARD (\%), the impact of selecting attributes ($W=12$), where 6 attributes are ``glucose\_level'', ``meal'', ``bolus'', ``finger\_stick'', ``sleep'' and ``exercise''.}
\resizebox{1.05\textwidth}{!}{
    \begin{tabular}{lccccccccccccc}
    \hline
    \multicolumn{1}{c}{Method}                    & 559             & 563             & 570            & 575             & 588             & 591             & 540             & 544             & 552             & 567             & 584             & 596             & Average         \\ \hline
    GAM 6 attributes + basal                      & 15.507          & 13.149          & 10.424         & 17.137          & 12.921          & 20.759          & 20.319          & 12.625          & 17.299          & 20.630          & 18.601          & 14.260          & 16.136          \\
    GAM 6 attributes +   basis\_skin\_temperature & \textbf{14.919} & 12.853          & 9.628          & 17.181          & 12.817          & 20.799          & 19.759          & 12.283          & 17.280          & 19.875          & 18.601          & \textbf{13.946} & 15.829          \\
    GAM 6 attributes + basis\_gsr                 & 15.767          & 12.905          & 9.677          & 17.352          & 12.718          & 21.061          & 19.654          & 12.588          & 17.526          & 20.758          & \textbf{18.447} & 14.151          & 16.050          \\
    GAM 6 attributes +   basis\_sleep             & 15.510          & 12.972          & 9.477          & 16.931          & 12.649          & \textbf{20.272} & 20.323          & 12.886          & \textbf{16.846} & 20.356          & 19.120          & 14.208          & 15.962          \\
    GAM 6 attributes +   acceleration             & 15.682          & 12.993          & 9.813          & 17.081          & 12.767          & 20.707          & 19.894          & 12.699          & 17.192          & 21.329          & 18.826          & 14.206          & 16.099          \\
    GAM 6 attributes + work                       & 15.808          & 13.087          & 9.781          & \textbf{16.668} & 12.535          & 20.956          & 20.082          & \textbf{12.268} & 17.204          & 20.995          & 18.479          & 14.249          & 16.009          \\
    GAM 6 attributes +   basis\_steps             & 15.729          & 12.902          & 9.758          & 17.201          & 13.019          & 20.955          & 20.282          & 12.695          & 17.220          & 21.197          & 19.281          & 14.443          & 16.223          \\
    GAM 6 attributes +   basis\_heart\_rate       & 15.695          & 12.988          & 9.466          & 17.271          & 13.243          & 20.486          & 20.027          & 12.654          & 17.230          & 21.242          & 18.677          & 14.404          & 16.115          \\
    GAM 6 attributes +   basis\_air\_temperature  & 15.474          & 12.792          & 9.933          & 17.451          & 13.249          & 20.822          & 20.151          & 12.514          & 17.165          & 21.249          & 19.072          & 14.352          & 16.185          \\
    GAM 6 attributes +   day\_of\_week            & 15.603          & 12.831          & 9.774          & 17.117          & 12.551          & 20.865          & 20.487          & 12.537          & 16.922          & 21.288          & 19.291          & 14.070          & 16.111          \\
    GAM 6 attributes +   total\_seconds           & 15.501          & \textbf{12.774} & \textbf{9.406} & 16.906          & \textbf{12.423} & 21.009          & \textbf{19.169} & 12.361          & 17.045          & \textbf{20.168} & 18.907          & 13.994          & \textbf{15.805} \\
    GAM 6 attributes                              & 15.517          & 12.851          & 9.421          & 16.857          & 12.762          & 20.721          & 20.003          & 12.655          & 17.173          & 20.955          & 19.039          & 14.262          & 16.018          \\ \hline
    \end{tabular}
}
\label{table:12_mard_attri}
\end{table}

\begin{table}[tb]
\centering
\caption{In terms of MAE (mg/dL), the impact of selecting attributes ($W=12$), where 6 attributes are ``glucose\_level'', ``meal'', ``bolus'', ``finger\_stick'', ``sleep'' and ``exercise''.}
\resizebox{1.05\textwidth}{!}{
    \begin{tabular}{lccccccccccccc}
    \hline
    \multicolumn{1}{c}{Method}                    & 559             & 563             & 570             & 575             & 588             & 591             & 540             & 544             & 552             & 567             & 584             & 596             & Average         \\ \hline
    GAM 6 attributes + basal                      & 22.677          & 21.253          & 20.745          & 24.419          & 21.933          & 25.173          & 29.481          & 19.131          & 22.084          & 27.153          & 27.665          & 19.916          & 23.469          \\
    GAM 6 attributes +   basis\_skin\_temperature & \textbf{22.084} & 20.806          & 19.388          & 24.574          & 21.814          & 25.450          & 28.760          & 18.827          & 22.120          & 26.488          & 27.008          & \textbf{18.804} & 23.010          \\
    GAM 6 attributes + basis\_gsr                 & 22.954          & 20.900          & 19.430          & 24.580          & 21.476          & 25.652          & 28.640          & 19.406          & 22.438          & 27.231          & \textbf{26.930} & 18.950          & 23.216          \\
    GAM 6 attributes +   basis\_sleep             & 22.813          & 21.104          & 19.103          & \textbf{24.333} & 21.392          & \textbf{24.780} & 29.487          & 19.447          & \textbf{21.713} & 26.866          & 27.524          & 18.903          & 23.122          \\
    GAM 6 attributes +   acceleration             & 22.989          & 21.166          & 19.923          & 24.653          & 21.518          & 25.049          & 28.988          & 19.269          & 21.959          & 27.990          & 27.282          & 18.811          & 23.300          \\
    GAM 6 attributes + work                       & 23.426          & 21.306          & 19.551          & 24.343          & 21.137          & 25.340          & 29.078          & \textbf{18.694} & 22.234          & 27.641          & 26.874          & 19.061          & 23.224          \\
    GAM 6 attributes +   basis\_steps             & 22.953          & 20.964          & 19.545          & 24.707          & 22.111          & 25.355          & 29.460          & 19.192          & 22.082          & 27.663          & 27.695          & 19.243          & 23.414          \\
    GAM 6 attributes +   basis\_heart\_rate       & 22.570          & 21.064          & 19.202          & 24.719          & 22.224          & 24.985          & 29.155          & 19.144          & 22.056          & 27.671          & 27.149          & 19.258          & 23.266          \\
    GAM 6 attributes +   basis\_air\_temperature  & 22.750          & 20.876          & 19.880          & 25.250          & 22.529          & 25.317          & 29.282          & 19.004          & 21.875          & 27.678          & 27.435          & 19.162          & 23.420          \\
    GAM 6 attributes +   day\_of\_week            & 22.740          & 20.760          & 19.549          & 24.580          & 21.337          & 25.259          & 29.446          & 19.096          & 21.777          & 28.126          & 27.810          & 18.679          & 23.263          \\
    GAM 6 attributes +   total\_seconds           & 22.889          & \textbf{20.696} & \textbf{19.017} & 24.531          & \textbf{20.985} & 25.190          & \textbf{28.347} & 18.772          & 21.917          & \textbf{26.465} & 27.298          & 18.655          & \textbf{22.897} \\
    GAM 6 attributes                              & 22.653          & 20.901          & 19.028          & 24.383          & 21.678          & 25.094          & 29.114          & 19.300          & 21.960          & 27.540          & 27.488          & 19.235          & 23.198          \\ \hline
    \end{tabular}
}
\label{table:12_mae_attri}
\end{table}

As for other attributes, in Tables \ref{table:6_rmse_attri}-\ref{table:12_mae_attri}, we can observe that, in most cases, introducing new attributes will not improve the performance.
For example, only when $W=6$, considering ``basis\_sleep'' can improve the performance in a manner, while it cannot improve the performance obviously when $W=12$.
Hence, ``basis\_sleep'' is not taken into account in the selection of attributes.
Similarly, temporal attributes, i.e., ``day\_of\_week'' and ``total\_seconds'' cannot bring about obvious improvements in most cases, where ``total\_seconds'' is the total seconds transformed from the time of a day.
For example, when the time is ``5:10:01'', the ``total\_seconds'' is $5*60*60+10*60+1=18601$.
Therefore, temporal attributes are also ignored in the final combination of attributes.
Hence, except the 6 attributes, no more attributes are considered.

Besides, we can also find an interesting observation. 
In the 6 attributes, excluding ``glucose\_level'', all attributes are self-reported.
Meanwhile, other attributes collected by devices, e.g., ``basis\_heart\_rate'' cannot ensure the improvement of performance in most cases.
In this scenario, self-reported attributes are more important than the attributes collected by sensors, except ``glucose\_level''.

\section{Impact of Hyperparameters}
\label{sec:imp_hyper}

We select four important hyperparameters of GAM to figure out the impact of these hyperparameters, including the hidden size ($H$) of GRU-based memory, no. of heads ($M$) of GAT, no. of layers ($L$) of GAT and historical sequence length ($T$).

The hidden size ($H$) controls the memory ability of GAM.
Within certain limits, the larger the hidden size, the higher the memory ability.
We use 5 different hidden sizes in this experiment, as $H\in\{64, 128, 256, 512, 1024\}$.

Changing the no. of heads ($M$) can vary the representation ability of GAM.
A larger $M$ enables GAM to have multiple groups of attention weights, and graph embeddings have more ways to be merged with each other.
For example, if the ``glucose\_level'' node and ``meal'' node are activated, they can generate and send messages to each other.
If $M=1$, in terms of ``glucose\_level'' node, it may only summarize the messages from ``glucose\_level'' node and ``meal'' node via a group of weights $(0.8, 0.2)$.
If $M=2$, in terms of ``glucose\_level'' node, apart from the group of weights $(0.8, 0.2)$, the ``glucose\_level'' node may have one more group of weights, e.g., $(0.6, 0.4)$ to aggregate messages that it has received.
As for node $i$, it will generate $M$ new tensors after aggregating messages, and these tensors will be further aggregated to one tensor based on Equation \ref{Eqn:multi_head}.
We use 5 different no. of heads, as $M\in\{1,2,3,4,5\}$.

Varying no. of layers ($L$) of GAT can also change the representation ability of GAM from another perspective.
A larger $L$ makes the GAM more powerful.
For example, if $L=1$, in terms of ``glucose\_level'' node, the node can only receive messages from the adjacent nodes.
If $L=2$, in the first layer, the message sending and receiving are the same as the process of $L=1$.
In the second layer, even though the ``glucose\_level'' node still continuously receives messages from the adjacent nodes, the messages are not the original graph embedding.
The received messages have been mixed from the previous layer before aggregating, so deeper layer make graph messages merge with each other more times in a complicated way, similar to extending the no. of layers of convolution neural networks (CNNs).
We use 5 different no. of layers in this experiment, as $H\in\{1, 2, 3, 4, 5\}$.

The historical length $T$ of RMTS decides how much past information is considered in the prediction.
A larger $T$ means more historical records are leveraged.
We use 5 different historical length $T$ in this experiment, as $T\in\{6, 12, 24, 32, 48\}$.

\textbf{Hyperparameter settings}: prediction window is $W=6$ or $W=12$.
When hidden size $H$ changes, other hyperparameters are fixed, as $M=1$, $L=2$ and $T=24$. 
When no. of heads $M$ varies, other hyperparameters are fixed, as $H=512$, $L=2$ and $T=24$. 
When no. of layers $L$ changes, other hyperparameters are fixed, as $H=512$, $M=1$ and $T=24$.
When the length of the historical time series ($T$) changes, other hyperparameters are fixed, as $H=512$, $M=1$ and $L=2$.
The training algorithm is based on Algorithm \ref{Alg:training_no_fl}, and no. of global training epochs in the first or second training step is $T^{global}=10,000$ or $T^{person}=800$; evaluation interval in the first or second training step is $T^{eval1}=1,000$ or $T^{eval2}=160$; batch size is $B=128$; learning rate in the first or second training step is $0.001$ or $0.00001$.

\subsection{Experiment Settings}
\label{sec:hyper_setting}
\subsection{Result Analysis}
\label{sec:hyper_res}

\begin{figure*}[tb]
	\centering
	\subfigure[]{
		\begin{minipage}[t]{0.33\textwidth}
			\centering
			\includegraphics[width=1\textwidth]{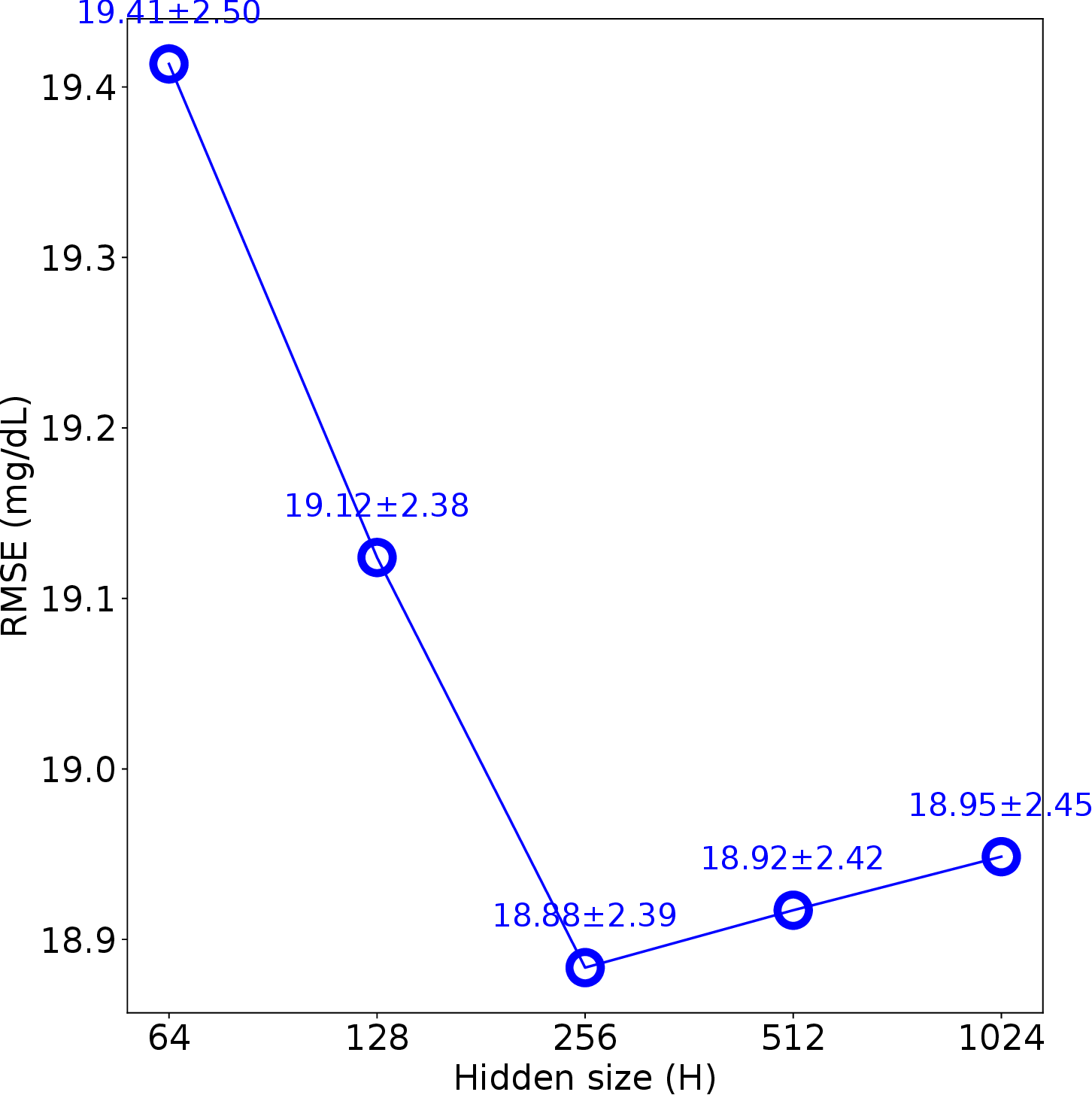}
		\end{minipage}%
		\label{Fig:6_rmse_hidden}
	}%
	\subfigure[]{
		\begin{minipage}[t]{0.33\textwidth}
			\centering
			\includegraphics[width=1\textwidth]{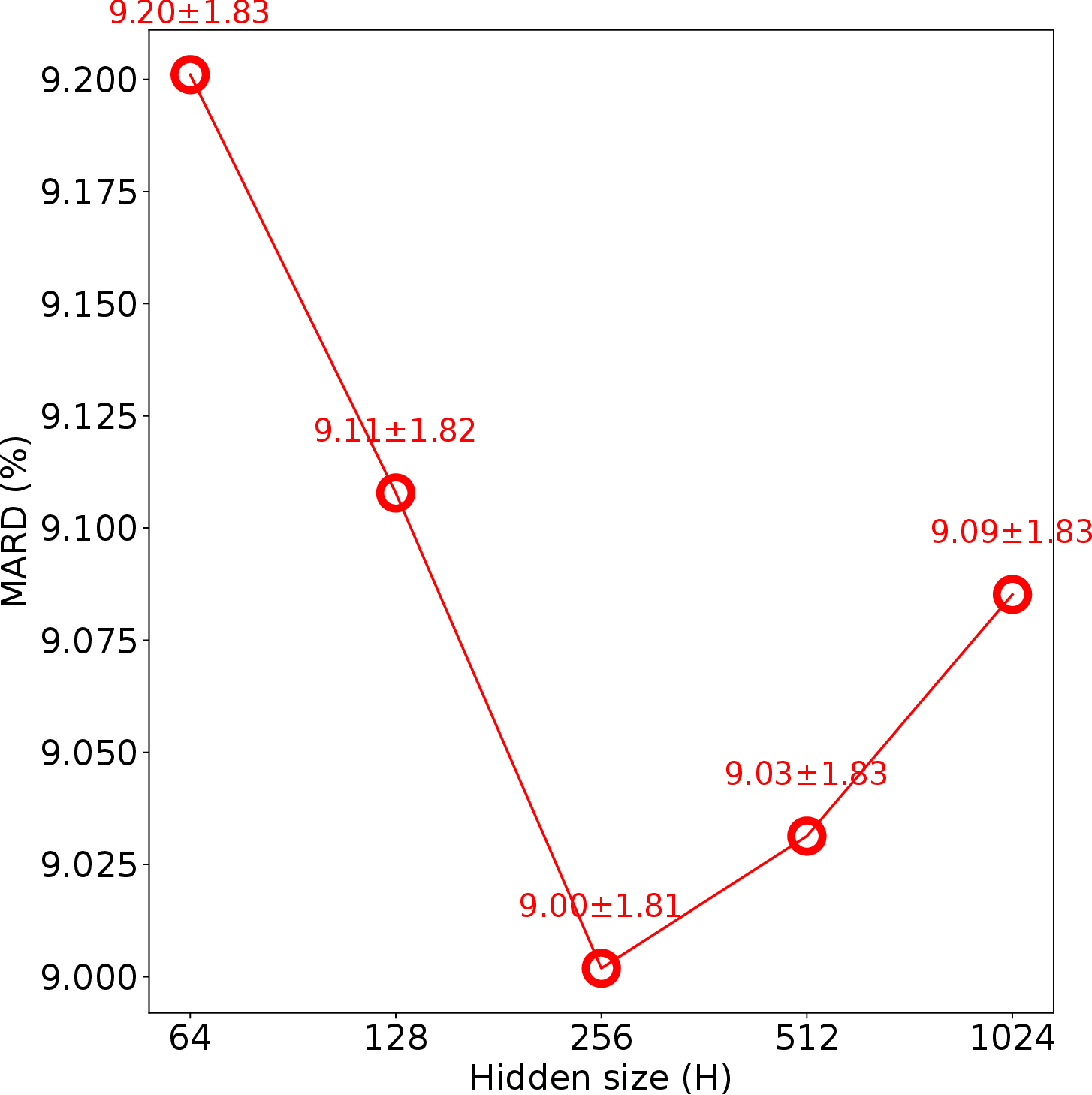}
		\end{minipage}%
		\label{Fig:6_mard_hidden}
	}%
	\subfigure[]{
		\begin{minipage}[t]{0.33\textwidth}
			\centering
			\includegraphics[width=1\textwidth]{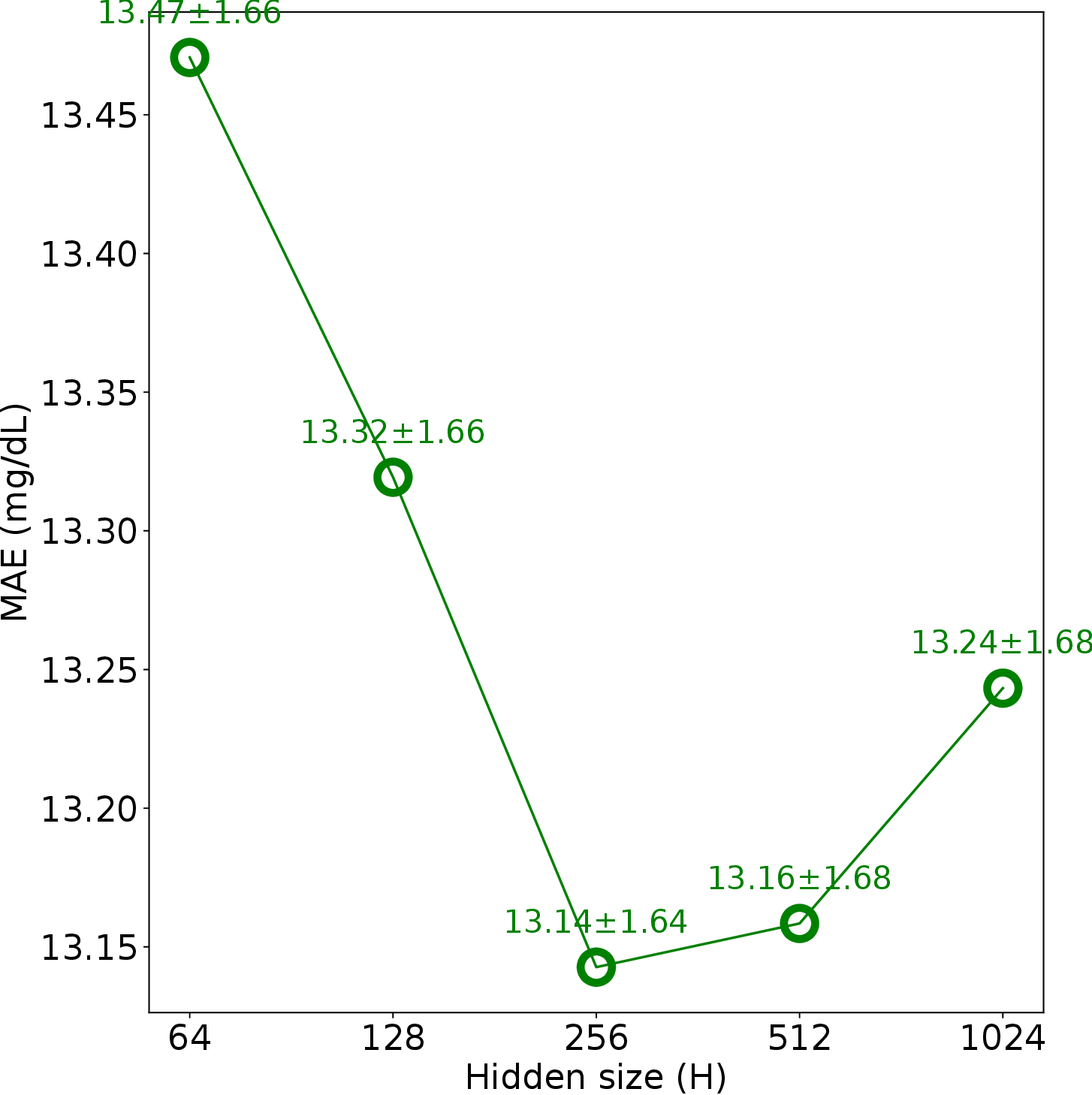}
		\end{minipage}%
		\label{Fig:6_mae_hidden}
	}%
	\vspace{-2mm}
	\caption{The impact of changing hidden size ($H$) when $W=6$.}
	\label{Fig:6_hidden}
\end{figure*}

\begin{figure*}[tb]
	\centering
	\subfigure[]{
		\begin{minipage}[t]{0.33\textwidth}
			\centering
			\includegraphics[width=1\textwidth]{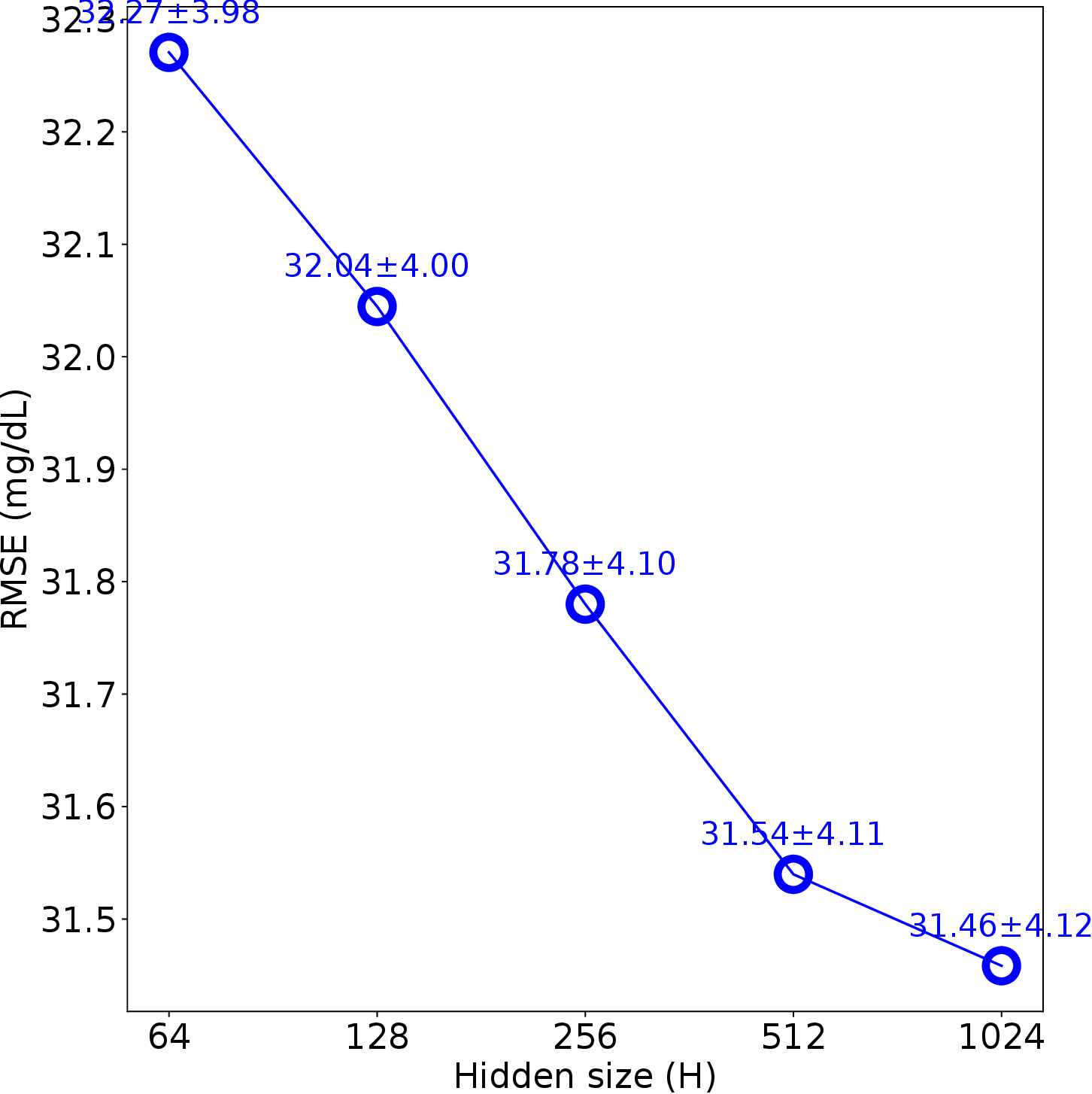}
		\end{minipage}%
		\label{Fig:12_rmse_hidden}
	}%
	\subfigure[]{
		\begin{minipage}[t]{0.33\textwidth}
			\centering
			\includegraphics[width=1\textwidth]{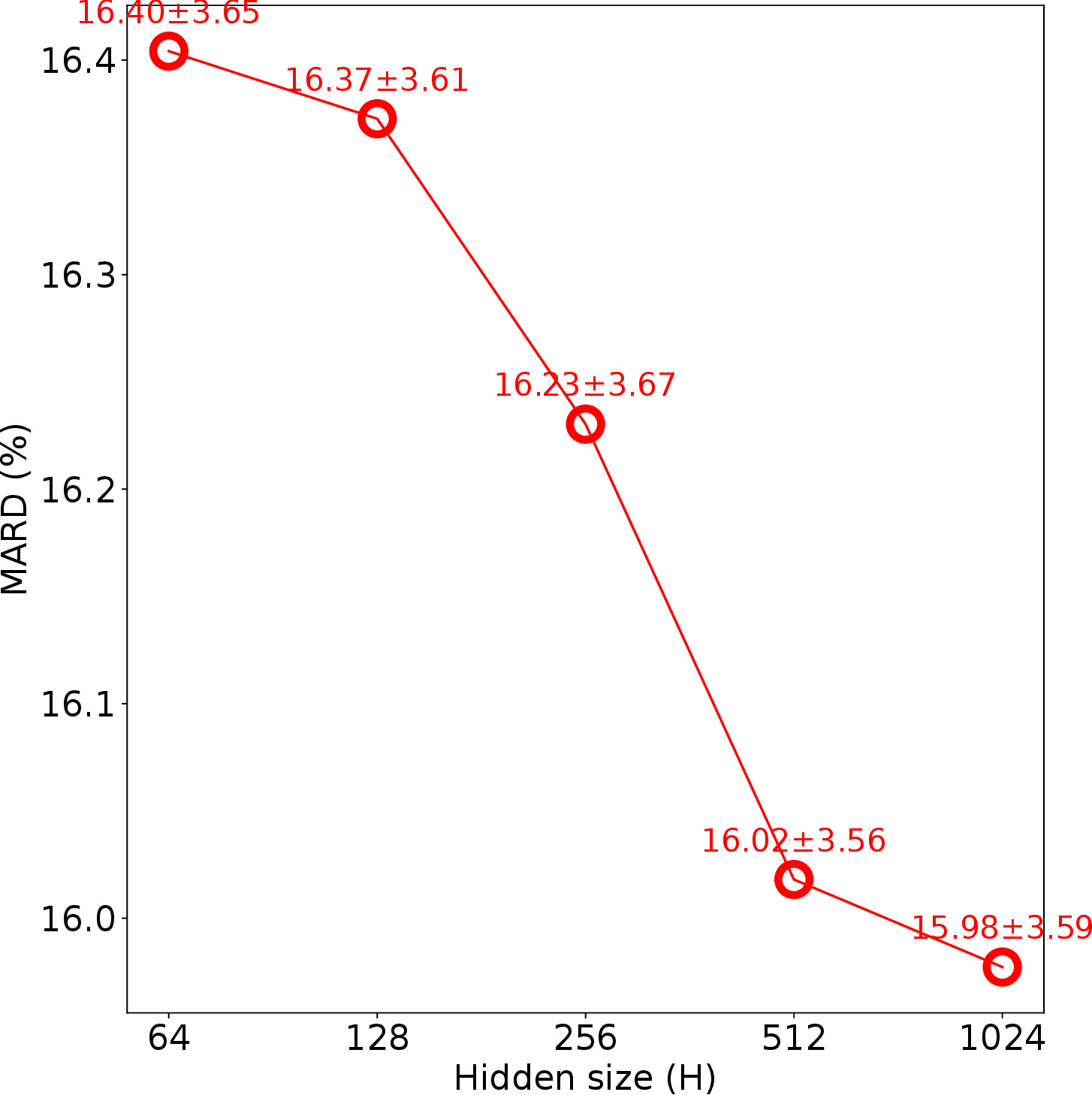}
		\end{minipage}%
		\label{Fig:12_mard_hidden}
	}%
	\subfigure[]{
		\begin{minipage}[t]{0.33\textwidth}
			\centering
			\includegraphics[width=1\textwidth]{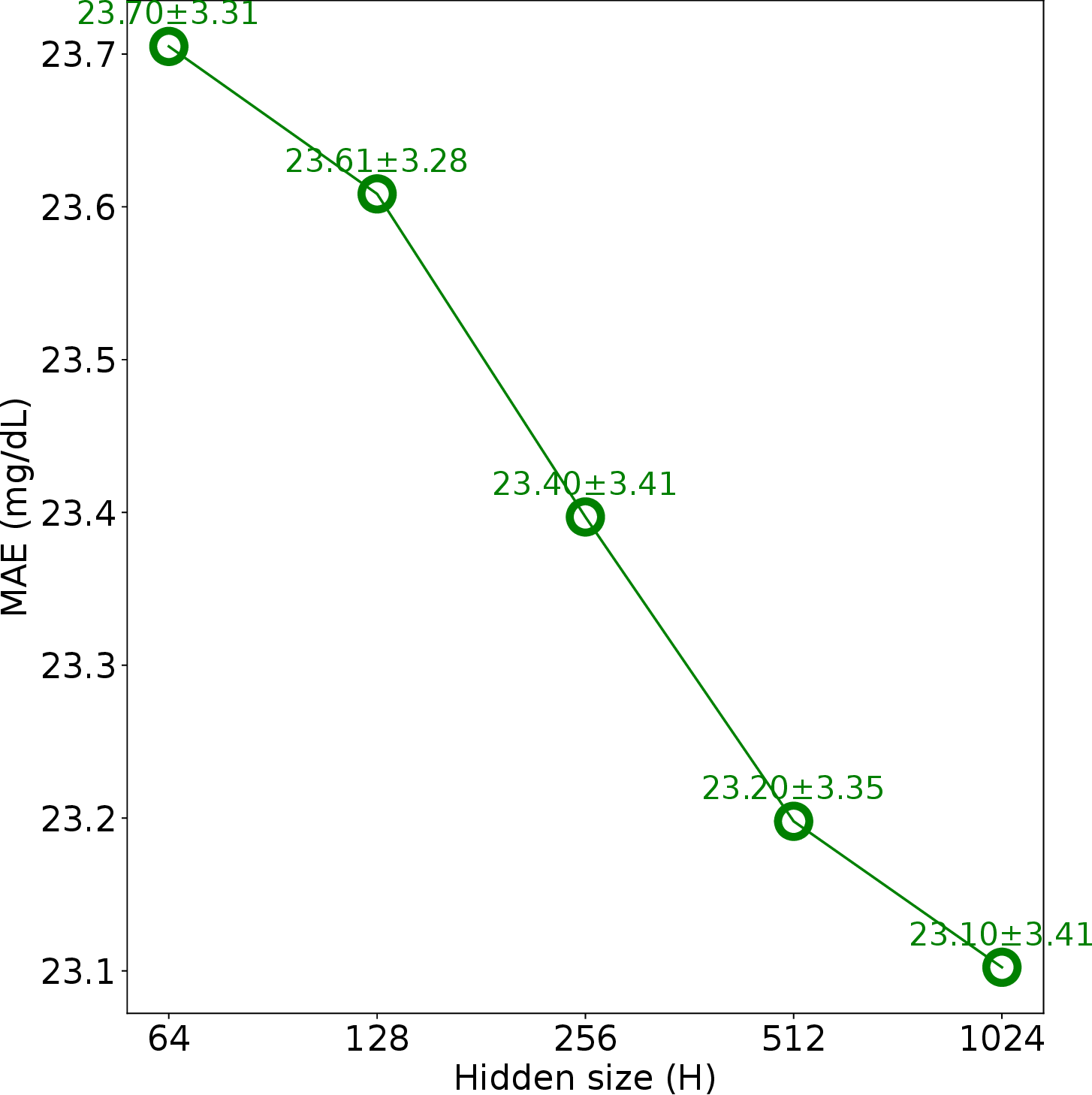}
		\end{minipage}%
		\label{Fig:12_mae_hidden}
	}%
	\vspace{-2mm}
	\caption{The impact of changing hidden size ($H$) when $W=12$.}
	\label{Fig:12_hidden}
\end{figure*}

\begin{figure*}[tb]
	\centering
	\subfigure[]{
		\begin{minipage}[t]{0.33\textwidth}
			\centering
			\includegraphics[width=1\textwidth]{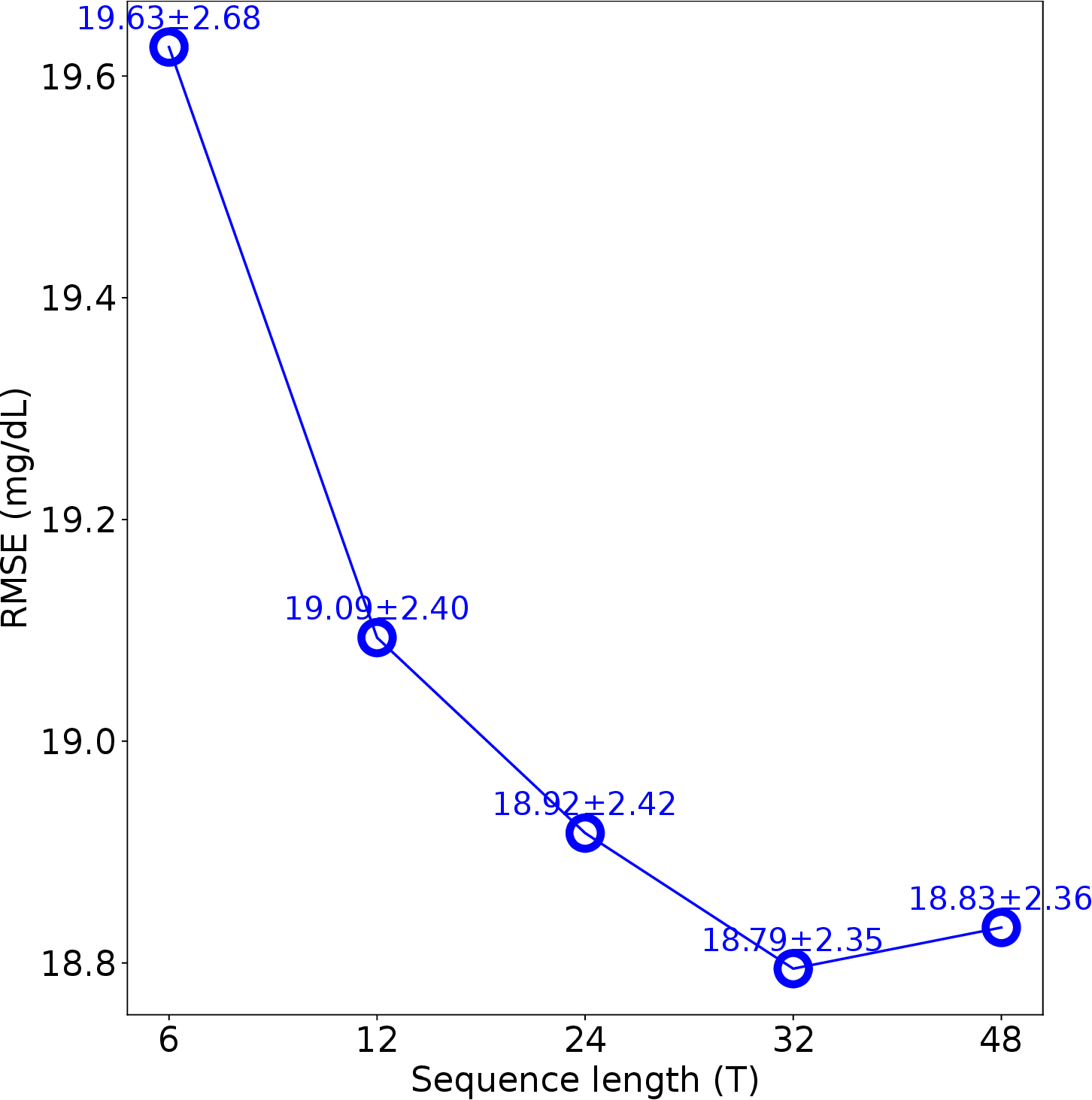}
		\end{minipage}%
		\label{Fig:6_rmse_seq_len}
	}%
	\subfigure[]{
		\begin{minipage}[t]{0.33\textwidth}
			\centering
			\includegraphics[width=1\textwidth]{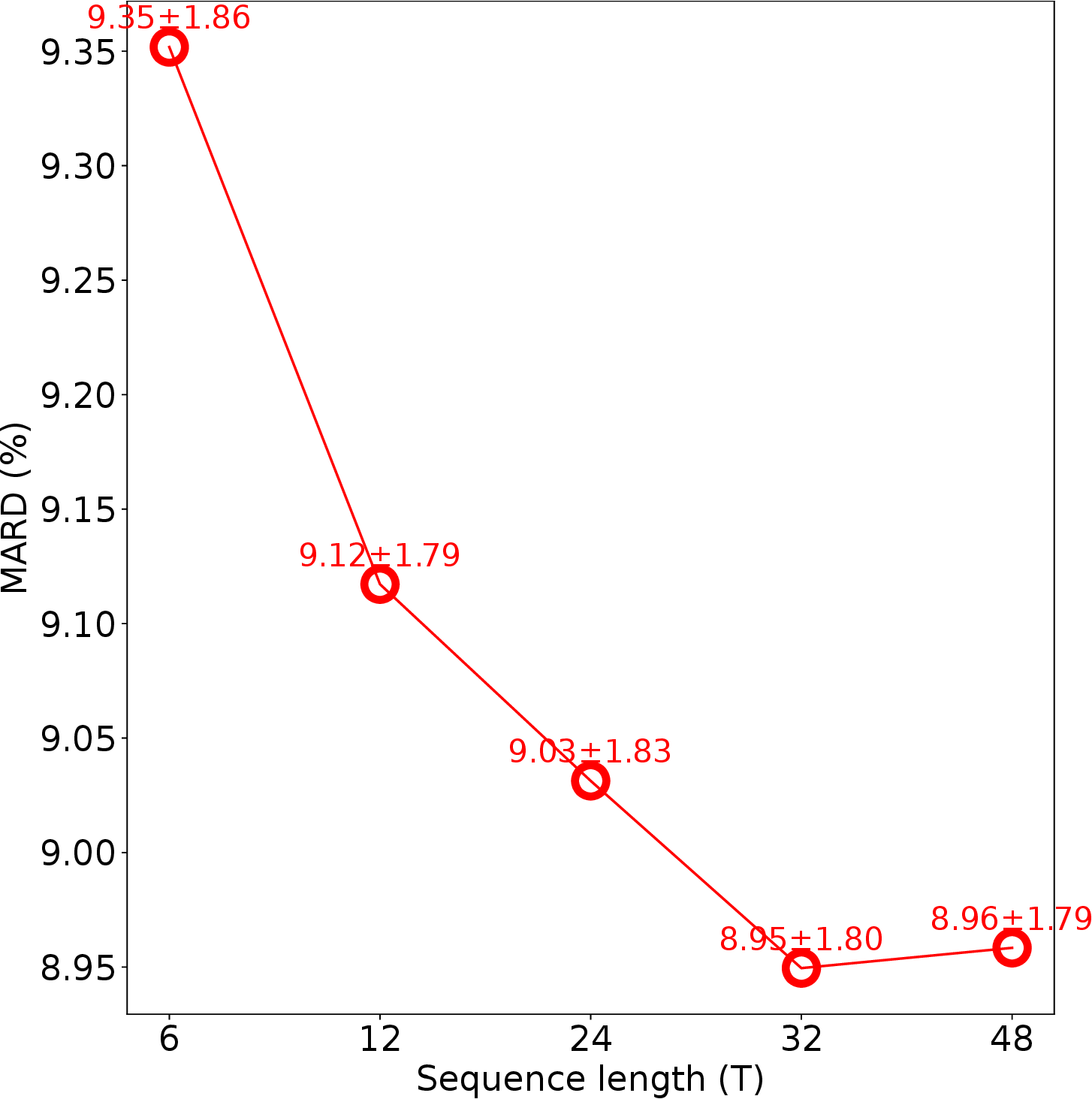}
		\end{minipage}%
		\label{Fig:6_mard_seq_len}
	}%
	\subfigure[]{
		\begin{minipage}[t]{0.33\textwidth}
			\centering
			\includegraphics[width=1\textwidth]{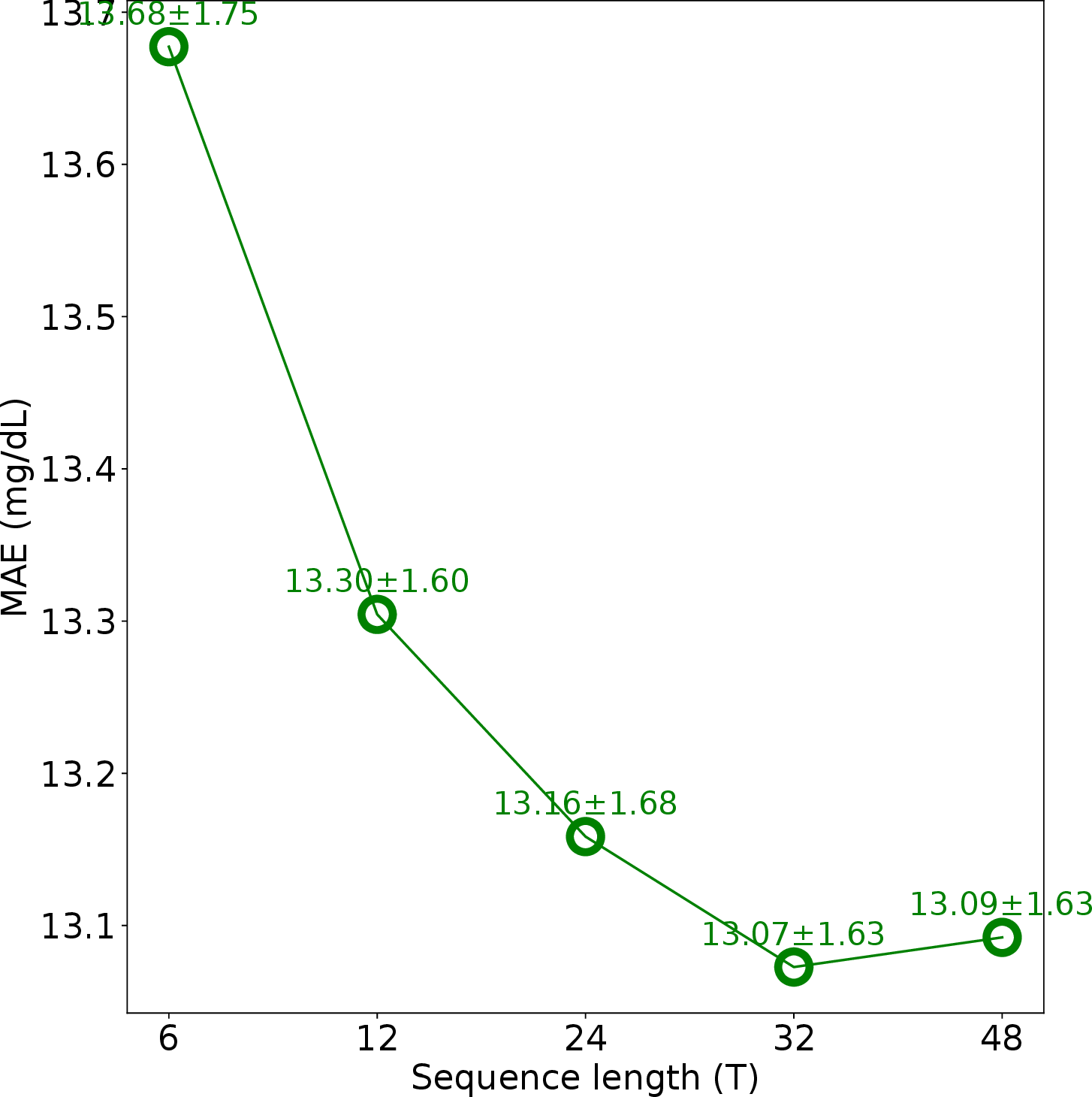}
		\end{minipage}%
		\label{Fig:6_mae_seq_len}
	}%
	\vspace{-2mm}
	\caption{The impact of changing sequence length ($T$) when $W=6$.}
	\label{Fig:6_seq_len}
\end{figure*}

\begin{figure*}[tb]
	\centering
	\subfigure[]{
		\begin{minipage}[t]{0.33\textwidth}
			\centering
			\includegraphics[width=1\textwidth]{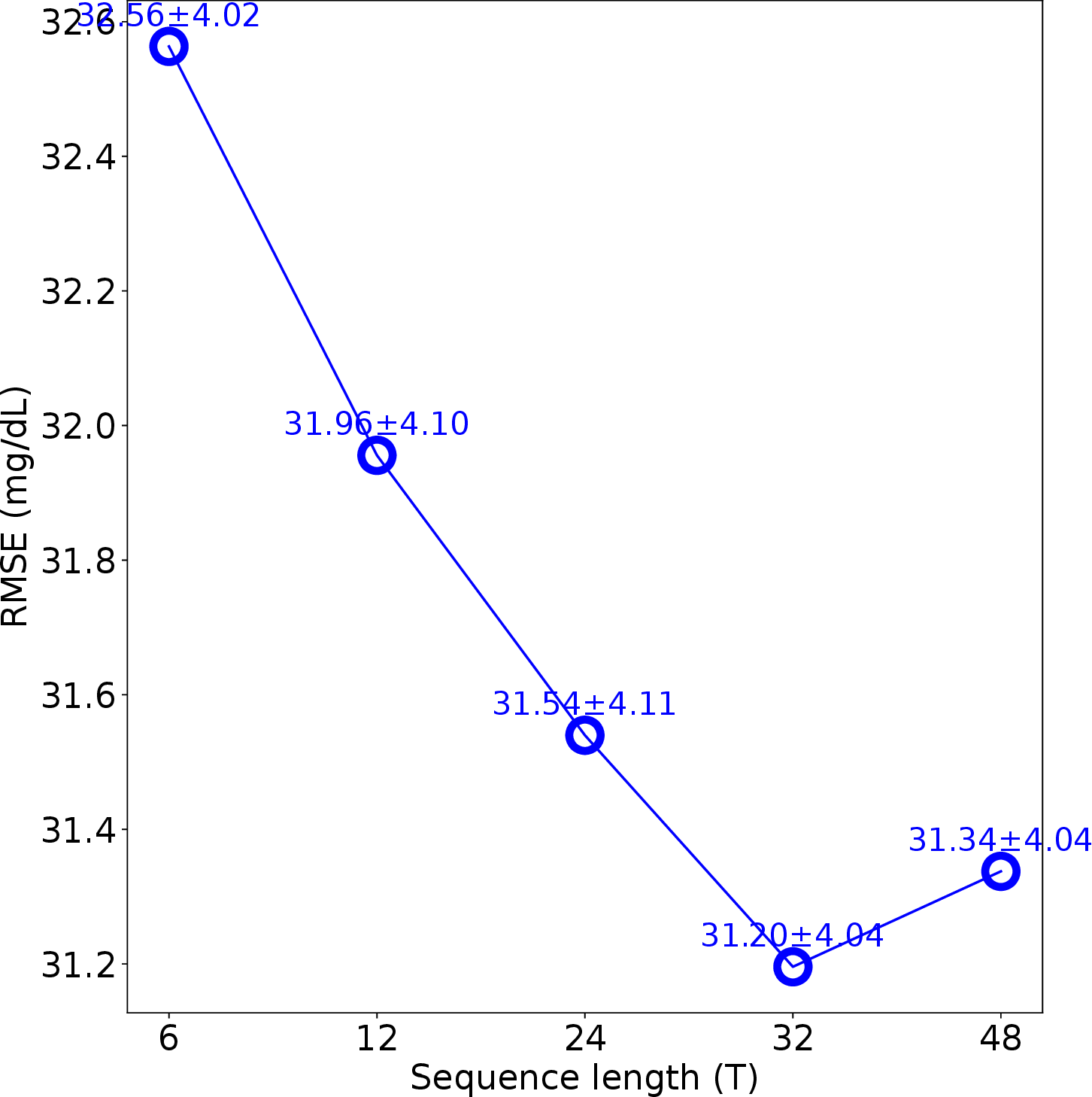}
		\end{minipage}%
		\label{Fig:12_rmse_seq_len}
	}%
	\subfigure[]{
		\begin{minipage}[t]{0.33\textwidth}
			\centering
			\includegraphics[width=1\textwidth]{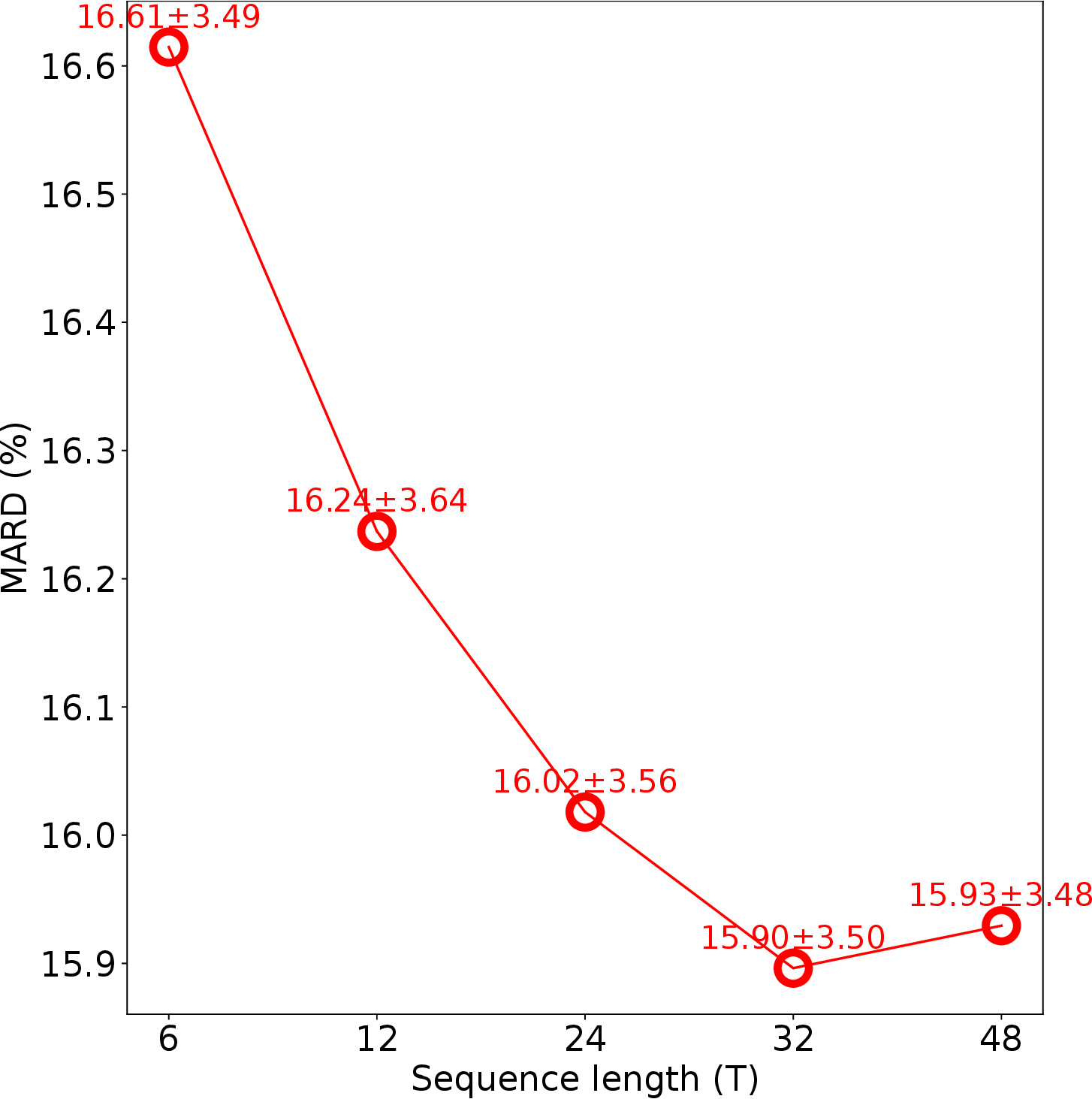}
		\end{minipage}%
		\label{Fig:12_mard_seq_len}
	}%
	\subfigure[]{
		\begin{minipage}[t]{0.33\textwidth}
			\centering
			\includegraphics[width=1\textwidth]{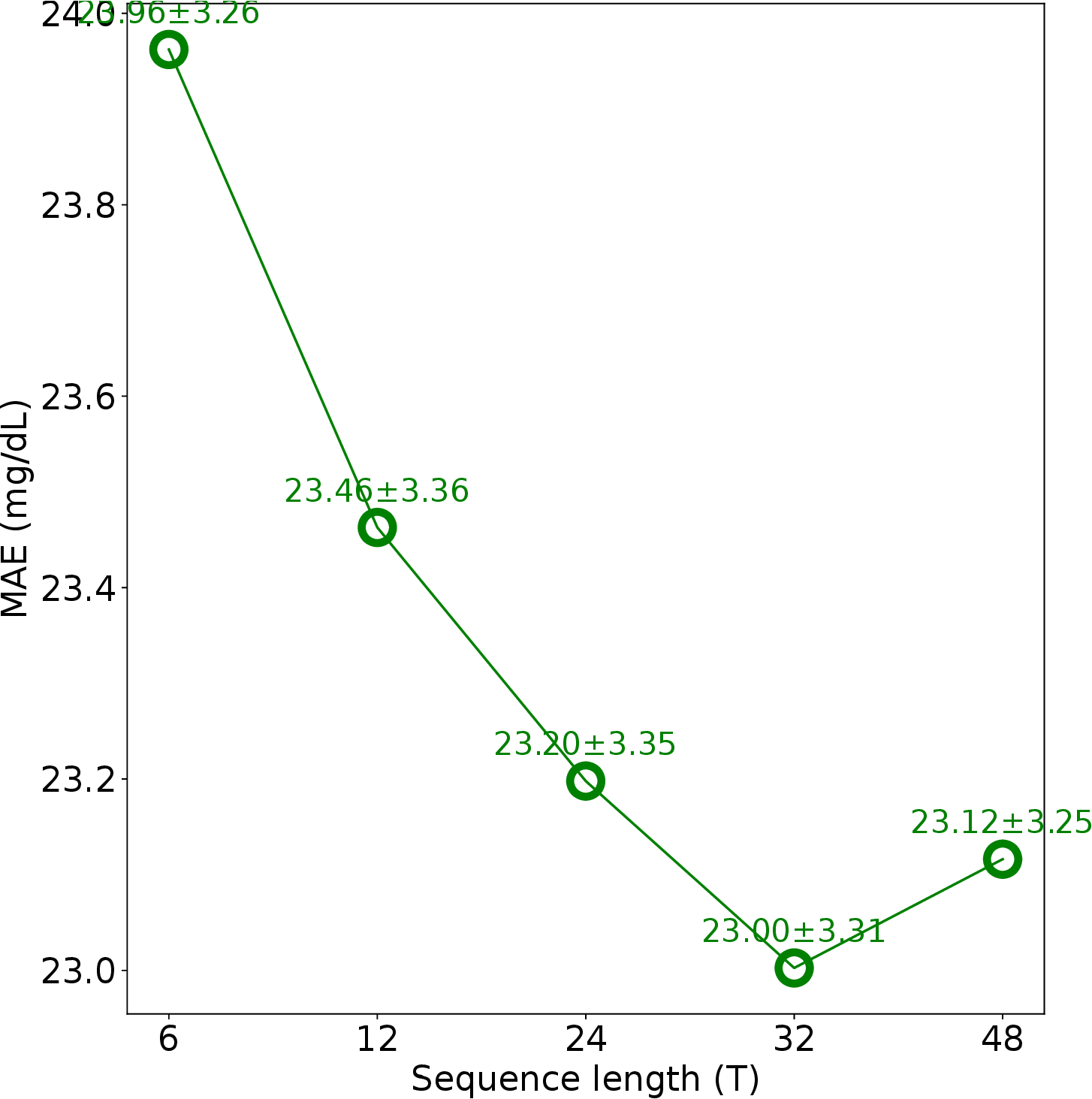}
		\end{minipage}%
		\label{Fig:12_mae_seq_len}
	}%
	\vspace{-2mm}
	\caption{The impact of changing sequence length ($T$) when $W=12$.}
	\label{Fig:12_seq_len}
\end{figure*}

The impact of hidden size ($H$) can be seen in Figures \ref{Fig:6_hidden}-\ref{Fig:12_hidden}.
We can find that if $W=6$, GAM achieves the best performance when $H=256$.
Larger $H$, i.e., $H=512$ and $H=1024$, will not lead to better performance.
Interestingly, when $W=12$, the performance is obviously improved when increasing $H$ from $64$ to $1024$.
This is because when the prediction window $W$ gets larger, the prediction is more challenging, and more historical information is needed to refer to.
Larger $H$ enables GAM to memorize more past information.
When the prediction window is not very large, i.e. $W=6$, more history, i.e., $H>256$, brings more unnecessary information, contributing to relatively worse performance.
On the other hand, we can hold the view that the hidden size ($H$) can obviously affect the whole performance.
For example, when $W=6$ and $H=256$, the average RMSE is 18.88 mg/dL, while the average RMSE changed to 19.41 mg/dL when $H=64$ (see Figure \ref{Fig:6_rmse_hidden}).
Similarly, in Figures \ref{Fig:6_seq_len}-\ref{Fig:12_seq_len}, we can also find that within a certain range, increasing $T$ can obviously and positively affect performance.
For example, in Figure \ref{Fig:6_rmse_seq_len}, when increase $T$ from $6$ to $32$, the average RMSE decreases from 19.63 mg/dL to 18.79 mg/dL.
This is because larger $T$ provides GAM with more useful historical information.

\begin{figure*}[tb]
	\centering
	\subfigure[]{
		\begin{minipage}[t]{0.33\textwidth}
			\centering
			\includegraphics[width=1\textwidth]{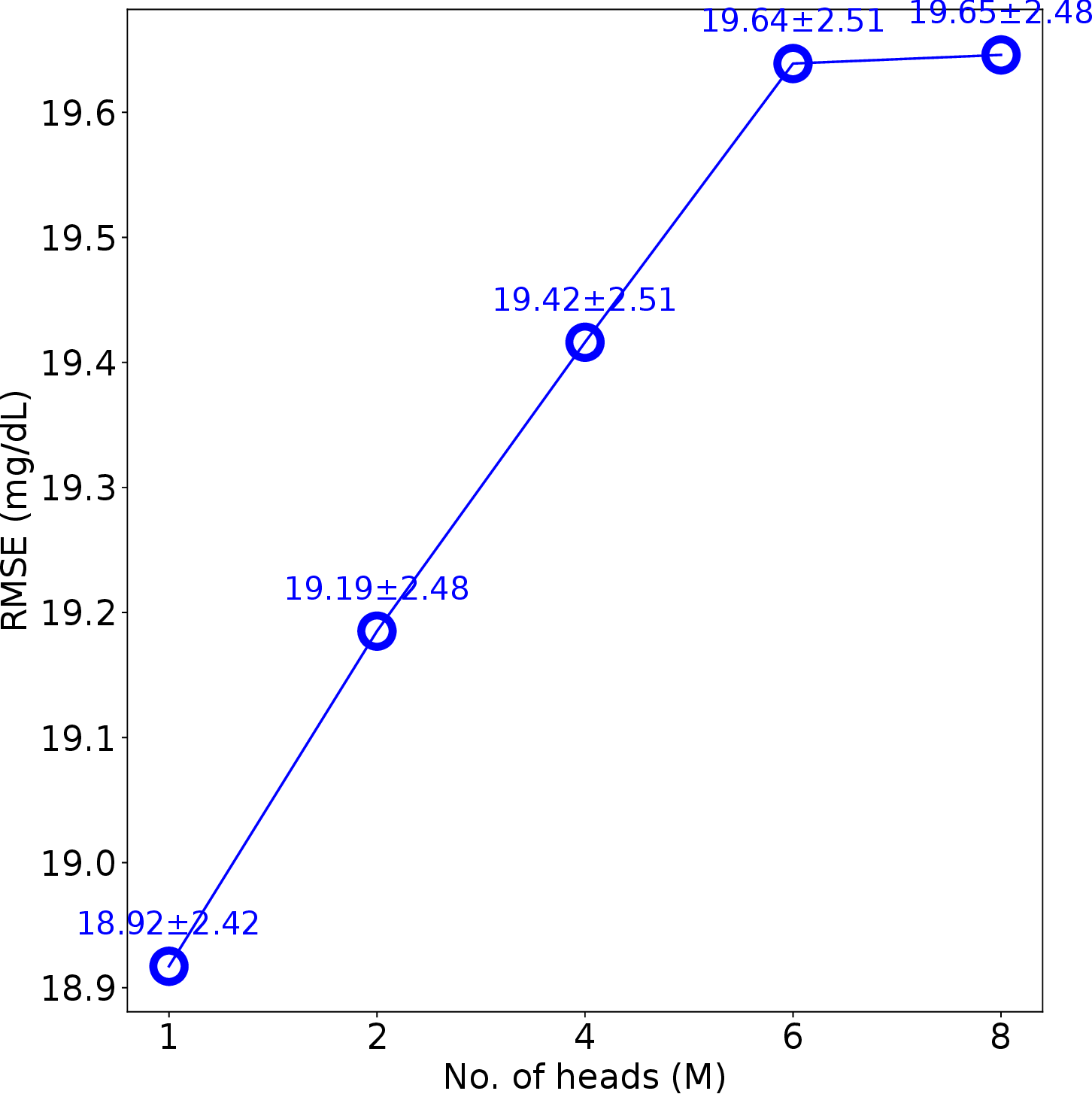}
		\end{minipage}%
		\label{Fig:6_rmse_nheads}
	}%
	\subfigure[]{
		\begin{minipage}[t]{0.33\textwidth}
			\centering
			\includegraphics[width=1\textwidth]{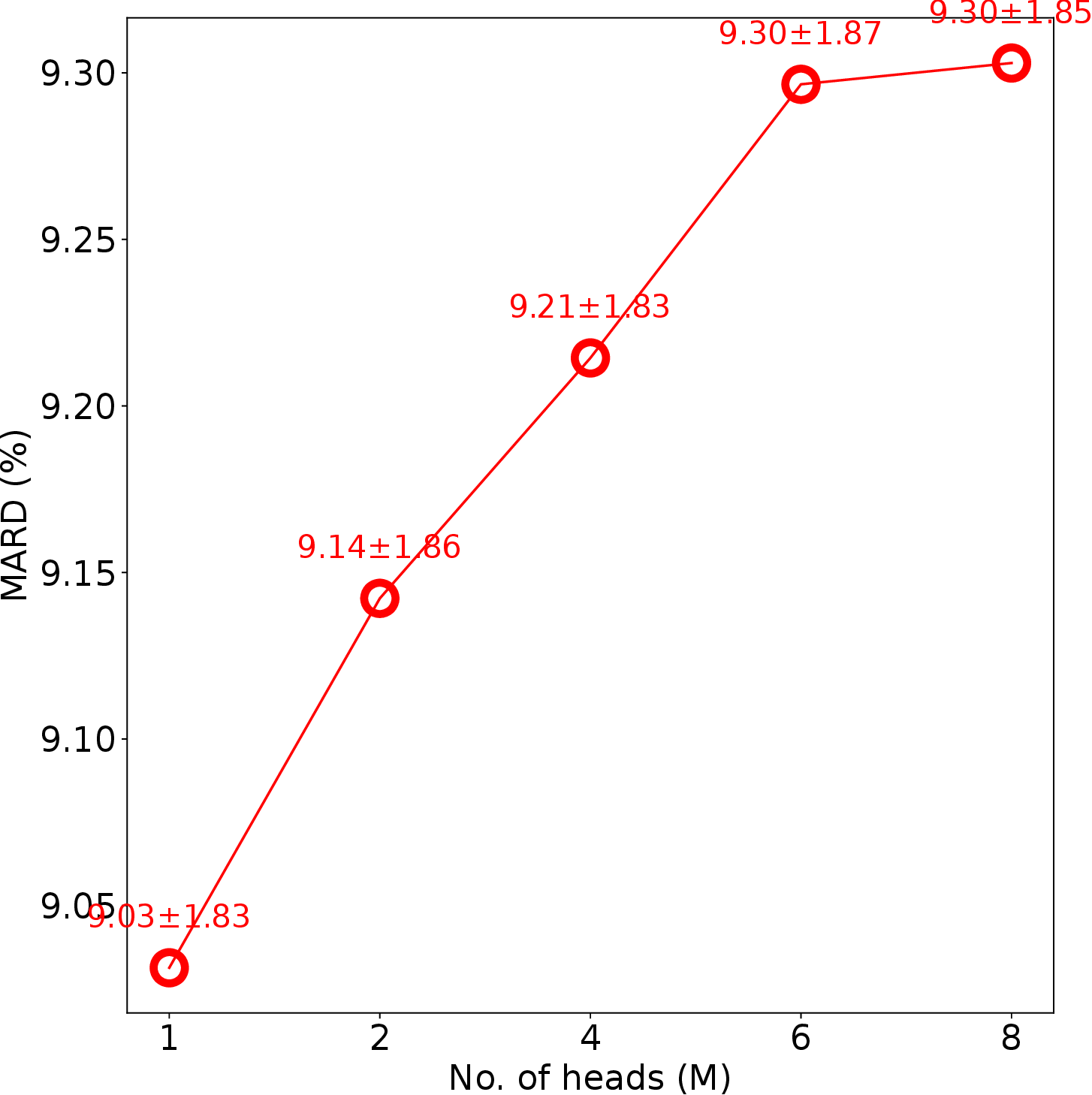}
		\end{minipage}%
		\label{Fig:6_mard_nheads}
	}%
	\subfigure[]{
		\begin{minipage}[t]{0.33\textwidth}
			\centering
			\includegraphics[width=1\textwidth]{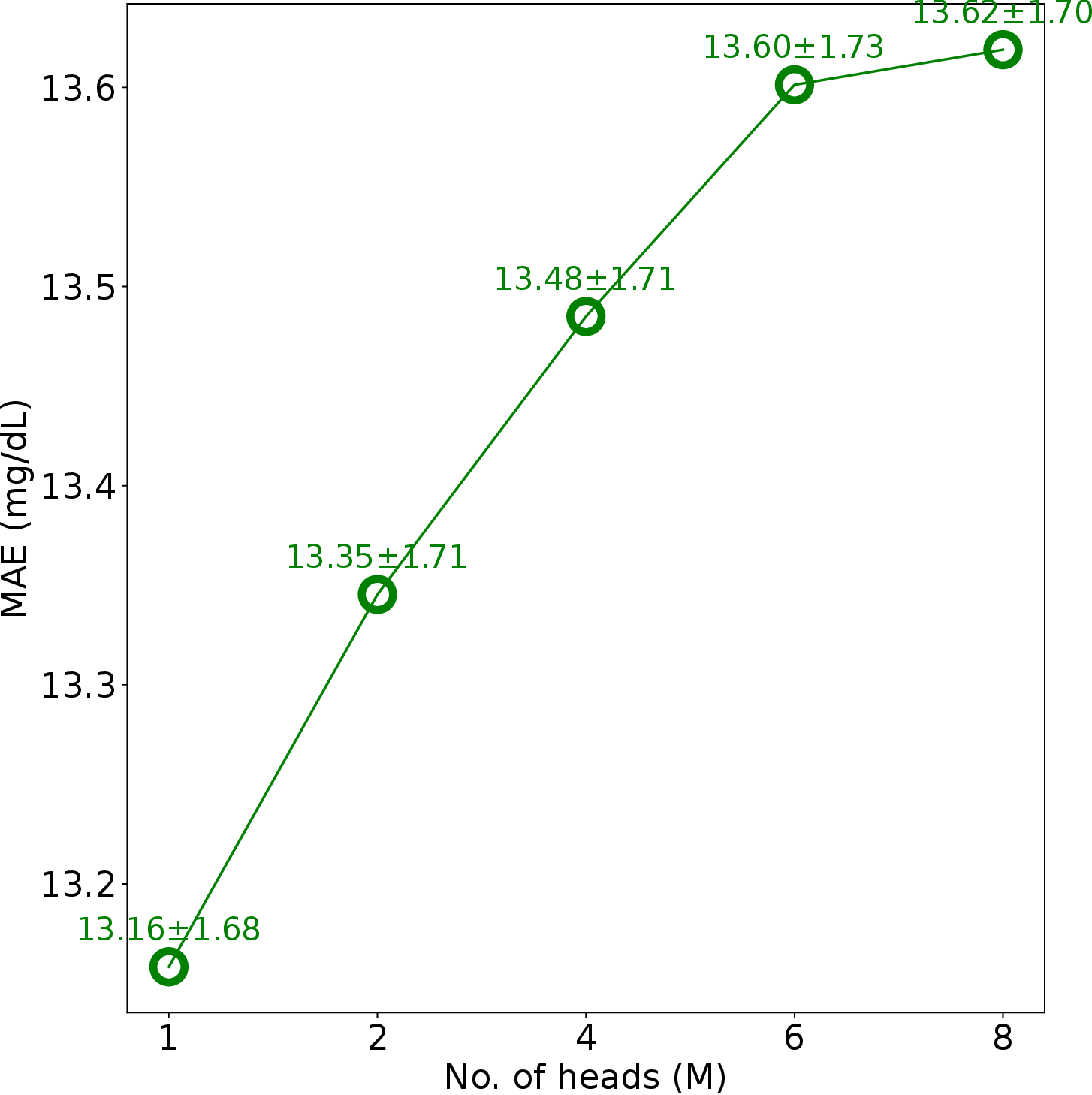}
		\end{minipage}%
		\label{Fig:6_mae_nheads}
	}%
	\vspace{-2mm}
	\caption{The impact of changing no. of heads ($M$) when $W=6$.}
	\label{Fig:6_nheads}
\end{figure*}

\begin{figure*}[tb]
	\centering
	\subfigure[]{
		\begin{minipage}[t]{0.33\textwidth}
			\centering
			\includegraphics[width=1\textwidth]{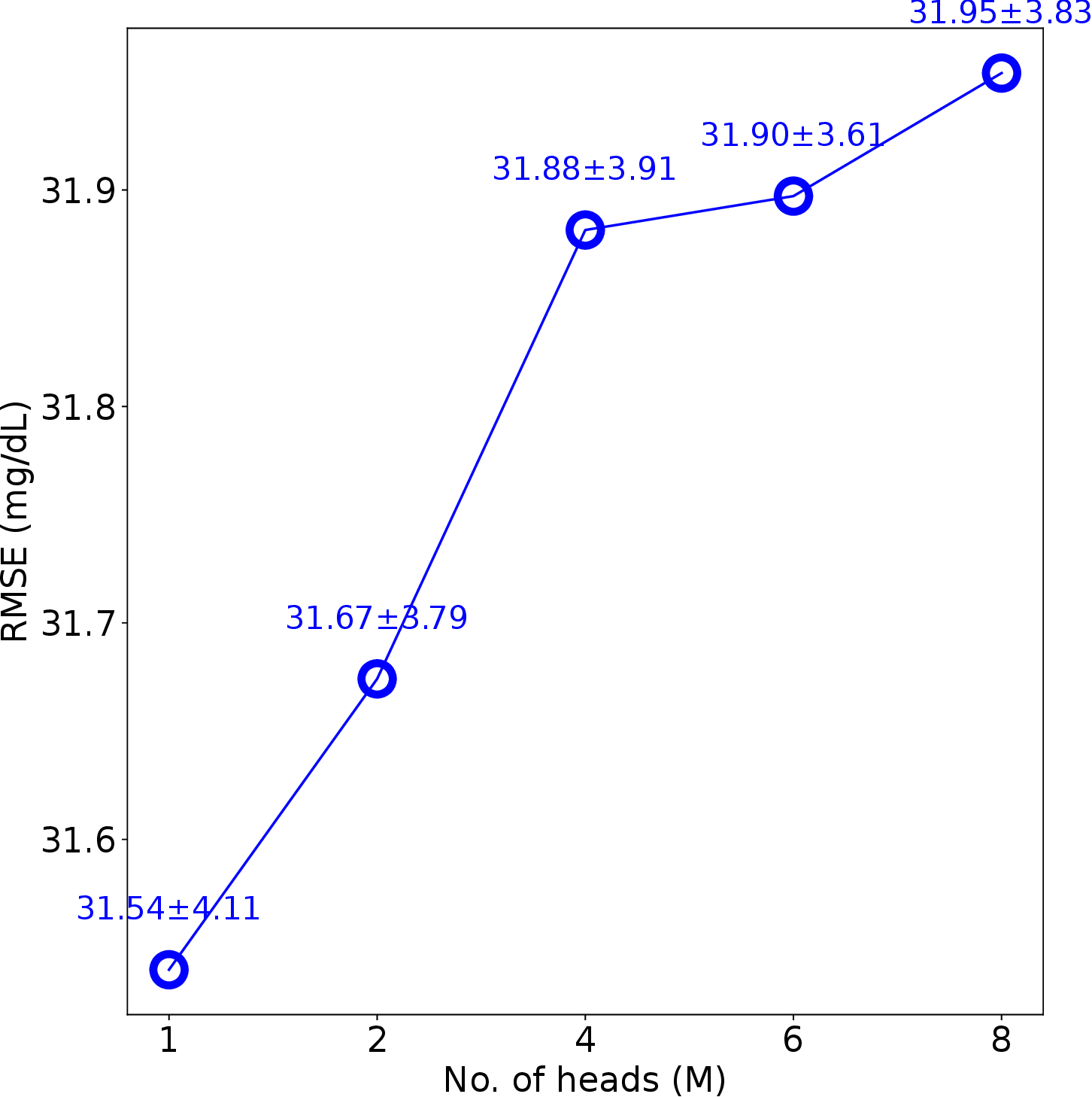}
		\end{minipage}%
		\label{Fig:12_rmse_nheads}
	}%
	\subfigure[]{
		\begin{minipage}[t]{0.33\textwidth}
			\centering
			\includegraphics[width=1\textwidth]{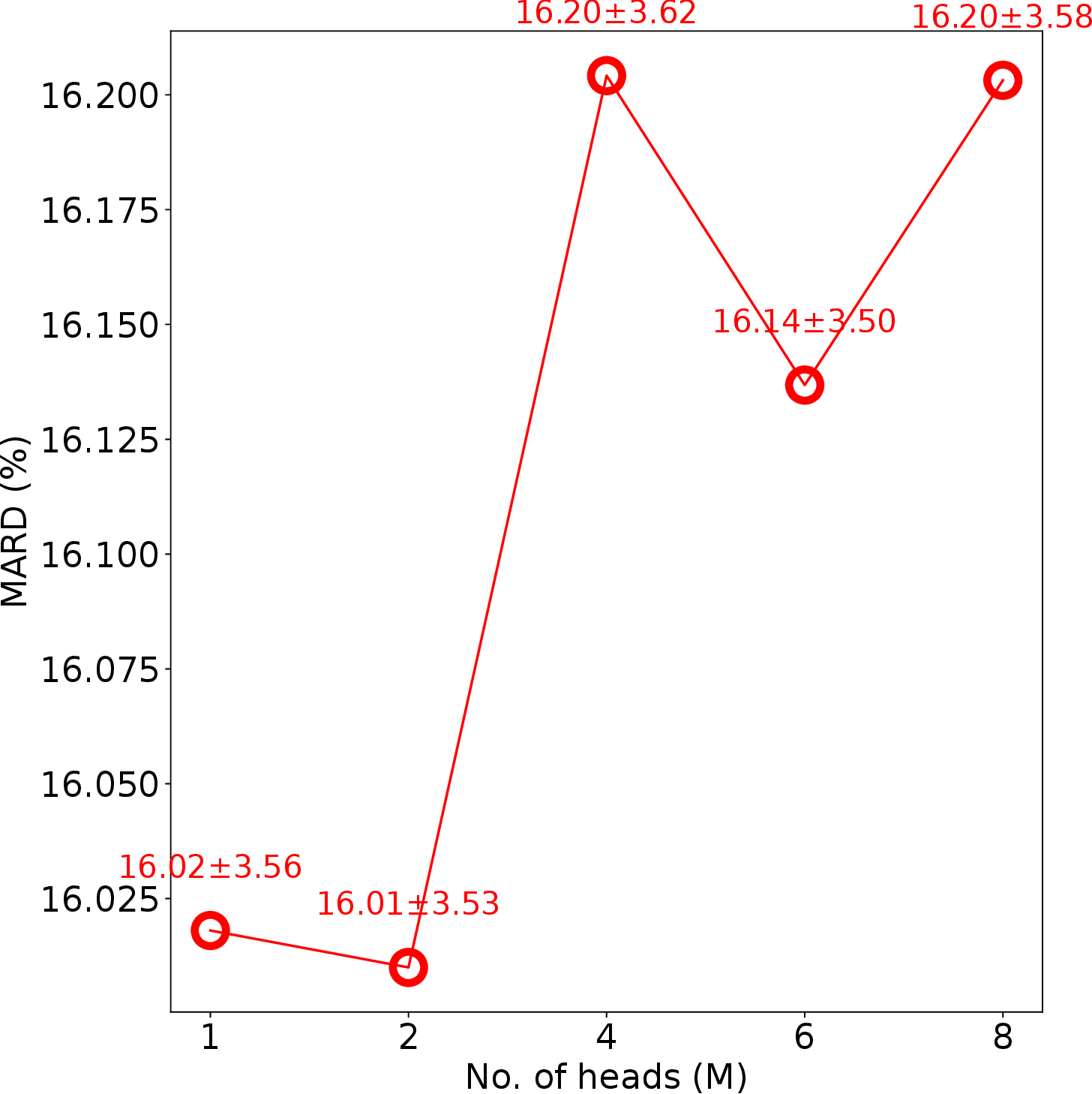}
		\end{minipage}%
		\label{Fig:12_mard_nheads}
	}%
	\subfigure[]{
		\begin{minipage}[t]{0.33\textwidth}
			\centering
			\includegraphics[width=1\textwidth]{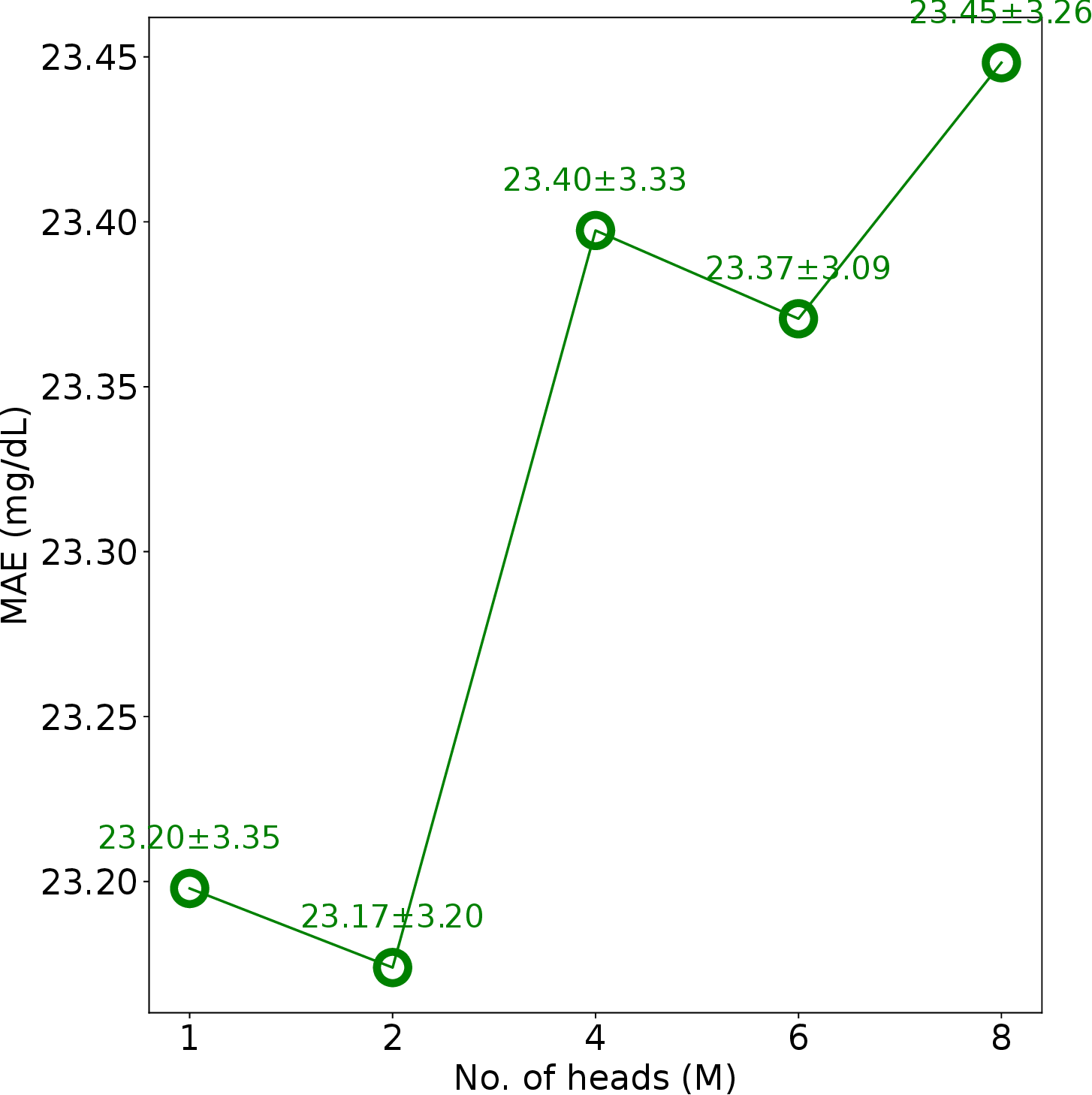}
		\end{minipage}%
		\label{Fig:12_mae_nheads}
	}%
	\vspace{-2mm}
	\caption{The impact of changing no. of heads ($M$) when $W=12$.}
	\label{Fig:12_nheads}
\end{figure*}

\begin{figure*}[tb]
	\centering
	\subfigure[]{
		\begin{minipage}[t]{0.33\textwidth}
			\centering
			\includegraphics[width=1\textwidth]{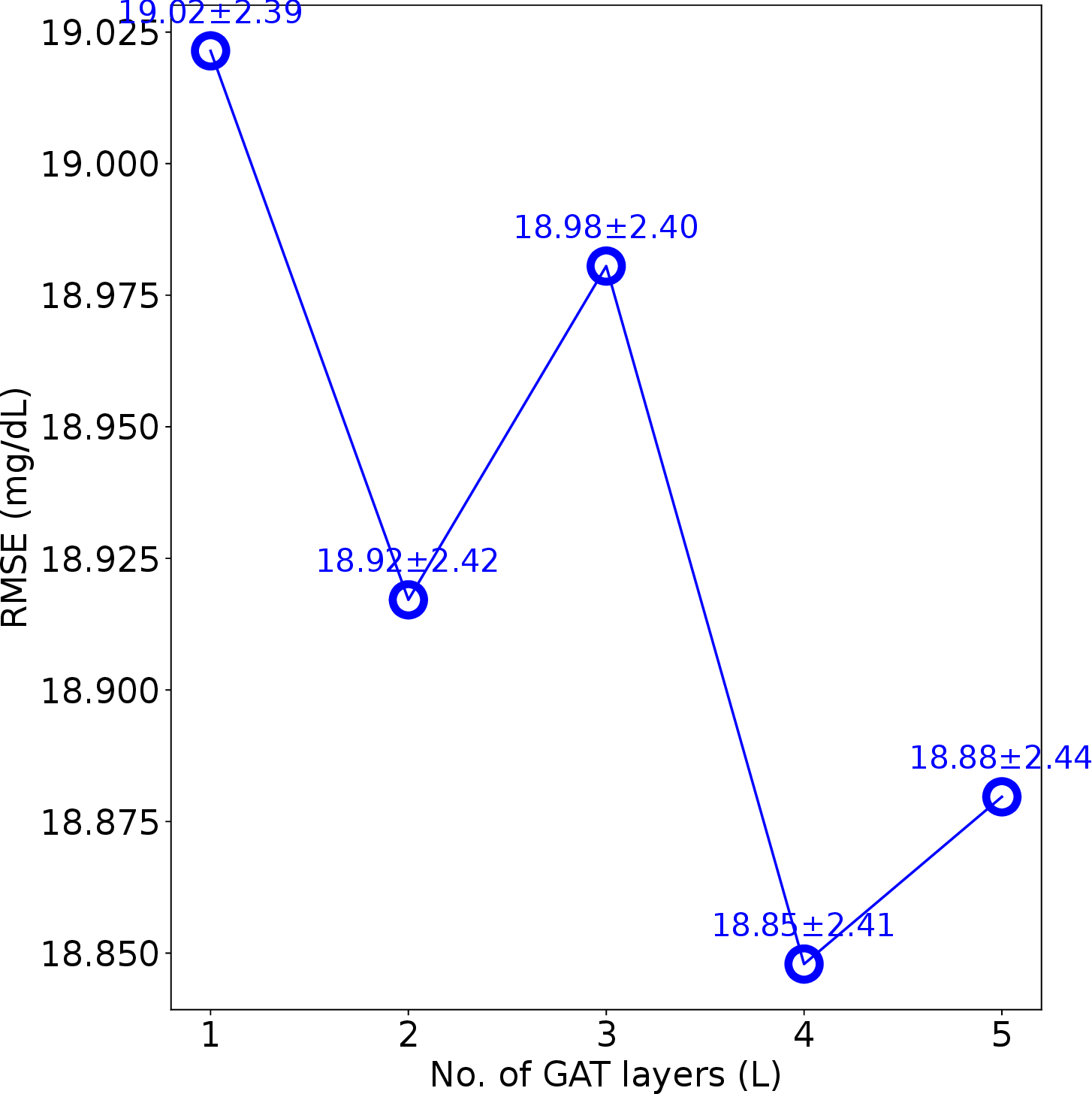}
		\end{minipage}%
		\label{Fig:6_rmse_num_layers}
	}%
	\subfigure[]{
		\begin{minipage}[t]{0.33\textwidth}
			\centering
			\includegraphics[width=1\textwidth]{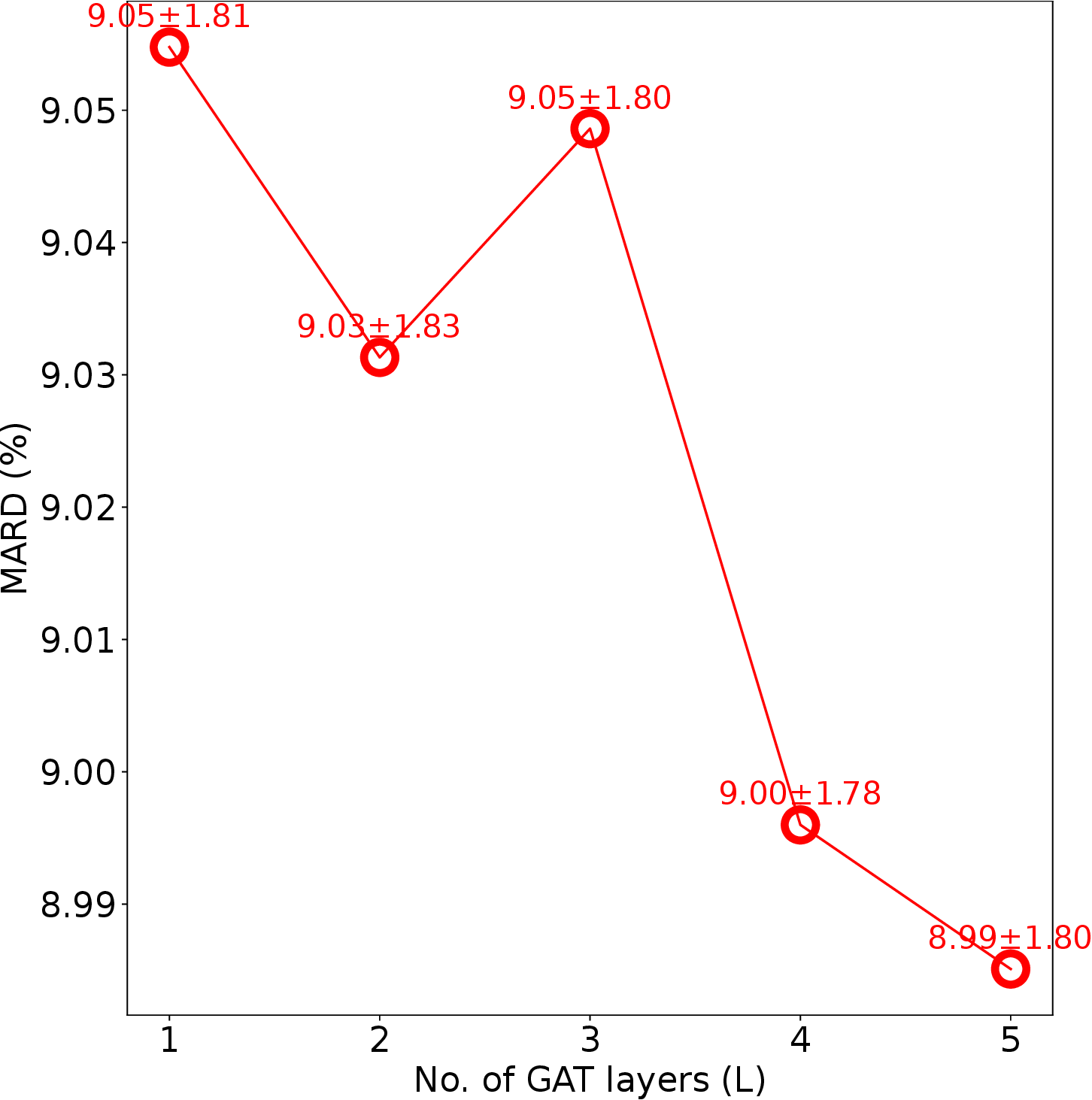}
		\end{minipage}%
		\label{Fig:6_mard_num_layers}
	}%
	\subfigure[]{
		\begin{minipage}[t]{0.33\textwidth}
			\centering
			\includegraphics[width=1\textwidth]{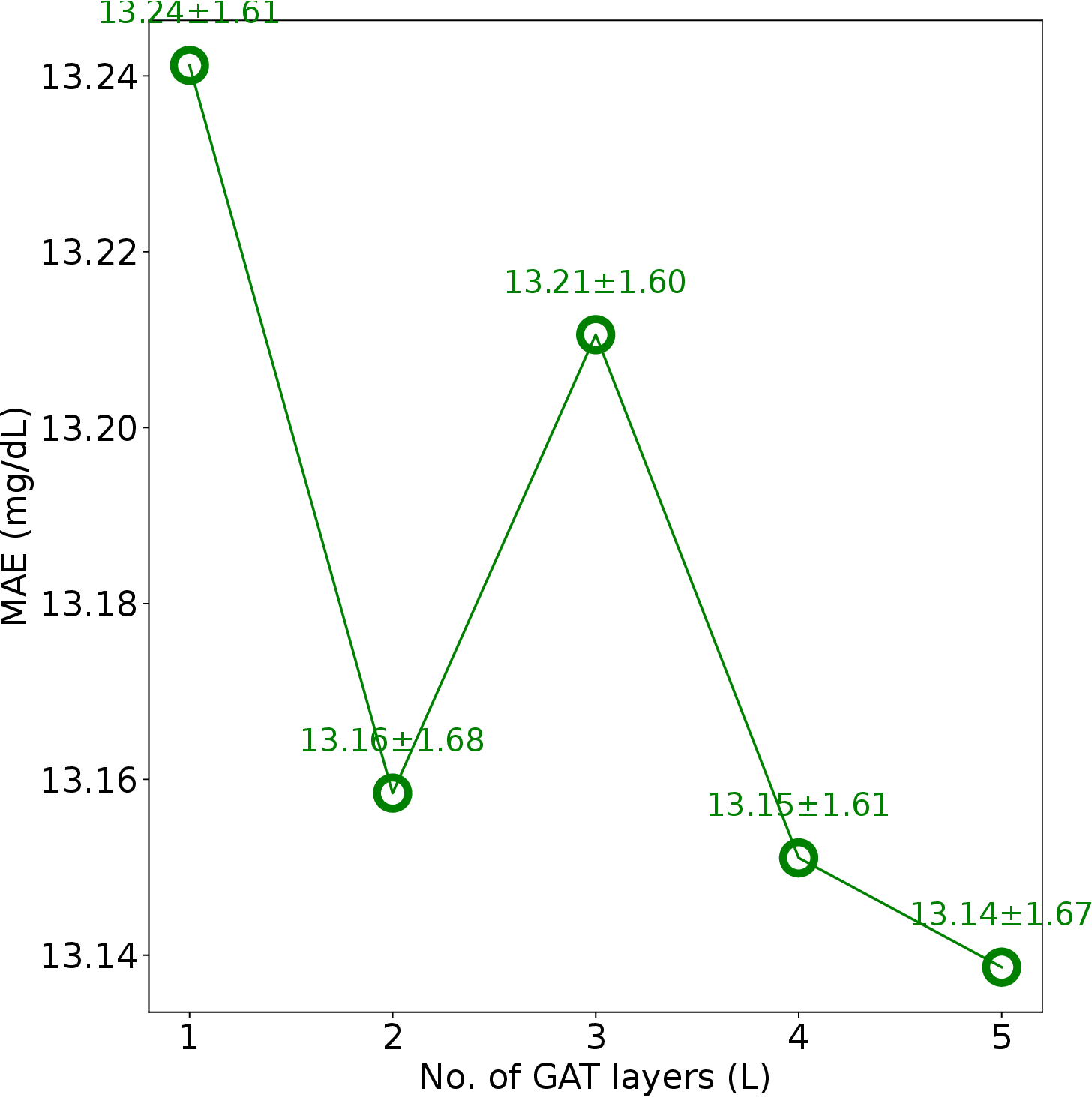}
		\end{minipage}%
		\label{Fig:6_mae_num_layers}
	}%
	\vspace{-2mm}
	\caption{The impact of changing no. of layers ($L$) when $W=6$.}
	\label{Fig:6_num_layers}
\end{figure*}

\begin{figure*}[tb]
	\centering
	\subfigure[]{
		\begin{minipage}[t]{0.33\textwidth}
			\centering
			\includegraphics[width=1\textwidth]{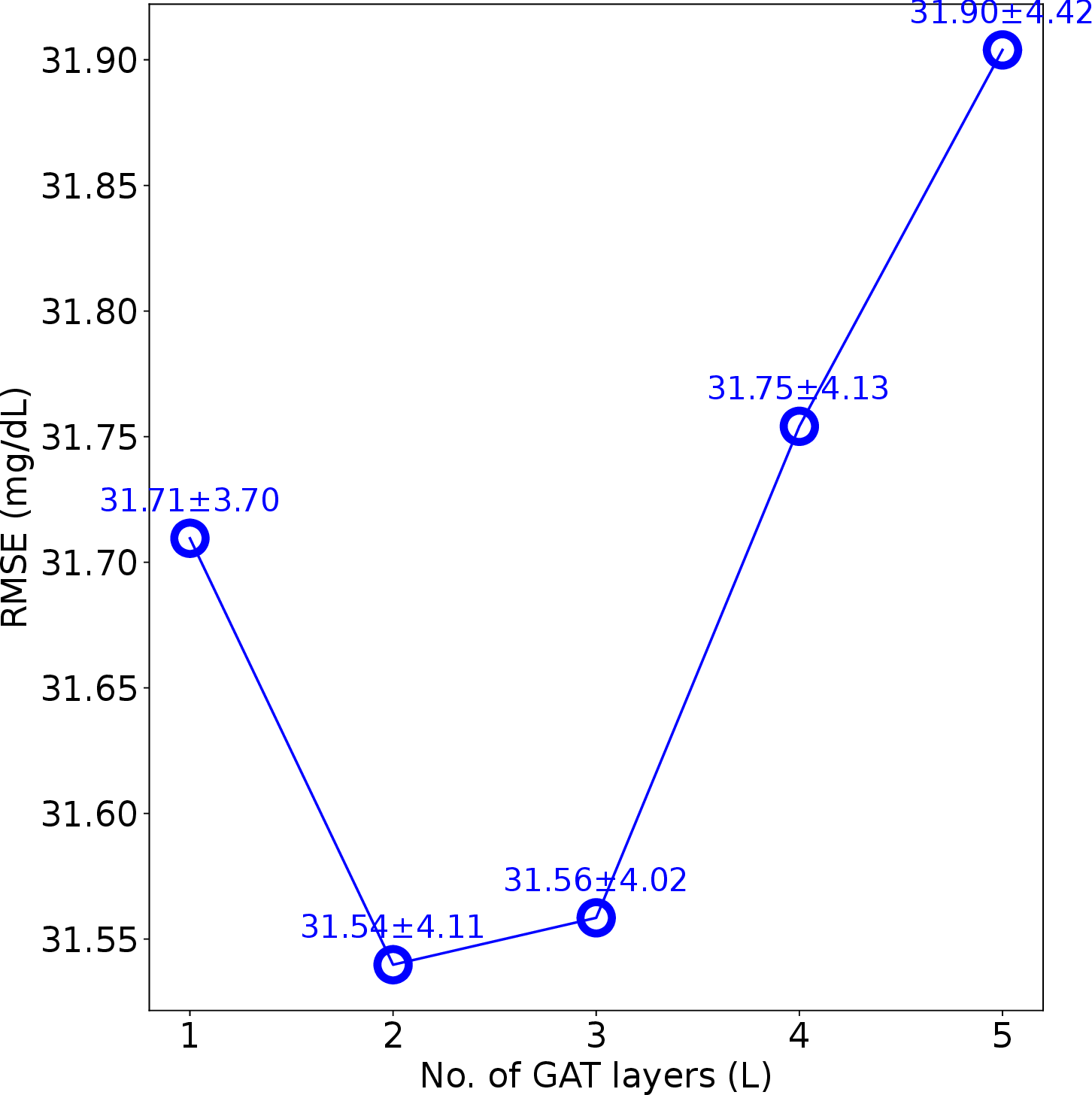}
		\end{minipage}%
		\label{Fig:12_rmse_num_layers}
	}%
	\subfigure[]{
		\begin{minipage}[t]{0.33\textwidth}
			\centering
			\includegraphics[width=1\textwidth]{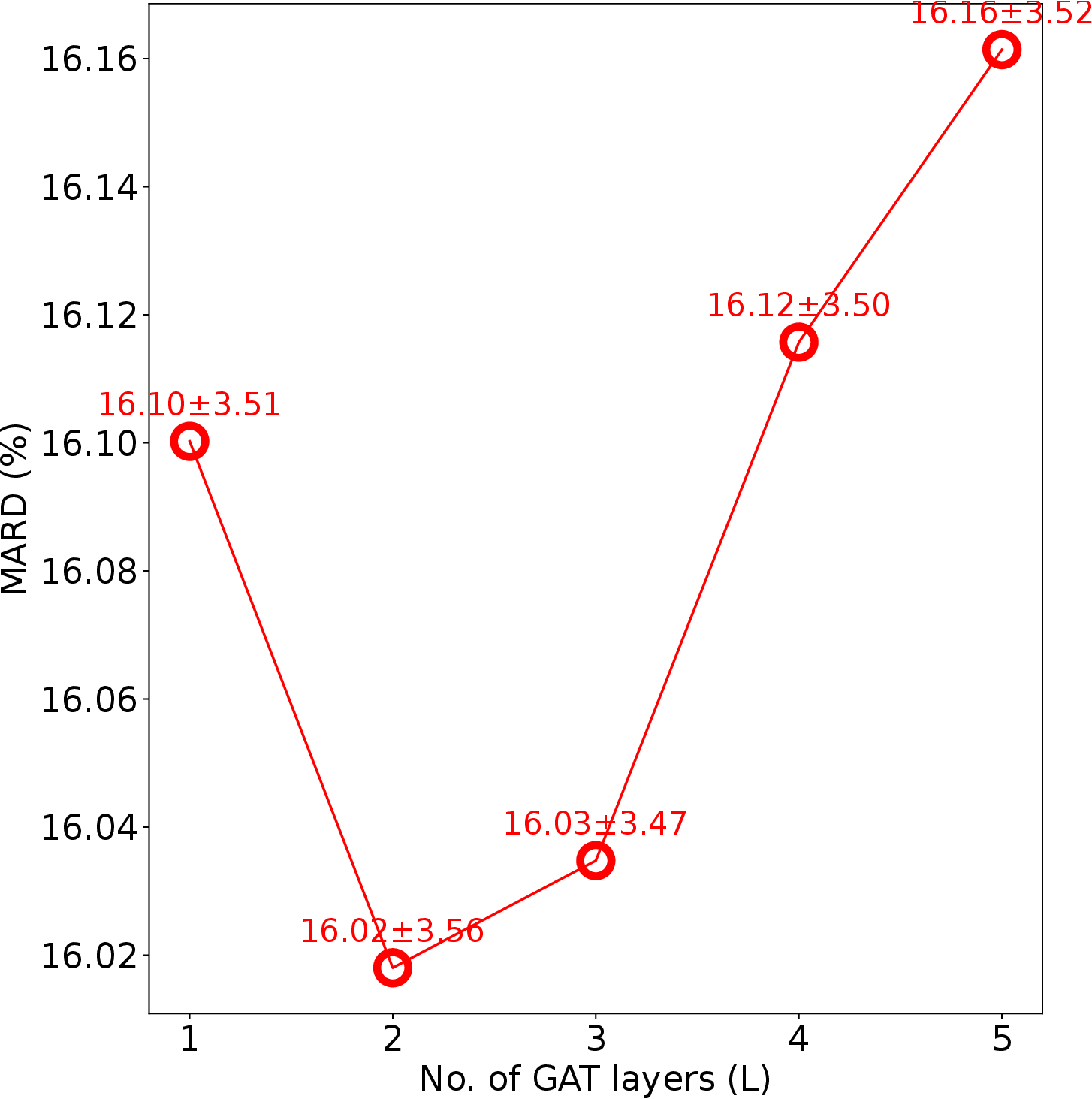}
		\end{minipage}%
		\label{Fig:12_mard_num_layers}
	}%
	\subfigure[]{
		\begin{minipage}[t]{0.33\textwidth}
			\centering
			\includegraphics[width=1\textwidth]{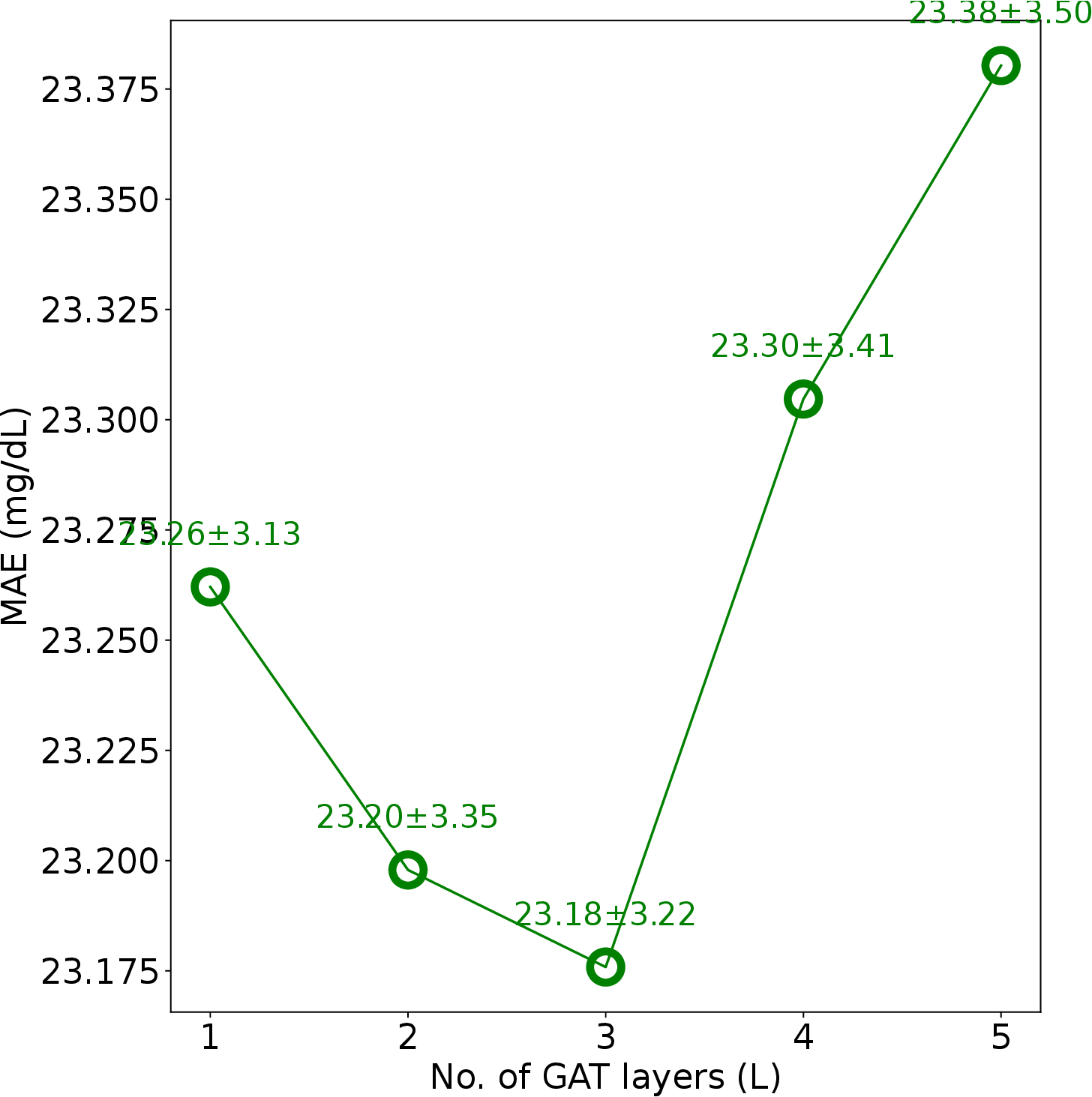}
		\end{minipage}%
		\label{Fig:12_mae_num_layers}
	}%
	\vspace{-2mm}
	\caption{The impact of changing no. of layers ($L$) when $W=12$.}
	\label{Fig:12_num_layers}
\end{figure*}

The sensitivity of no. of heads $M$ can be viewed in Figures \ref{Fig:6_nheads}-\ref{Fig:12_nheads}.
We can draw a conclusion that more heads cannot enable better performance in this scenario.
Meanwhile, from Figure \ref{Fig:6_num_layers}-\ref{Fig:12_num_layers}, within certain ranges, increasing $L$ only brings minor improvements.
For example, in terms of RMSE, when increase $L$ from 1 to 4, the average RMSE is improved from 19.02 mg/dL to 18.85 mg/dL (see Figure \ref{Fig:6_rmse_num_layers}).
Hence, we can hold the view that compared with $H$ and $T$, increasing the ability of GAT, i.e., $M$ and $L$, cannot bring about obvious improvements in this scenario.
This is because the correlations among these attributes are not very complex, so it does not need a very powerful GAT.
More powerful GAT to model simple correlations might cause overfitting, increasing RMSE, MARD and MAE.

In summary, temporal hyperparameters $H$ and $T$ are relatively more important in this scenario.
They control the storing ability and the amount of historical information infused into the neural networks.
They can significantly affect the model's performance.
However, increasing the hyperparameters, i.e., $M$ and $L$, of GAT cannot obviously improve the prediction accuracy, and it may even worsen the performance in some cases since the correlations among the attribute are not very complex.

\section{Impact of FL}
\label{sec:imp_fl}
\subsection{Experiment Settings}
\label{sec:fl_setting}
Introducing FL can keep the privacy of participants in a great manner.
However, it may sacrifice some prediction accuracy.
This experiment tries to figure out how much prediction ability is lost when considering FL in the training process. 
Hence, we directly compare the performance of ``GAM + 6 attributes'' with ``GAM + 6 attributes + FL''.
In order to show that the reduction in performance is not caused by our proposed model when incorporating FL.
We also introduce LSTM \citep{DBLP:journals/neco/HochreiterS97} and use the same training algorithm to figure out whether leveraging FL will worsen the model performance.
Hence, we also compare the performance of ``LSTM + 6 attributes'' with ``LSTM + 6 attributes + FL''.
Besides, we leverage LSTM to show whether using FL will reduce the bonus of other attributes excluding ``glucose\_level''.
Therefore, we need to compare the performance of ``LSTM + glucose\_level + FL'' with ``LSTM + 6 attributes + FL''.

The reason why we introduce LSTM rather than other methods for comparisons is that LSTM is relatively more lightweight, which is suitable and applicable to mobile devices.
Given that the target of FL is to leverage the computational resources in mobile devices and to utilize the data generated in the mobile devices directly, how many computational resources the model need should be concerned carefully. This is why we did not use N-BEATS \citep{DBLP:conf/iclr/OreshkinCCB20} based method for comparisons, even though an N-BEATS based method \citep{RubinFalconeFW20} achieved first place in the BGLP 2020 challenge, but N-BEATS based method needs very deep layers to improve the performance. On the contrary, only one layer of LSTM can still achieve a satisfying performance \citep{DBLP:conf/ecai/BevanC20}.

LSTM is a self-gated recurrent neural network, as:
\begin{subequations}
\label{Eqn:LSTM}
\begin{align}
    \label{Eqn:LSTM_1}
    & \bi_t = \sigma(\bW_{xi} \bx_t + b_{xi} + \bW_{hi}\bh_{t-1} + b_{hi}), \\
    \label{Eqn:LSTM_2}
    & \bff_t = \sigma(\bW_{xf}\bx_t + b_{xf} + \bW_{hf}\bh_{t-1} + b_{hf}), \\
    \label{Eqn:LSTM_3}
    & \bj_t = \tanh(\bW_{xj}\bx_t + b_{xj} + \bW_{hj}\bh_{t-1} + b_{hj}), \\
    \label{Eqn:LSTM_4}
    & \bo_t = \sigma(\bW_{xo}\bx_t + b_{xo} + \bW_{ho}\bh_{t-1} + b_{ho}), \\
    \label{Eqn:LSTM_5}
    & \bc_t = \bff_t * \bc_{t-1} + \bi_t * \bj_t, \\
    \label{Eqn:LSTM_6}
    & \bh_t = \bo_t * \tanh(\bc_t), \ \ \ \ \forall t\in\{1, ..., T\},
\end{align}
\end{subequations}
where $\bW_{xi}, \bW_{xf}, \bW_{xj}, \bW_{xo} \in\mathbb{R}^{E\times H}$ and $\bW_{hi}, \bW_{hf}, \bW_{hj}, \bW_{ho} \in\mathbb{R}^{H\times H}$ are learnable weights; 
$\bx_t$ is defined in section \ref{sec:problem_def}.
First of all, the current input $\bx_t$ and previous hidden state $\bh_{t-1}$ will be mapped to an input gate $\bi_t$, forget gate $\bff_t$, new information $\bj_t$ and output gate $\bo_t$, as Equation (\ref{Eqn:LSTM_1})-(\ref{Eqn:LSTM_4}).
In Equation (\ref{Eqn:LSTM_5}), when forget gate $\bff_t$ is activated ($\bff_t$ is close to $\mathbf{0}$), all historical information of the previous cell state $\bc_{t-1}$ is ignored, and it is not passed to the current cell state $\bc_t$. 
When the input gate $\bi_t$ is activated ($\bi_t$ is close to $\mathbf{1}$), all the new information of $\bj_t$ can be infused to the current cell state $\bc_t$.
Hence, with the help of the input gate $\bi_t$ and forget gate $\bff_t$, LSTM can decide how much historical information and new information should be aggregated to the current cell state $\bc_t$.
Comparably, the output gate $\bo_t$ can decide how much information of the current cell state $\bc_t$ can be passed to the current hidden state $\bh_t$.
The self-gated mechanism helps LSTM to avoid gradient descent problems by leveraging an accumulator $\bc_t$.
Finally, we use Equation (\ref{eqn:model_5}) to estimate the future blood glucose level $\hat{y}_{T+W}$.

\textbf{Hyperparameter settings}: prediction window is $W=6$ or $W=12$; historical sequence length $T=48$; both GRU and LSTM hidden size are $H=512$; no. of heads $M=1$; no. of GAT layers $L=2$. The training algorithm of ``GAM + 6 attributes'' and ``LSTM + 6 attributes'' are based on Algorithm \ref{Alg:training_no_fl}, and no. of global training epochs in the first or second training step is $T^{global}=15,000$ or $T^{person}=1,600$; evaluation interval in the first or second training step is $T^{eval1}=1,000$ or $T^{eval2}=160$; batch size is $B=128$; learning rate in the first or second training step is $0.001$ or $0.00001$. 

The training algorithm of ``GAM + 6 attributes + FL'', ``LSTM + 6 attributes + FL'' and ``LSTM + 6 attributes + FL'' are according to Algorithm \ref{Alg:training_with_fl}, and no. of FL training epochs in the first training step (FL) are $T^{total}=50$ and $T^{client}=80$; no. of FL training epochs in the second training step is $T^{person}=1,600$; evaluation interval in the first or second training step is $T^{eval1}=2$ or $T^{eval2}=160$; batch size is $B=128$; learning rate in the first or second training step is $0.001$ or $0.00001$.
\subsection{Result Analysis}
\label{sec:fl_res}

In Tables \ref{table:6_rmse_fl}-\ref{table:12_MAE_fl}, we can find that ``GAT + 6 attributes'' is consistently much better than ``GAT + 6 attributes + FL''.
For example, when $W=6$, the average RMSE of ``GAT + 6 attributes'' is 18.753 mg/dL, while the average RMSE of ``GAT + 6 attributes + FL'' is 20.489 mg/dL (see Table \ref{table:6_rmse_fl}).
We will have a similar observation when comparing ``LSTM + 6 attributes'' with ``LSTM + 6 attributes + FL''.
For instance, the average RMSE of ``LSTM + 6 attributes'' is 19.333 mg/dL when $W=6$, while the average RMSE of ``LSTM + 6 attributes + FL'' is 20.633 mg/dL.
Accordingly, we can conclude that introducing FL sacrifices some prediction accuracy, and it is not caused by the unstable of our proposed model.

Furthermore, ``GAM + 6 attributes'' is consistently better than ``LSTM + 6 attributes''. 
``GAM + 6 attributes + FL'' is better than ``LSTM + 6 attributes + FL'' in most cases.
It shows that our proposed model, GAM, is better than LSTM.
Besides, when comparing ``LSTM + 6 attributes + FL'' with ``LSTM + glucose\_level + FL'', there is a minor improvement when considering other attributes, excluding ``glucose\_level'', in FL.
We suppose that FL worsens the model performance by reducing the benefits of considering other attributes.

\begin{table}[tb]
\centering
\caption{In terms of RMSE (mg/dL), the impact of FL ($W=6$), where 6 attributes are ``glucose\_level'', ``meal'', ``bolus'', ``finger\_stick'', ``sleep'' and ``exercise''.}
\resizebox{1.05\textwidth}{!}{
    \begin{tabular}{lccccccccccccc}
    \hline
    \multicolumn{1}{c}{Method} & 559    & 563    & 570    & 575    & 588    & 591    & 540    & 544    & 552    & 567    & 584    & 596    & Average \\ \hline
    LSTM + glucose\_level + FL & 20.357 & 18.556 & 19.789 & 23.682 & 18.912 & 22.589 & 23.261 & 18.332 & 17.746 & 23.366 & 23.876 & 18.312 & 20.732  \\
    LSTM + 6 attributes + FL   & 20.590 & 18.654 & 19.495 & 26.110 & 18.139 & 22.079 & 21.990 & 18.083 & 17.059 & 22.931 & 24.275 & 18.195 & 20.633  \\
    LSTM + 6 attributes        & 18.664 & 18.117 & 17.166 & 23.804 & 17.083 & 21.130 & 21.568 & 16.512 & 16.842 & 21.969 & 22.372 & 16.765 & 19.333  \\ \hline
    GAM + 6 attributes + FL    & 20.156 & 18.589 & 18.988 & 25.059 & 17.890 & 22.030 & 21.960 & 17.376 & 17.543 & 23.269 & 24.708 & 18.300 & 20.489  \\
    GAM + 6 attributes         & 19.222 & 17.926 & 16.337 & 21.677 & 17.010 & 20.833 & 20.816 & 15.784 & 16.463 & 21.619 & 21.486 & 15.864 & 18.753  \\ \hline
    \end{tabular}
}
\label{table:6_rmse_fl}
\end{table}

\begin{table}[tb]
\centering
\caption{In terms of MARD (\%), the impact of FL ($W=6$), where 6 attributes are ``glucose\_level'', ``meal'', ``bolus'', ``finger\_stick'', ``sleep'' and ``exercise''.}
\resizebox{1.05\textwidth}{!}{
    \begin{tabular}{lccccccccccccc}
    \hline
    \multicolumn{1}{c}{Method} & 559   & 563   & 570   & 575    & 588   & 591    & 540    & 544   & 552    & 567    & 584    & 596   & Average \\ \hline
    LSTM + glucose\_level + FL & 9.404 & 8.086 & 6.124 & 10.747 & 8.306 & 12.676 & 11.730 & 8.882 & 10.002 & 11.594 & 11.625 & 9.427 & 9.883   \\
    LSTM + 6 attributes + FL   & 9.150 & 8.101 & 6.046 & 11.297 & 7.942 & 12.601 & 11.133 & 8.125 & 9.395  & 11.399 & 11.938 & 9.368 & 9.708   \\
    LSTM + 6 attributes        & 8.479 & 7.856 & 5.456 & 10.240 & 7.452 & 12.109 & 10.865 & 7.664 & 9.219  & 10.765 & 10.809 & 8.724 & 9.136   \\ \hline
    GAM + 6 attributes + FL    & 9.121 & 8.099 & 5.901 & 10.904 & 7.842 & 12.356 & 10.991 & 8.050 & 9.836  & 12.064 & 12.181 & 9.333 & 9.723   \\
    GAM + 6 attributes         & 8.709 & 7.700 & 5.267 & 9.330  & 7.387 & 11.853 & 10.412 & 7.437 & 9.077  & 10.746 & 10.459 & 8.343 & 8.893   \\ \hline
    \end{tabular}
}
\label{table:6_mard_fl}
\end{table}

\begin{table}[tb]
\centering
\caption{In terms of MAE (mg/dL), the impact of FL ($W=6$), where 6 attributes are ``glucose\_level'', ``meal'', ``bolus'', ``finger\_stick'', ``sleep'' and ``exercise''.}
\resizebox{1.05\textwidth}{!}{
    \begin{tabular}{lccccccccccccc}
    \hline
    \multicolumn{1}{c}{Method} & 559    & 563    & 570    & 575    & 588    & 591    & 540    & 544    & 552    & 567    & 584    & 596    & Average \\ \hline
    LSTM + glucose\_level + FL & 14.054 & 12.900 & 12.430 & 15.391 & 13.781 & 15.995 & 16.889 & 13.091 & 13.008 & 15.718 & 16.951 & 12.852 & 14.422  \\
    LSTM + 6 attributes + FL   & 13.897 & 12.946 & 12.412 & 16.681 & 13.222 & 15.658 & 15.909 & 12.358 & 12.361 & 15.415 & 17.473 & 12.569 & 14.242  \\
    LSTM + 6 attributes        & 12.662 & 12.612 & 11.011 & 14.880 & 12.426 & 15.091 & 15.689 & 11.680 & 12.129 & 14.644 & 16.024 & 11.649 & 13.375  \\ \hline
    GAM + 6 attributes + FL    & 13.826 & 12.918 & 11.929 & 15.989 & 13.034 & 15.560 & 15.835 & 12.005 & 12.891 & 16.072 & 17.805 & 12.599 & 14.205  \\
    GAM + 6 attributes         & 12.921 & 12.449 & 10.713 & 13.499 & 12.452 & 14.915 & 14.995 & 11.312 & 11.963 & 14.600 & 15.415 & 11.100 & 13.028  \\ \hline
    \end{tabular}
}
\label{table:6_MAE_fl}
\end{table}

\begin{table}[tb]
\centering
\caption{In terms of RMSE (mg/dL), the impact of FL ($W=12$), where 6 attributes are ``glucose\_level'', ``meal'', ``bolus'', ``finger\_stick'', ``sleep'' and ``exercise''.}
\resizebox{1.05\textwidth}{!}{
    \begin{tabular}{lccccccccccccc}
    \hline
    \multicolumn{1}{c}{Method} & 559    & 563    & 570    & 575    & 588    & 591    & 540    & 544    & 552    & 567    & 584    & 596    & Average \\ \hline
    LSTM + glucose\_level + FL & 34.438 & 30.795 & 32.335 & 37.425 & 31.505 & 34.621 & 43.097 & 31.124 & 30.504 & 38.700 & 38.626 & 29.905 & 34.423  \\
    LSTM + 6 attributes + FL   & 34.052 & 30.581 & 30.928 & 39.173 & 30.774 & 33.938 & 41.058 & 31.602 & 32.694 & 38.883 & 38.722 & 29.322 & 34.310  \\
    LSTM + 6 attributes        & 31.428 & 29.078 & 29.238 & 34.465 & 29.380 & 33.414 & 38.641 & 28.268 & 29.433 & 37.275 & 36.547 & 27.067 & 32.019  \\ \hline
    GAM + 6 attributes + FL    & 35.751 & 29.722 & 30.567 & 39.775 & 29.666 & 34.028 & 40.754 & 28.391 & 30.598 & 39.585 & 40.051 & 28.198 & 33.924  \\
    GAM + 6 attributes         & 31.616 & 29.578 & 26.776 & 33.806 & 28.416 & 32.911 & 38.348 & 26.916 & 28.410 & 37.073 & 36.098 & 25.352 & 31.275  \\ \hline
    \end{tabular}
}
\label{table:12_rmse_fl}
\end{table}

\begin{table}[tb]
\centering
\caption{In terms of MARD (\%), the impact of FL ($W=12$), where 6 attributes are ``glucose\_level'', ``meal'', ``bolus'', ``finger\_stick'', ``sleep'' and ``exercise''.}
\resizebox{1.05\textwidth}{!}{

    \begin{tabular}{lccccccccccccc}
    \hline
    \multicolumn{1}{c}{Method} & 559    & 563    & 570    & 575    & 588    & 591    & 540    & 544    & 552    & 567    & 584    & 596    & Average \\ \hline
    LSTM + glucose\_level + FL & 17.770 & 13.860 & 11.030 & 18.578 & 13.900 & 21.424 & 21.996 & 16.689 & 17.952 & 21.872 & 19.849 & 16.591 & 17.626  \\
    LSTM + 6 attributes + FL   & 17.254 & 14.113 & 10.477 & 19.025 & 13.426 & 21.120 & 21.346 & 14.558 & 19.185 & 22.008 & 20.299 & 15.578 & 17.366  \\
    LSTM + 6 attributes        & 16.230 & 12.922 & 10.287 & 16.981 & 12.654 & 20.850 & 20.526 & 13.981 & 17.013 & 21.051 & 18.640 & 14.930 & 16.339  \\ \hline
    GAM + 6 attributes + FL    & 17.473 & 13.202 & 10.335 & 19.321 & 13.012 & 21.737 & 21.345 & 13.360 & 18.559 & 22.629 & 21.054 & 15.740 & 17.314  \\
    GAM + 6 attributes         & 15.658 & 12.857 & 9.819  & 16.701 & 12.166 & 20.668 & 19.812 & 12.739 & 16.994 & 20.837 & 18.759 & 13.822 & 15.903  \\ \hline
    \end{tabular}

}
\label{table:12_mard_fl}
\end{table}

\begin{table}[tb]
\centering
\caption{In terms of MAE (mg/dL), the impact of FL ($W=12$), where 6 attributes are ``glucose\_level'', ``meal'', ``bolus'', ``finger\_stick'', ``sleep'' and ``exercise''.}
\resizebox{1.05\textwidth}{!}{
    \begin{tabular}{lccccccccccccc}
    \hline
    \multicolumn{1}{c}{Method} & 559    & 563    & 570    & 575    & 588    & 591    & 540    & 544    & 552    & 567    & 584    & 596    & Average \\ \hline
    LSTM + glucose\_level + FL & 25.717 & 22.508 & 22.347 & 26.522 & 23.092 & 26.257 & 32.182 & 24.032 & 23.046 & 28.775 & 28.989 & 22.148 & 25.468  \\
    LSTM + 6 attributes + FL   & 25.169 & 22.593 & 21.394 & 27.618 & 22.764 & 25.474 & 30.626 & 22.097 & 24.849 & 28.600 & 29.226 & 21.020 & 25.119  \\
    LSTM + 6 attributes        & 23.063 & 21.018 & 20.527 & 23.997 & 21.381 & 25.074 & 29.473 & 21.077 & 22.309 & 27.453 & 27.281 & 19.880 & 23.544  \\ \hline
    GAM + 6 attributes + FL    & 26.123 & 21.363 & 20.888 & 27.969 & 21.991 & 25.761 & 31.136 & 20.321 & 23.648 & 29.232 & 30.220 & 20.934 & 24.966  \\
    GAM + 6 attributes         & 22.763 & 20.872 & 19.400 & 23.913 & 20.883 & 25.008 & 28.947 & 19.523 & 21.853 & 27.455 & 27.161 & 18.501 & 23.023  \\ \hline
    \end{tabular}
}
\label{table:12_MAE_fl}
\end{table}

\section{Impact of Time-Aware Attention}
\label{sec:imp_ta}

\subsection{Experiment Settings}
\label{sec:ta_setting}

According to the results of section \ref{sec:attri_res}, we can find that temporal features, i.e., ``day\_of\_week'' and ``total\_seconds'' cannot obviously improve the performance of our model.
However, in section \ref{sec:imp_of_tf}, we can find some apparent temporal features which are relevant to the variations of ``glucose\_level'', when we analyze the processed ``glucose\_level'' data.
Given that we use zero padding to make adjacent tensors, i.e., $\bx_t$ and $\bx_{t+1}$, of RMTS have equal time intervals, i.e., 5 minutes.
Hence, maybe our proposed model can infer some temporal features from RMTS directly, which is already enough for the predictions, and no more temporal features are needed.
For example, given $[\bx_1, ..., \bx_{12}]$, the model can infer that the total period of this RMTS is 1 hour, and the model can also infer the time interval between any $\bx_t$ ($t>1$) and $\bx_1$.
Maybe the inference of temporal features is enough, or our proposed model cannot leverage temporal features.

In order to find the answers to the above questions, we firstly compare ``LSTM + 6 attributes'' with ``LSTM + 6 attributes + total\_seconds''.
If ``LSTM + 6 attributes + total\_seconds'' is also worse than ``LSTM + 6 attributes'', which is the same as the observation in section \ref{sec:attri_res}, we can hold the view that bad performance caused by introducing temporal features is not the problem of our proposed model.
Otherwise, the problem may be attributed to our proposed model.
On the other hand, we want to add time-aware attention (TA, \cite{DBLP:journals/tkde/LiuWPDYWW22}) to GAM so as to find whether the temporal features can be utilized by introducing a new module.
Therefore, we need to compare ``GAT + 6 attributes'' with ``GAT + TA + 6 attributes + total\_seconds'', and the time-aware attention is as follows ($t, t^{\prime}\in\mathcal{T}$):
\begin{subequations}
\label{Eqn:ta}
\begin{align}
    \label{Eqn:ta_1}
    &\beta_t = \frac{\exp({\rm f}^{time}(d_t)^\mathsf{T}\bW^{TA}{\rm f}^{time}(d_{T+W}))}{\sum_{t^{\prime}}\exp({\rm f}^{time}(d_{t^{\prime}})^\mathsf{T}\bW^{TA}{\rm f}^{time}(d_{T+W}))}, \ \forall t\in \mathcal{T}\\
    \label{Eqn:ta_2}    
    &\bh_T^{\prime} = \sum_t \beta_t \bh_t,
\end{align}
\end{subequations}
where $d_t\in\mathbb{R}^1$ is the ``total\_seconds'' of the time slot $t$, and ``total\_seconds'' is explained in section \ref{sec:attri_setting};
${\rm f}^{time}(\cdot)$ maps a scalar $d_t$ to a tensor, as ${\rm f}^{time}(d_t)\in\mathbb{R}^{E^{\prime}}$;
$\bW^{TA}\in\mathbb{R}^{E^\prime \times E^\prime}$ is learnable parameters; 
$\bh_t$ is the hidden state of Equation (\ref{eqn:model_4}).

${\rm f}^{time}(d_t)^\mathsf{T}\bW^{TA}{\rm f}^{time}(d_{T+W})$ calculates a similarity score between $d_t$ and $d_{T+W}$, where $\forall t\in\mathcal{T}$.
The normalized score is the time-aware attention weight $\beta_t$ (see Equation \ref{Eqn:ta_1}).
In terms of time slot $t$, if the temporal feature of $t$ is more relevant with $T+W$, the time-aware attention weight $\beta_t$ is larger.
Then, more information of $\bh_t$ is passed to $\bh_T^{\prime}$ (see Equation \ref{Eqn:ta_2}).
This mechanism enables the estimation to leverage the information from $\{\bh_1, ..., \bh_t,..., \bh_T\}$ with the time-aware attention weights rather than only utilizing $\bh_T$.
The focus of the time-aware attention weights is decided by the temporal features of the target time slot, i.e., $T+W$.
Compared with only using the information of $\bh_T$, this mechanism may further leverage some distant historical information which has been forgotten by the GRU memory.
In this way, GAM can explicitly leverage temporal features.

\textbf{Hyperparameter settings}: prediction window is $W=6$ or $W=12$; historical sequence length $T=48$; both GRU and LSTM hidden size are $H=512$; no. of heads $M=1$; no. of GAT layers $L=2$. The training algorithm is based on Algorithm \ref{Alg:training_no_fl}, and no. of global training epochs in the first or second training step is $T^{global}=15,000$ or $T^{person}=1,600$; evaluation interval in the first or second training step is $T^{eval1}=1,000$ or $T^{eval2}=160$; batch size is $B=128$; learning rate in the first or second training step is $0.001$ or $0.00001$.

\subsection{Result Analysis}
\label{sec:ta_res}

\begin{table}[tb]
\centering
\caption{In terms of RMSE (mg/dL), the impact of TA ($W=6$), where 6 attributes are ``glucose\_level'', ``meal'', ``bolus'', ``finger\_stick'', ``sleep'' and ``exercise''.}
\resizebox{1.05\textwidth}{!}{
    \begin{tabular}{lccccccccccccc}
    \hline
    \multicolumn{1}{c}{Method}               & 559    & 563    & 570    & 575    & 588    & 591    & 540    & 544    & 552    & 567    & 584    & 596    & Average \\ \hline
    LSTM + 6 attributes + total\_seconds     & 18.555 & 17.606 & 17.496 & 22.419 & 17.075 & 20.720 & 21.778 & 16.292 & 16.594 & 21.860 & 22.538 & 16.359 & 19.108  \\
    GAM + TA + 6 attributes + total\_seconds & 19.704 & 20.303 & 17.941 & 22.811 & 18.176 & 21.580 & 23.184 & 16.859 & 17.311 & 22.856 & 23.032 & 17.544 & 20.108  \\
    LSTM + 6 attributes                      & 18.664 & 18.117 & 17.166 & 23.804 & 17.083 & 21.130 & 21.568 & 16.512 & 16.842 & 21.969 & 22.372 & 16.765 & 19.333  \\
    GAM + 6 attributes                       & 19.222 & 17.926 & 16.337 & 21.677 & 17.010 & 20.833 & 20.816 & 15.784 & 16.463 & 21.619 & 21.486 & 15.864 & 18.753  \\ \hline
    \end{tabular}
}
\label{table:6_rmse_ta}
\end{table}

\begin{table}[tb]
\centering
\caption{In terms of MARD (\%), the impact of TA ($W=6$), where 6 attributes are ``glucose\_level'', ``meal'', ``bolus'', ``finger\_stick'', ``sleep'' and ``exercise''.}
\resizebox{1.05\textwidth}{!}{
    \begin{tabular}{lccccccccccccc}
    \hline
    \multicolumn{1}{c}{Method}                 & 559   & 563   & 570   & 575    & 588   & 591    & 540    & 544   & 552   & 567    & 584    & 596   & Average \\ \hline
    LSTM + 6 attributes +   total\_seconds     & 8.370 & 7.745 & 5.545 & 9.852  & 7.349 & 12.008 & 10.706 & 7.579 & 8.962 & 10.796 & 10.909 & 8.622 & 9.037   \\
    GAM + TA + 6 attributes +   total\_seconds & 9.162 & 8.656 & 6.137 & 10.300 & 7.768 & 12.375 & 11.881 & 7.896 & 9.737 & 12.012 & 11.513 & 9.202 & 9.720   \\
    LSTM + 6 attributes                        & 8.479 & 7.856 & 5.456 & 10.240 & 7.452 & 12.109 & 10.865 & 7.664 & 9.219 & 10.765 & 10.809 & 8.724 & 9.136   \\
    GAM + 6 attributes                         & 8.709 & 7.700 & 5.267 & 9.330  & 7.387 & 11.853 & 10.412 & 7.437 & 9.077 & 10.746 & 10.459 & 8.343 & 8.893   \\ \hline
    \end{tabular}
}
\label{table:6_mard_ta}
\end{table}

\begin{table}[tb]
\centering
\caption{In terms of MAE (mg/dL), the impact of TA ($W=6$), where 6 attributes are ``glucose\_level'', ``meal'', ``bolus'', ``finger\_stick'', ``sleep'' and ``exercise''.}
\resizebox{1.05\textwidth}{!}{
    \begin{tabular}{lccccccccccccc}
    \hline
    \multicolumn{1}{c}{Method}                 & 559    & 563    & 570    & 575    & 588    & 591    & 540    & 544    & 552    & 567    & 584    & 596    & Average \\ \hline
    LSTM + 6 attributes +   total\_seconds     & 12.675 & 12.421 & 11.164 & 14.313 & 12.326 & 14.915 & 15.619 & 11.524 & 11.851 & 14.644 & 16.097 & 11.431 & 13.248  \\
    GAM + TA + 6 attributes +   total\_seconds & 13.759 & 14.214 & 12.058 & 14.695 & 13.316 & 15.455 & 17.169 & 11.948 & 12.692 & 15.962 & 16.688 & 12.219 & 14.181  \\
    LSTM + 6 attributes                        & 12.662 & 12.612 & 11.011 & 14.880 & 12.426 & 15.091 & 15.689 & 11.680 & 12.129 & 14.644 & 16.024 & 11.649 & 13.375  \\
    GAM + 6 attributes                         & 12.921 & 12.449 & 10.713 & 13.499 & 12.452 & 14.915 & 14.995 & 11.312 & 11.963 & 14.600 & 15.415 & 11.100 & 13.028  \\ \hline
    \end{tabular}
}
\label{table:6_mae_ta}
\end{table}

\begin{table}[tb]
\centering
\caption{In terms of RMSE (mg/dL), the impact of TA ($W=12$), where 6 attributes are ``glucose\_level'', ``meal'', ``bolus'', ``finger\_stick'', ``sleep'' and ``exercise''.}
\resizebox{1.05\textwidth}{!}{
    \begin{tabular}{lccccccccccccc}
    \hline
    \multicolumn{1}{c}{Method}             & 559    & 563    & 570    & 575    & 588    & 591    & 540    & 544    & 552    & 567    & 584    & 596    & Average \\ \hline
    LSTM + 6 attributes +   total\_seconds & 32.550 & 29.456 & 29.255 & 34.727 & 29.044 & 32.738 & 38.774 & 27.057 & 29.716 & 36.582 & 36.972 & 27.267 & 32.011  \\
    GAM + TA + 6 attributes +   total\_seconds  & 34.653 & 33.940 & 29.479 & 33.819 & 32.195 & 34.490 & 43.031 & 29.283 & 30.606 & 39.103 & 38.528 & 27.596 & 33.893  \\
    LSTM + 6 attributes                    & 31.428 & 29.078 & 29.238 & 34.465 & 29.380 & 33.414 & 38.641 & 28.268 & 29.433 & 37.275 & 36.547 & 27.067 & 32.019  \\
    GAM + 6 attributes                     & 31.616 & 29.578 & 26.776 & 33.806 & 28.416 & 32.911 & 38.348 & 26.916 & 28.410 & 37.073 & 36.098 & 25.352 & 31.275  \\ \hline
    \end{tabular}
}
\label{table:12_rmse_ta}
\end{table}

\begin{table}[tb]
\centering
\caption{In terms of MARD (\%), the impact of TA ($W=12$), where 6 attributes are ``glucose\_level'', ``meal'', ``bolus'', ``finger\_stick'', ``sleep'' and ``exercise''.}
\resizebox{1.05\textwidth}{!}{
    \begin{tabular}{lccccccccccccc}
    \hline
    \multicolumn{1}{c}{Method}             & 559    & 563    & 570    & 575    & 588    & 591    & 540    & 544    & 552    & 567    & 584    & 596    & Average \\ \hline
    LSTM + 6 attributes +   total\_seconds & 16.403 & 13.237 & 10.164 & 17.072 & 12.663 & 20.650 & 19.974 & 13.109 & 17.546 & 19.933 & 18.955 & 14.903 & 16.217  \\
    GAM + TA + 6 attributes +   total\_seconds  & 17.394 & 14.533 & 10.963 & 17.814 & 13.949 & 22.338 & 22.278 & 13.673 & 18.591 & 22.813 & 20.167 & 15.176 & 17.474  \\
    LSTM + 6 attributes                    & 16.230 & 12.922 & 10.287 & 16.981 & 12.654 & 20.850 & 20.526 & 13.981 & 17.013 & 21.051 & 18.640 & 14.930 & 16.339  \\
    GAM + 6 attributes                     & 15.658 & 12.857 & 9.819  & 16.701 & 12.166 & 20.668 & 19.812 & 12.739 & 16.994 & 20.837 & 18.759 & 13.822 & 15.903  \\ \hline
    \end{tabular}
}
\label{table:12_mard_ta}
\end{table}

\begin{table}[tb]
\centering
\caption{In terms of MAE (mg/dL), the impact of TA ($W=12$), where 6 attributes are ``glucose\_level'', ``meal'', ``bolus'', ``finger\_stick'', ``sleep'' and ``exercise''.}
\resizebox{1.05\textwidth}{!}{
    \begin{tabular}{lccccccccccccc}
    \hline
    \multicolumn{1}{c}{Method}             & 559    & 563    & 570    & 575    & 588    & 591    & 540    & 544    & 552    & 567    & 584    & 596    & Average \\ \hline
    LSTM + 6 attributes +   total\_seconds & 24.249 & 21.358 & 20.127 & 24.358 & 21.278 & 24.742 & 29.211 & 19.918 & 22.572 & 26.254 & 27.626 & 19.823 & 23.460  \\
    GAM + TA + 6 attributes +   total\_seconds  & 25.543 & 24.167 & 21.509 & 24.517 & 23.782 & 26.610 & 32.683 & 21.115 & 23.899 & 29.167 & 29.235 & 20.162 & 25.199  \\
    LSTM + 6 attributes                    & 23.063 & 21.018 & 20.527 & 23.997 & 21.381 & 25.074 & 29.473 & 21.077 & 22.309 & 27.453 & 27.281 & 19.880 & 23.544  \\
    GAM + 6 attributes                     & 22.763 & 20.872 & 19.400 & 23.913 & 20.883 & 25.008 & 28.947 & 19.523 & 21.853 & 27.455 & 27.161 & 18.501 & 23.023  \\ \hline
    \end{tabular}
}
\label{table:12_mae_ta}
\end{table}

From Tables \ref{table:6_rmse_ta}-\ref{table:12_mae_ta}, we have some observations.
Compared with ``LSTM + 6 attributes'', after including temporal features, ``LSTM + 6 attributes + total\_seconds'' is consistently better.
For example, in Table \ref{table:6_rmse_ta}, the average RMSE of ``LSTM + 6 attributes + total\_seconds'' is 19.108 mg/dL, while the average RMSE of ``LSTM + 6 attributes'' is 20.108 mg/dL.
In Table \ref{table:12_rmse_ta}, the average RMSE of ``LSTM + 6 attributes + total\_seconds'' is 32.011 mg/dL, while the average RMSE of ``LSTM + 6 attributes'' is 33.893 mg/dL.
Hence, introducing temporal features do helpful for the BGLP-RMTS.

However, even though the temporal features can positively affect the prediction tasks, our proposed model still failed to leverage temporal features (see section \ref{sec:attri_res}).
In this experiment, when explicitly utilizing temporal features by adding time-aware attention to GAM, ``GAM + TA + 6 attribute + total\_seconds'' is consistently worse than ``GAM + 6 attribute''.
For example, in Table \ref{table:6_rmse_ta}, the average RMSE of ``GAM + TA + 6 attribute + total\_seconds'' is 19.333 mg/dL, while the average RMSE of ``GAM + 6 attribute'' is 18.753 mg/dL.
In Table \ref{table:12_rmse_ta}, the average RMSE of ``GAM + TA + 6 attribute + total\_seconds'' is 32.019 mg/dL, while the average RMSE of ``GAM + 6 attribute'' is 31.275 mg/dL.
Explicitly modeling temporal features by GAM still cannot improve the performance.

Hence, according to the above observations and some guesses (see the first paragraph of section \ref{sec:ta_setting}), we further guess that maybe GAM can extract temporal features directly from the RMTS without ``total\_seconds'', and more extra temporal features can be seen as redundancy, which might negatively affect the prediction performance.

\section{Answers to the Research Questions}
\label{sec:ans}

The research questions raised from the research proposal are answered as follows.

Question 1: can the correlations among multiple variables, e.g. glucose level, meal, insulin, sleep, etc., be modeled by graph neural networks?

The answer to Question 1: according to the observations and conclusions of section \ref{sec:attri_res}, we find that there are 6 attributes, i.e., ``glucose\_level'', ``meal'', ``bolus'', ``finger\_stick'', ``sleep'' and ``exercise'' are helpful for the BGLP-RMTS when leveraging graph attention based sequential neural network, i.e., GAM.
Besides, in section \ref{sec:fl_res}, we can obviously observe the benefits when using a graph-based structure compared with a purely LSTM-based method. 
However, in section \ref{sec:hyper_res}, we can find that the benefits are limited because we guess the correlations among these attributes are not very complex, and the graph structure should not be very sophisticated, i.e., very large no. of heads $M$ or layers $L$.

Question 2: can a time-aware attention mechanism directly deal with time series whose adjacent items have different time intervals?

The answer to Question 2: first of all, we transform IMTS into RMTS via zero padding. Then, based on the results of section \ref{sec:ta_res}, we find that a time-aware attention mechanism cannot further improve the performance of GAM when incorporating temporal features, i.e., ``total\_seconds''. 
This may be because that RMTS has equal time intervals between the adjacent items, and some temporal features might be directly inferred (see the first paragraph of section \ref{sec:ta_setting}).
Besides, the memory of GAM is based on GRU which is good at dealing with RMTS rather than IMTS.

Question 3: in order to protect patient privacy, can a non-personalized model be trained with fully decentralized local data aided by federated learning?

The answer to Question 3: in section \ref{sec:fl_res}, we can observe the performance of FL in two different models, i.e., an LSTM-based model and a graph-based model. 
We can find that the two models sacrifice some prediction accuracy when introducing FL.
Meanwhile, the benefits of considering more attributes are also reduced.
Note that we do not use non-personalized models but use personalized models which are generated by fine-tuning the non-personalized model after FL.
This is because we find that the body conditions and behaviors vary a lot among the participants in section \ref{sec:data_preprocessing}.
\chapter{Discussion}
\label{chap:dis}
In this chapter, we compare our proposed model with the methods in Ohio 2020 Challenge \citep{DBLP:conf/ecai/2020kdh} in section \ref{sec:comp}. 
Then, we further analyze the selected 6 attributes, figuring out whether these attributes are helpful (see section \ref{sec:dis_attri}).
Next, we visualize the GAM and testing samples, showing that our proposed model is explainable (see section \ref{sec:vis}).
Then, we discuss the convergence and training time of our proposed method (see section \ref{sec:cong}).
Finally, we make a brief conclusion, discussing the advantages and limitations of this work, and we also talk about future work and research directions in section \ref{sec:conclusion}.
\section{Compare with Other Methods}
\label{sec:comp}
\begin{table}[tb]
\centering
\caption{Compare with other methods in Ohio 2020 challenge.}
\resizebox{1.05\textwidth}{!}{
    \begin{tabular}{ccccccc}
    \hline
    Author                                                  & Method                        & RMSE (mg/dL, $W=6$) & MAE(mg/dL, $W=6$) & RMSE(mg/dL, $W=12$) & MAE(mg/dL, $W=12$) & Overall \\ \hline
    \cite{RubinFalconeFW20}                & N-BEATS                       & 18.22               & 12.83             & 31.66               & 23.60              & 86.31   \\
   \cite{DBLP:conf/ecai/HameedK20}        & RNN                           & 19.21               & 13.08             & 31.77               & 23.09              & 87.15   \\
    \textbf{Proposed model}                                          & \textbf{GAM}                           & \textbf{18.77}                & \textbf{13.39}              & \textbf{32.04}                & \textbf{23.86}               & \textbf{88.06}    \\
    \cite{DBLP:conf/ecai/ZhuYLHG20}        & GAN (GRU; CNN)                & 18.34               & 13.37             & 32.21               & 24.20              & 88.12   \\
    \cite{DBLP:conf/ecai/YangWTWMZYL20}    & Multi-Scale LSTM              & 19.05               & 13.50             & 32.03               & 23.83              & 88.41   \\
    \cite{DBLP:conf/ecai/BevanC20}         & LSTM+Attention                & 18.23               & 14.37             & 31.10               & 25.75              & 89.45   \\
    \cite{DBLP:conf/ecai/SunRSHBASC20}     & Latent-Variable-based Model   & 19.37               & 13.76             & 32.59               & 24.64              & 90.36   \\
    \cite{DBLP:conf/ecai/JoedickeKCWVCH20} & Genetic Programming           & 19.60               & 14.25             & 34.12               & 25.99              & 93.96   \\
    \cite{DBLP:conf/ecai/MaZWYWTYL20}      & Residual compensation network & 20.03               & 14.52             & 34.89               & 26.41              & 95.85   \\ \hline
    \end{tabular}
}
\label{table:comp}
\end{table}

We compare our proposed method, GAM, with the top 8 methods in Ohio 2020 Challenge (see Table \ref{table:comp}).
These methods are also briefly explained as follows.
\begin{itemize}
    \item \textbf{N-BEATS \citep{RubinFalconeFW20}}: this method was based on N-BEATS \citep{DBLP:conf/iclr/OreshkinCCB20}.
    They utilize OhioT1DM'18 (6 participants) and Tidepool data (100 participants) to pretrain the models. Then, they use 80\% of the training data of OhioT1DM'20 for training and 20\% of the training data for validation. They leverage 5 attributes for the predictions, including ``glucose\_level'', ``bolus'', ``meal'', ``finger\_stick'', ``time\_of\_day''.
    
    \item \textbf{RNN  \citep{DBLP:conf/ecai/HameedK20}}: this approach was based on an RNN.
    They use OhioT1DM'18, excluding the last 20\% of the data, to pretrain the models. Next, they use 100\% of the training data of OhioT1DM'20 for training and the last 20\% of the training data of OhioT1DM'18 for validation. They only use ``glucose\_level'' as the input.
    
    \item \textbf{GAN (GRU; CNN) \citep{DBLP:conf/ecai/ZhuYLHG20}}: this model was based on GAN.
    They utilize OhioT1DM'18 to pretrain the models. Then, they use 80\% of the training data of OhioT1DM'20 for training and 20\% of the training data for validation. They consider 3 attributes for the predictions, i.e., ``glucose\_level'', ``bolus'', and ``meal''.
    
    \item \textbf{Multi-Scale LSTM \citep{DBLP:conf/ecai/YangWTWMZYL20}}: this work was according to LSTM.
    They do not mention whether they pretrain the models. They use 90\% of the training data of OhioT1DM'20 for training and 10\% of the training data for validation. They consider 4 attributes for the predictions, and the attributes are ``glucose\_level'', ``bolus'', ``meal'' and ``basal''.
    
    \item \textbf{LSTM + Attention \citep{DBLP:conf/ecai/BevanC20}}: this work was also based on LSTM.
    They leverage OhioT1DM'18 to pretrain the models. They use 90\% of the training data of OhioT1DM'20 for training and 10\% of the training data for validation. They only consider``glucose\_level'' for the predictions. 
    
    \item \textbf{Latent-Variable-based Model \citep{DBLP:conf/ecai/SunRSHBASC20}}: this work was based on latent variables.
    They do not mention whether they pretrain the models. They use 90\% of the training data of OhioT1DM'20 for training and 10\% of the training data for validation. They consider 3 attributes for the predictions, and the attributes are ``glucose\_level'', ``bolus'' and ``basal''.
    
    \item \textbf{Genetic Programming \citep{DBLP:conf/ecai/JoedickeKCWVCH20}}: this work was according to genetic programming.
    They consider many variables, e.g., ``glucose\_level'', ``bolus'', ``bolus\_type'', ``basal'', ``basis\_gsr'' and ``basis\_skin\_temperature''. Besides, they take some future attributes, including ``bolus'' and ``basal''. The future attributes appear within $[T+1, T+W]$, because they hold the view that the values of ``bolus'' and ``basal'' are predictable for the participants.
    
    \item \textbf{Residual compensation network \citep{DBLP:conf/ecai/MaZWYWTYL20}}:  this work was based on the residual compensation network.
    They only consider ``glucose\_level'' as the input during the predictions.
\end{itemize}

In order to compare our proposed model with the baselines above, our model is also pretrained by OhioT1DM'18, followed by fine-tuning in OhioT1DM'20, based on Algorithm \ref{Alg:training_no_fl}.
We do not compare ``GAM + FL'' with the above baselines because FL obviously worsens the performance, and we mainly want to show the performance differences among these deep models.
The hyperparameters are the same as the ones in section \ref{sec:ta_setting}.

Overall, we can observe that our proposed model ranks third place.
However, it cannot prove that our proposed model is worse than the top 2 models.
This is because the N-BEATS-based model is pretrained by a super huge dataset which contains a hundred participants. In comparison, other methods are only pretrained/trained by relatively small datasets which contains 12 participants in total.
On the other hand, the N-BEATS uses some special tricks.
For example, they trained several models at a time with different hyperparameters, and they used the median of the predictions of these models as their final output.
Even though they have an excellent performance, they cannot prove that such performance is caused by the model or the tricks.
Similarly, the second method in Table \ref{table:comp} also leverages special tricks.
For instance, they use the first 80\% of data of OhioT1DM'18 to pretrain the models. Next, they use 100\% of the training data of OhioT1DM'20 for training and the last 20\% of the training data of OhioT1DM'18 for validation.
Such cross validation method is much different from others.
Different cross validations can affect the performance obviously since we can observe that 80\%-20\% cross validation methods (N-BEATS, RNN, GAM and GAN (GRU; CNN)) all perform better than 90\%-10\% cross validation methods (Multi-Scale LSTM, LSTM+Attention and Latent-Variable-based Model).
Without the help of such tricks (or other tricks that we do not find), We do not think it is reasonable that a simple RNN-based method can exceed many advanced LSTM-based models, i.e., Multi-Scale LSTM and LSTM + Attention.

Another observation is that neural networks (N-BEATS, RNN, GAM, GAN (GRU; CNN), Multi-Scale LSTM and LSTM+Attention) are better than other methods, where most of these top methods are based on neural networks.
In terms of attributes, there are 3 methods (RNN, LSTM + Attention and residual compensation network) that only use ``glucose\_level'' as the input.
Apart from the non-neural methods (residual compensation network) and the methods using special tricks (RNN), LSTM + Attention ranks the last one in these neural network-based methods.
It means considering other attributes will bring extra benefits during the prediction, which is the same as our results in section \ref{sec:attri_res}.
In terms of the neural-based methods which leverage multiple attributes, we can find that ``bolus'' and ``meal'' are very popular, which is in accord with our findings in section \ref{sec:attri_res}, as `bolus'', ``meal'' and ``finger\_stick'' can bring more benefits compared with other attributes excluding ``glucoses\_level''.

Note that among all these methods, we are the only ones who successfully leverage ``sleep'' and ``exercise'' as a part of the input, contributing to some improvements. 
We are also the only one that proposed an explainable model, which can explain the feature importance by graph attentions.
An example of the explanation will be discussed in section \ref{sec:vis}.

\section{Discussion of selected attributes}
\label{sec:dis_attri}

\begin{table}[tb]
\centering
\caption{Distributions of ``glucose\_level'' and ``finger\_stick'' in hyperglycemia, normal, hypoglycemia and all samples.}
\resizebox{1.05\textwidth}{!}{
    \begin{tabular}{ccccccccc}
    \hline
    PID & glucose\_level (hypo) & glucose\_level (normal) & glucose\_level (hyper) & glucose\_level (all) & finger\_stick (hypo) & finger\_stick (normal) & finger\_stick (hyper) & finger\_stick (all) \\ \hline
    559 & 59.90$\pm$7.76        & 124.73$\pm$30.93        & 237.74$\pm$48.41       & 167.23$\pm$69.91     & 108.36$\pm$70.97     & 123.75$\pm$56.18       & 280.47$\pm$88.02      & 194.69$\pm$108.09   \\
    563 & 62.56$\pm$6.59        & 128.42$\pm$29.14        & 215.00$\pm$32.07       & 149.80$\pm$49.75     & 75.80$\pm$32.12      & 133.51$\pm$37.98       & 228.06$\pm$50.16      & 166.32$\pm$64.47    \\
    570 & 63.00$\pm$5.06        & 133.52$\pm$30.89        & 237.76$\pm$38.81       & 192.95$\pm$64.10     & 63.38$\pm$15.13      & 132.54$\pm$35.90       & 240.72$\pm$43.86      & 196.51$\pm$69.61    \\
    575 & 58.31$\pm$8.07        & 122.56$\pm$29.29        & 228.68$\pm$43.27       & 143.34$\pm$60.40     & 67.92$\pm$25.90      & 120.24$\pm$38.98       & 246.89$\pm$70.40      & 154.67$\pm$78.95    \\
    588 & 59.81$\pm$8.64        & 136.67$\pm$27.18        & 219.32$\pm$32.55       & 166.82$\pm$50.33     & 88.56$\pm$28.70      & 131.79$\pm$35.23       & 212.74$\pm$48.11      & 153.33$\pm$55.21    \\
    591 & 58.39$\pm$9.53        & 126.44$\pm$29.71        & 222.62$\pm$33.26       & 153.75$\pm$56.94     & 68.97$\pm$18.19      & 133.84$\pm$44.21       & 233.82$\pm$41.77      & 161.75$\pm$67.97    \\
    540 & 60.18$\pm$7.65        & 120.58$\pm$30.50        & 226.76$\pm$38.71       & 141.35$\pm$58.21     & 68.24$\pm$21.77      & 123.23$\pm$34.87       & 245.74$\pm$54.35      & 155.83$\pm$73.33    \\
    544 & 63.06$\pm$5.49        & 130.55$\pm$29.28        & 229.99$\pm$41.09       & 163.44$\pm$58.87     & 69.80$\pm$10.45      & 119.33$\pm$30.75       & 238.55$\pm$45.97      & 140.12$\pm$58.23    \\
    552 & 61.70$\pm$5.71        & 124.38$\pm$29.61        & 224.30$\pm$35.86       & 145.47$\pm$54.23     & 85.25$\pm$27.48      & 127.85$\pm$32.66       & 228.85$\pm$56.39      & 150.69$\pm$59.01    \\
    567 & 58.11$\pm$7.92        & 129.12$\pm$29.74        & 227.28$\pm$40.60       & 152.46$\pm$59.97     & 52.12$\pm$12.80      & 137.99$\pm$31.98       & 231.05$\pm$48.13      & 187.09$\pm$64.75    \\
    584 & 56.96$\pm$8.83        & 139.97$\pm$26.62        & 240.41$\pm$50.25       & 188.50$\pm$65.19     & 199.67$\pm$12.22     & 167.90$\pm$36.24       & 246.32$\pm$66.97      & 213.25$\pm$66.82    \\
    596 & 62.56$\pm$6.53        & 127.31$\pm$29.09        & 217.29$\pm$30.57       & 147.33$\pm$49.52     & 134.29$\pm$48.56     & 131.86$\pm$33.78       & 197.22$\pm$41.67      & 141.35$\pm$42.74    \\ \hline
    \end{tabular}
}
\label{table:attri_hyper_hypo_gl_fg}
\end{table}

\begin{table}[tb]
\centering
\caption{Distributions of ``meal'' and ``bolus'' in hyperglycemia, normal, hypoglycemia and all samples.}
\resizebox{1.05\textwidth}{!}{
    \begin{tabular}{ccccccccc}
    \hline
    PID & meal (hypo)     & meal (normal)    & meal (hyper)     & meal (all)       & bolus (hypo)   & bolus (normal) & bolus (hyper)  & bolus (all)    \\ \hline
    559 & 25.17$\pm$12.50 & 35.19$\pm$14.90  & 37.46$\pm$17.99  & 35.55$\pm$15.95  & 4.70$\pm$1.13  & 3.44$\pm$1.62  & 4.25$\pm$1.71  & 3.92$\pm$1.71  \\
    563 & 36.50$\pm$12.48 & 29.21$\pm$15.83  & 31.47$\pm$18.28  & 29.95$\pm$16.33  & 10.90$\pm$3.25 & 6.94$\pm$4.22  & 9.16$\pm$3.73  & 8.00$\pm$4.14  \\
    570 & 76.43$\pm$38.05 & 109.75$\pm$40.05 & 104.70$\pm$43.63 & 105.70$\pm$42.15 & 2.96$\pm$1.21  & 8.17$\pm$3.78  & 7.50$\pm$3.58  & 7.62$\pm$3.74  \\
    575 & 31.94$\pm$18.88 & 41.42$\pm$23.75  & 48.54$\pm$24.55  & 40.56$\pm$23.43  & 3.15$\pm$1.61  & 4.60$\pm$2.30  & 5.68$\pm$3.10  & 4.78$\pm$2.55  \\
    588 & 25.00$\pm$14.72 & 35.69$\pm$35.74  & 25.71$\pm$14.63  & 32.73$\pm$31.37  & 4.95$\pm$0.07  & 5.00$\pm$2.35  & 3.30$\pm$1.94  & 4.31$\pm$2.33  \\
    591 & 26.84$\pm$12.78 & 30.83$\pm$13.04  & 37.05$\pm$17.44  & 31.43$\pm$14.20  & 3.32$\pm$1.15  & 3.06$\pm$1.53  & 3.39$\pm$2.10  & 3.20$\pm$1.76  \\
    540 & 58.11$\pm$36.10 & 55.13$\pm$28.49  & 54.00$\pm$32.66  & 55.32$\pm$29.37  & 4.25$\pm$2.04  & 4.14$\pm$2.51  & 3.58$\pm$2.39  & 3.99$\pm$2.48  \\
    544 & 52.20$\pm$33.77 & 75.98$\pm$38.36  & 66.33$\pm$31.67  & 73.32$\pm$37.18  & 14.13$\pm$4.04 & 12.99$\pm$4.95 & 9.45$\pm$5.45  & 12.49$\pm$5.16 \\
    552 & 39.86$\pm$23.18 & 58.76$\pm$33.35  & 36.60$\pm$28.51  & 53.23$\pm$32.88  & 5.30$\pm$0.00  & 4.31$\pm$3.88  & 3.22$\pm$2.18  & 3.84$\pm$3.29  \\
    567 & None            & 75.76$\pm$22.75  & 70.60$\pm$21.52  & 74.77$\pm$22.19  & 5.90$\pm$0.00  & 11.14$\pm$4.87 & 12.98$\pm$5.89 & 12.23$\pm$5.57 \\
    584 & 20.00$\pm$0.00  & 56.00$\pm$10.80  & 53.75$\pm$13.62  & 54.76$\pm$12.42  & None           & 7.39$\pm$2.60  & 7.53$\pm$2.95  & 7.48$\pm$2.81  \\
    596 & 30.50$\pm$11.84 & 25.40$\pm$13.39  & 21.70$\pm$15.24  & 25.28$\pm$13.60  & 3.54$\pm$1.00  & 3.02$\pm$1.49  & 3.03$\pm$1.35  & 3.03$\pm$1.44  \\ \hline
    \end{tabular}
}
\label{table:attri_hyper_hypo_ml_bl}
\end{table}

\begin{table}[tb]
\centering
\caption{Distributions of ``sleep'' and ``exercise'' in hyperglycemia, normal, hypoglycemia and all samples.}
\resizebox{1.05\textwidth}{!}{
    \begin{tabular}{ccccccccc}
    \hline
    PID & sleep (hypo)  & sleep (normal) & sleep (hyper) & sleep (all)   & exercise (hypo) & exercise (normal) & exercise (hyper) & exercise (all) \\ \hline
    559 & 1.91$\pm$0.49 & 2.26$\pm$0.54  & 2.32$\pm$0.61 & 2.25$\pm$0.56 & 5.00$\pm$0.00   & 5.18$\pm$1.06     & 5.16$\pm$0.93    & 5.17$\pm$0.99  \\
    563 & 2.05$\pm$0.22 & 2.57$\pm$0.50  & 2.74$\pm$0.44 & 2.60$\pm$0.49 & 4.22$\pm$1.00   & 4.51$\pm$1.28     & 4.72$\pm$0.59    & 4.52$\pm$1.23  \\
    570 & 2.09$\pm$0.29 & 2.18$\pm$0.62  & 2.10$\pm$0.43 & 2.13$\pm$0.51 & 4.89$\pm$0.31   & 5.20$\pm$0.61     & 4.79$\pm$0.87    & 5.02$\pm$0.74  \\
    575 & 1.87$\pm$0.50 & 2.74$\pm$0.48  & 2.75$\pm$0.54 & 2.71$\pm$0.53 & 7.00$\pm$0.00   & 5.45$\pm$1.02     & None             & 5.53$\pm$1.05  \\
    588 & 2.15$\pm$0.36 & 2.78$\pm$0.41  & 2.80$\pm$0.40 & 2.78$\pm$0.41 & 5.00$\pm$0.00   & 5.30$\pm$0.48     & 5.14$\pm$0.44    & 5.22$\pm$0.46  \\
    591 & 2.35$\pm$0.48 & 2.71$\pm$0.47  & 2.74$\pm$0.53 & 2.71$\pm$0.49 & 5.67$\pm$0.98   & 4.90$\pm$1.42     & 5.00$\pm$1.83    & 4.96$\pm$1.48  \\
    540 & None          & None           & None          & None          & None            & None              & None             & None           \\
    544 & 1.74$\pm$0.44 & 1.94$\pm$0.47  & 1.96$\pm$0.37 & 1.94$\pm$0.45 & None            & 6.68$\pm$0.48     & 7.00$\pm$0.00    & 6.83$\pm$0.38  \\
    552 & 3.00$\pm$0.00 & 2.64$\pm$0.64  & 2.58$\pm$0.74 & 2.64$\pm$0.65 & None            & 7.53$\pm$1.02     & None             & 7.53$\pm$1.02  \\
    567 & 2.00$\pm$0.00 & 1.90$\pm$0.86  & 2.40$\pm$0.73 & 2.01$\pm$0.77 & None            & None              & None             & None           \\
    584 & 3.00$\pm$0.00 & 2.99$\pm$0.10  & 2.96$\pm$0.20 & 2.97$\pm$0.17 & None            & 10.00$\pm$0.00    & 10.00$\pm$0.00   & 10.00$\pm$0.00 \\
    596 & 2.14$\pm$0.90 & 2.80$\pm$0.46  & 2.95$\pm$0.33 & 2.82$\pm$0.46 & 5.37$\pm$0.59   & 5.02$\pm$1.12     & 5.05$\pm$1.06    & 5.04$\pm$1.08  \\ \hline
    \end{tabular}
}
\label{table:attri_hyper_hypo_sl_ex}
\end{table}

Based on the results of section \ref{sec:attri_res}, we can find that ``meal'', ``bolus'', ``finger\_stick'', ``sleep'' and ``exercise'', apart from ``glucose\_level'', are useful during the predictions of future BG levels.
Then, we want to further explore whether we can directly observe some patterns from the statistics of these attributes.

According to Table \ref{table:attri_hyper_hypo_gl_fg}-\ref{table:attri_hyper_hypo_sl_ex}, we can observe the distributions of these attributes in terms of hypoglycemia (hypo), normal glucose levels (normal), hyperglycemia (hyper) and all samples (all).
Each attribute is aggregated through mean and standard deviation and grouped by these four statuses of ``glucose\_level''.

Then, we can find that the variations of ``finger\_stick'' have similar trends to ``glucose\_level'' (see Table \ref{table:attri_hyper_hypo_gl_fg}).
In most cases, when the ``glucose\_level'' belongs to hypo/normal/hyper, the values of ``finger\_stick'' are also in the same status.
For example, in terms of participant 570, the mean values of ``glucose\_level'' are 63.00/133.52/237.76 mg/dL respectively in hypo/normal/hyper, while the mean values of ``finger\_stick'' are 63.38/132.54/240.72, which is in accord with hypo/normal/hyper.
However, not all variation trends of ``finger\_stick' are consistent with the variation trends of ``glucose\_level''.
For example, in terms of participant 559, when the mean value, i.e., 59.90 mg/dL, of ``glucose\_level'' belongs to hypoglycemia, the mean value of ``finger\_stick'' is 108.36 mg/dL which is normal rather than hypoglycemia.
In this case, given that ``finger\_stick'' can be seen as the actual BG levels, the samples of ``finger\_stick'' can be used to calibrate the ``glucose\_level'' collected by CGM, since there might be some delays, estimations and mistakes of CGM, contributing to the inconsistencies with the actual BG levels.
Therefore, because of these inconsistencies between ``glucose\_level'' and ``finger\_stick'', it can prove why ``finger\_stick'' is helpful during the estimations of the future BG levels.

Furthermore, in Table \ref{table:attri_hyper_hypo_ml_bl}, we can find that there are 9 participants whose carbohydrate intake, ``meal'', is lower during hypoglycemia compared with the mean value of carbohydrate intake of all samples.
In other words, ``meal (hypo)'' is lower than ``meal (all)''.
This is reasonable since lower carbohydrate intake tends to contribute to lower BG levels.
However, other 3 participants tend to eat more food during hypoglycemia in order to increase their BG levels.
During hyperglycemia, there are 6 participants who choose to have more or similar to the mean carbohydrate intake of all samples, where the mean values of ``meal (hyper)'' are larger than or close to ``meal (all)''.
Other 6 participants try to eat less food during the hyperglycemia so as to reduce their BG levels. 
Hence, from these observations, we can find that only a relatively small part of participants control their BG levels via actively controlling the carbohydrate intake.
This is because the correct action during hypoglycemia/hyperglycemia is to have more/less carbohydrate intake, but actually, only a relatively small part of participants follow this rule. 

Comparably, in terms of ``bolus'', there are 8 participants whose mean values of ``bolus (hypo)'' are less than ``bolus (all)''.
Interestingly, during hyperglycemia, there are 7 participants whose mean values of ``bolus (hyper)'' are larger than or close to ``bolus (all)''.
This is because, having less insulin delivery during hypoglycemia can mildly reduce the BG levels, and delivering more bolus during hyperglycemia can badly decrease the BG levels.
Therefore, during hypoglycemia, the correct action is to have lower or no insulin delivery, while participants should have higher insulin delivery during hyperglycemia.
Hence, compared with carbohydrate intake, more participants follow the correct rules and well leverage insulin delivery to control their BG levels.

In terms of ``sleep'' in Table \ref{table:attri_hyper_hypo_sl_ex}, it is interesting to find that 9 participants have hypoglycemia when they have poor sleep quality, where ``sleep (hypo)'' is less than ``sleep (all)''.
Meanwhile, 10 participants have hyperglycemia when they have good sleep quality, where ``sleep (all)'' is larger than or close to ``sleep (hyper)''.
In terms of ``exercise'', more people tend to exercise when their BG levels are normal (exercise (normal) is not None).
This is also reasonable as people always feel uncomfortable when they have hypoglycemia or hyperglycemia.

In summary, we can find some interesting patterns from the statistics of these attributes in terms of hypoglycemia/normal BG levels/hyperglycemia.
Based on these observations, to some degree, we can understand why these attributes are helpful for the BGLP-RMTS problem.

\section{Visualization}
\label{sec:vis}

\begin{figure}[tb]
	\centering
	\includegraphics[width = 1.\columnwidth]{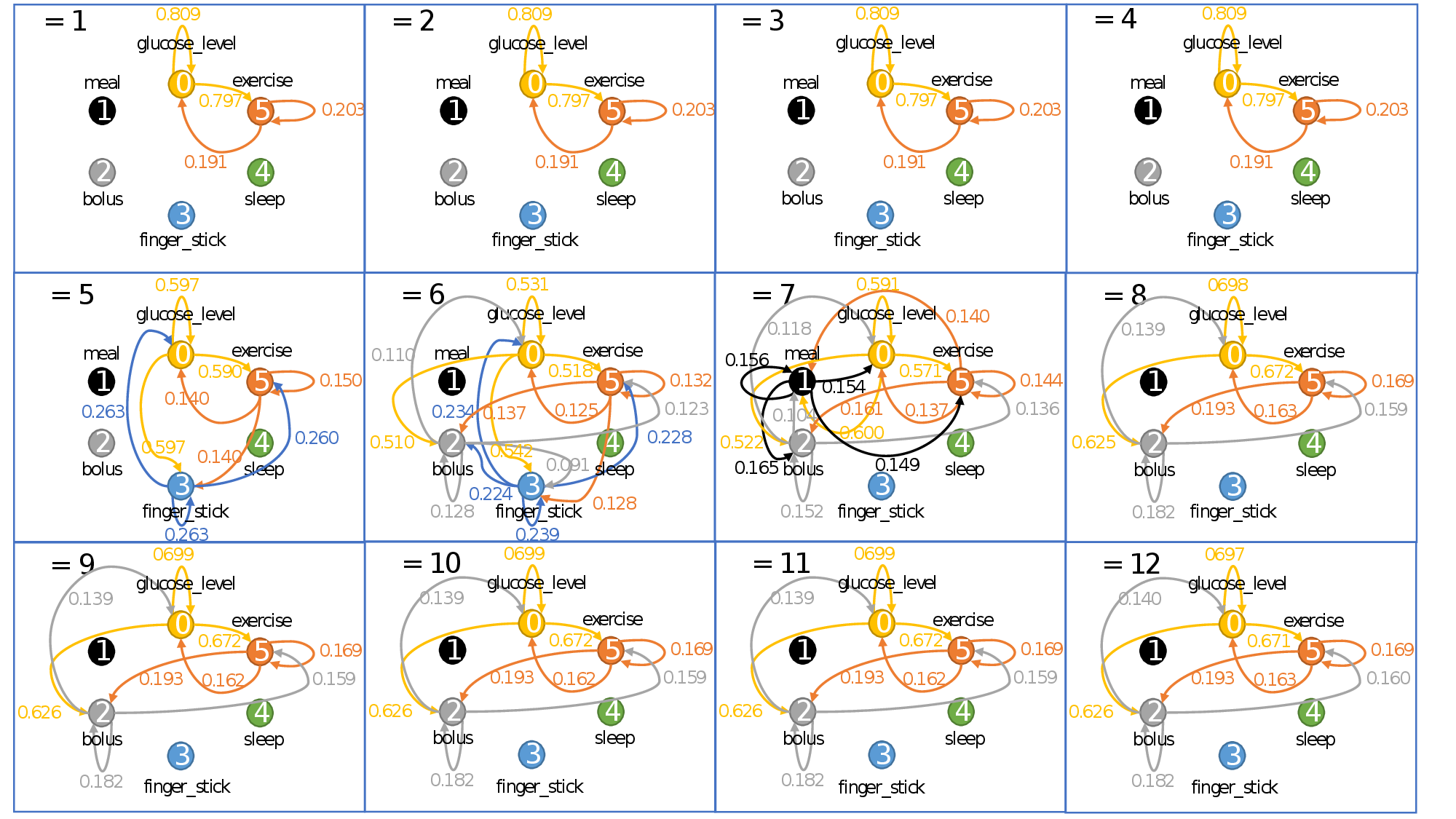}
	\caption{Example 1: visualization of graph attentions of participant 570.}

	\label{fig:vis_att1}
\end{figure}

\begin{figure}[tb]
	\centering
	\includegraphics[width = 1.\columnwidth]{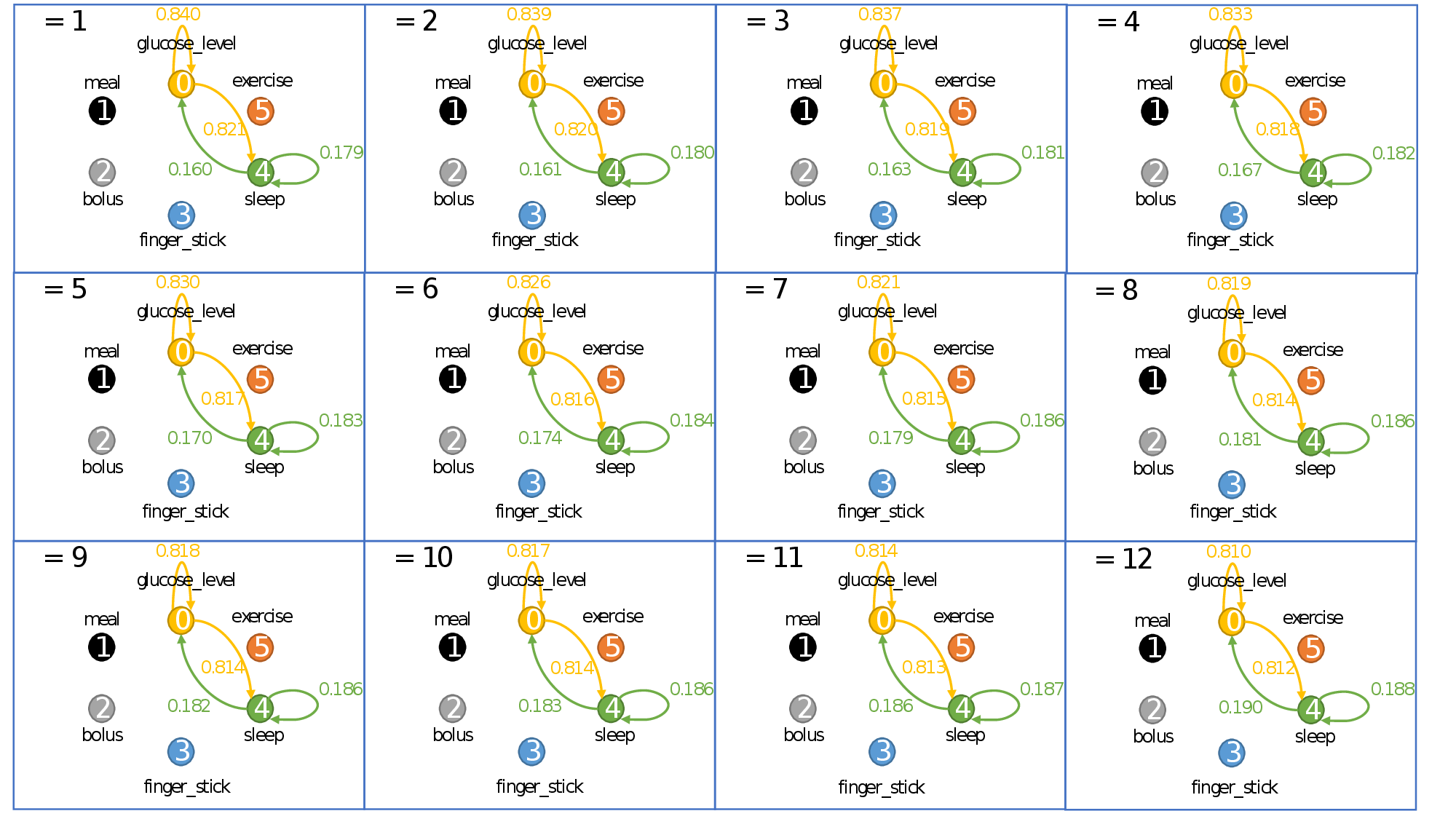}
	\caption{Example 2: visualization of graph attentions of participant 570.}

	\label{fig:vis_att2}
\end{figure}

We use two examples to visualize the GAM, as Figures \ref{fig:vis_att1}-\ref{fig:vis_att2}, where the examples are selected from the testing samples of participant 570 in OhioT1DM'18, given $W=6$ and $T=12$.
For each node, the summation of the weights of the edges coming into the node is 1, and the weights are attention weights calculated by Equation (\ref{eqn:gat_1}), and the receiving messages are aggregated by these attention weights according to Equation (\ref{eqn:gat_2}).

In Figure \ref{fig:vis_att1}, when $t\in\{1,...,4\}$, we can observe that only the node of ``glucose\_level'' and the node of ``exercise'' are activated.
Hence, only these two nodes send neural messages to each other, generating aggregated messages from these two nodes, respectively.
During the aggregation of neural messages, the neural messages from ``glucose\_level'' account for around 80\%, while the neural messages from ``exercise'' account for nearly 20\%.
According to the proportions of neural messages from different nodes during aggregation, we can get the ranking of attribute importance as ``glucose\_level'' $>$ ``exercise''.

When $t=5$, the node of ``finger\_stick'' is activated.
Then, there are 3 nodes sending neural messages to each other.
When aggregating neural messages, the neural messages from ``glucose\_level'' account for around 60\% which is decreased from nearly 80\%, while the neural messages from ``exercise'' account for nearly 14\% which is reduced from around 20\%.
The neural messages from the node of ``finger\_stick'' account for about 26\%.
The attribute importance changes when more nodes are activated.
In this case, the ranking of attribute importance is ``glucose\_level'' $>$ ``finger\_stick'' $>$ ``exercise''.
Similarly, when $t=6$, the node of ``bolus'' is activated, and the attention weights of all nodes change accordingly.
The ranking of attribute importance is ``glucose\_level'' $>$ ``finger\_stick'' $>$ ``exercise'' $>$ ``bolus''.
Then, the node of ``finger\_stick'' is deactivated, but the node of ``meal'' is activated, when $t=7$.
The ranking of attribute importance is ``glucose\_level'' $>$ ``meal'' $>$ ``exercise'' $>$ ``bolus''.
Finally, when $t\in\{8,...,12\}$, the ranking of attribute importance is stabilized as ``glucose\_level'' $>$ ``exercise'' $>$ ``bolus''.

In Figure \ref{fig:vis_att2}, when $t\in\{1,...,12\}$, there are only two active nodes, i.e., ``glucose\_level'' and ``sleep'', and the ranking of attribute importance is stabilized as ``glucose\_level'' $>$ ``sleep''.
However, the values of attention weights change with $t$.
In terms of ``glucose\_level'' (see the yellow lines), the attention weight decreases when $t$ increases.
On the contrary, in terms of ``sleep'' (see the green lines), the attention weight increases when $t$ increases.

In summary, only activated nodes generate, send, receive and aggregate neural messages.
When new nodes are activated, or active nodes are deactivated, the attention weights of all active nodes also change accordingly.
In some cases, even though the ranking of attribute importance is stable when increasing $t$, the attention weights might still change.
Based on these findings, our proposed model is efficient and flexible since not all nodes are active at a time, and the structure of the graph is dynamically changing.
Notably, the attention weights give a powerful explanation of our model.

\begin{figure}[tb]
	\centering
	\includegraphics[width = 1\columnwidth]{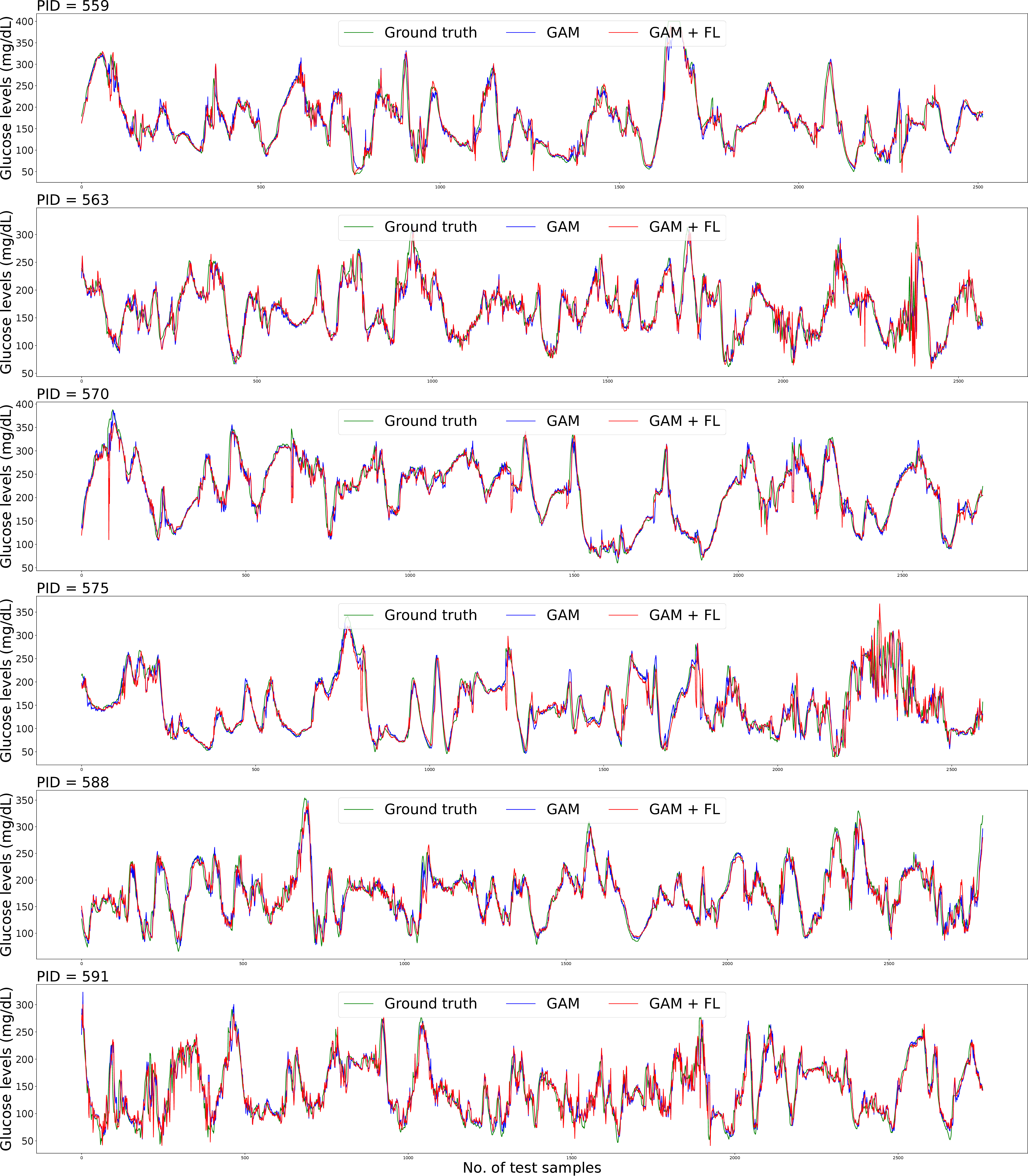}
	\caption{Example 1: visualization of predictions of GAM ($W=6$).}

	\label{fig:vis_pred_6_1}
\end{figure}

\begin{figure}[tb]
	\centering
	\includegraphics[width = 1\columnwidth]{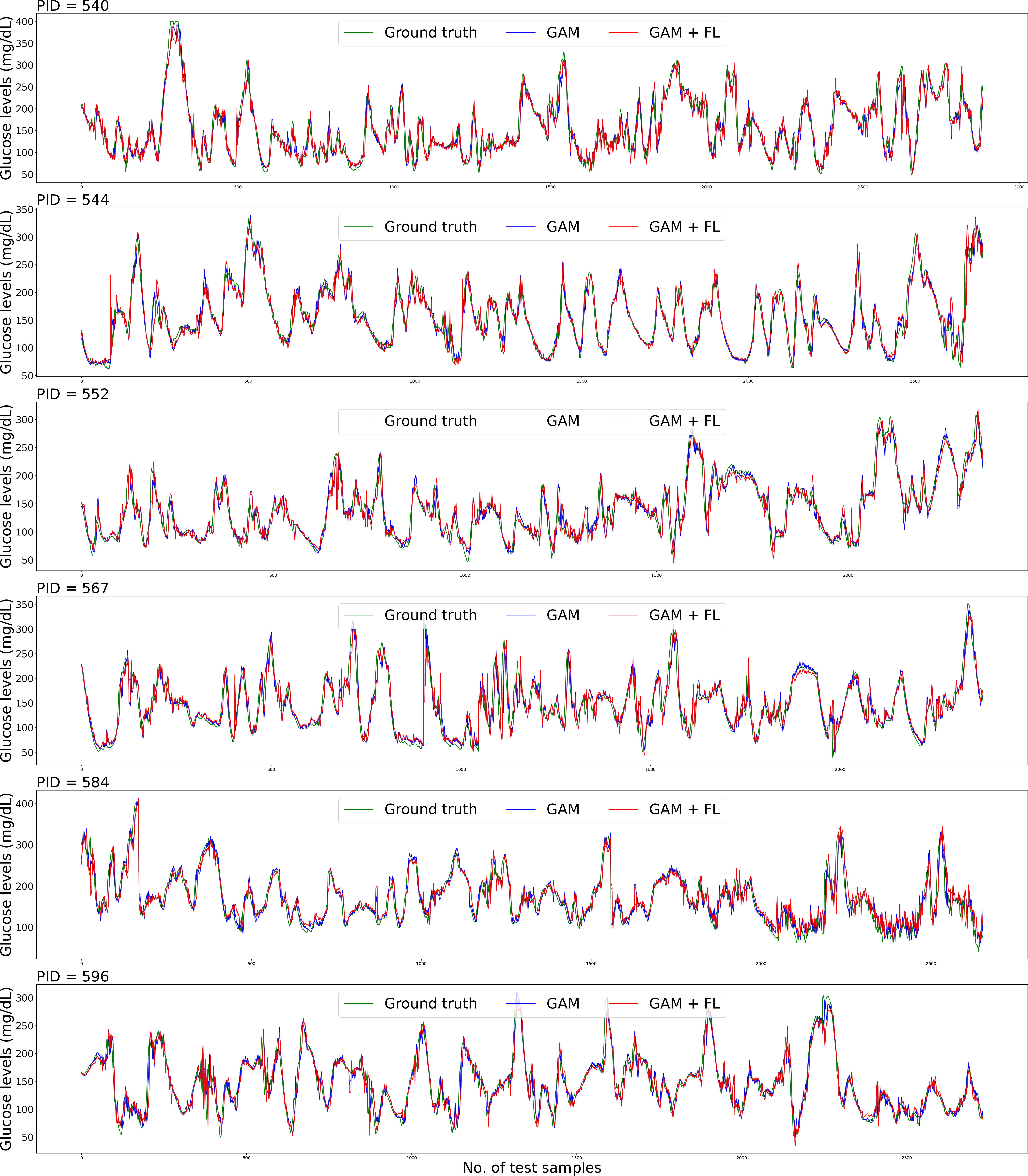}
	\caption{Example 2: visualization of predictions of GAM ($W=6$).}

	\label{fig:vis_pred_6_2}
\end{figure}

\begin{figure}[tb]
	\centering
	\includegraphics[width = 1\columnwidth]{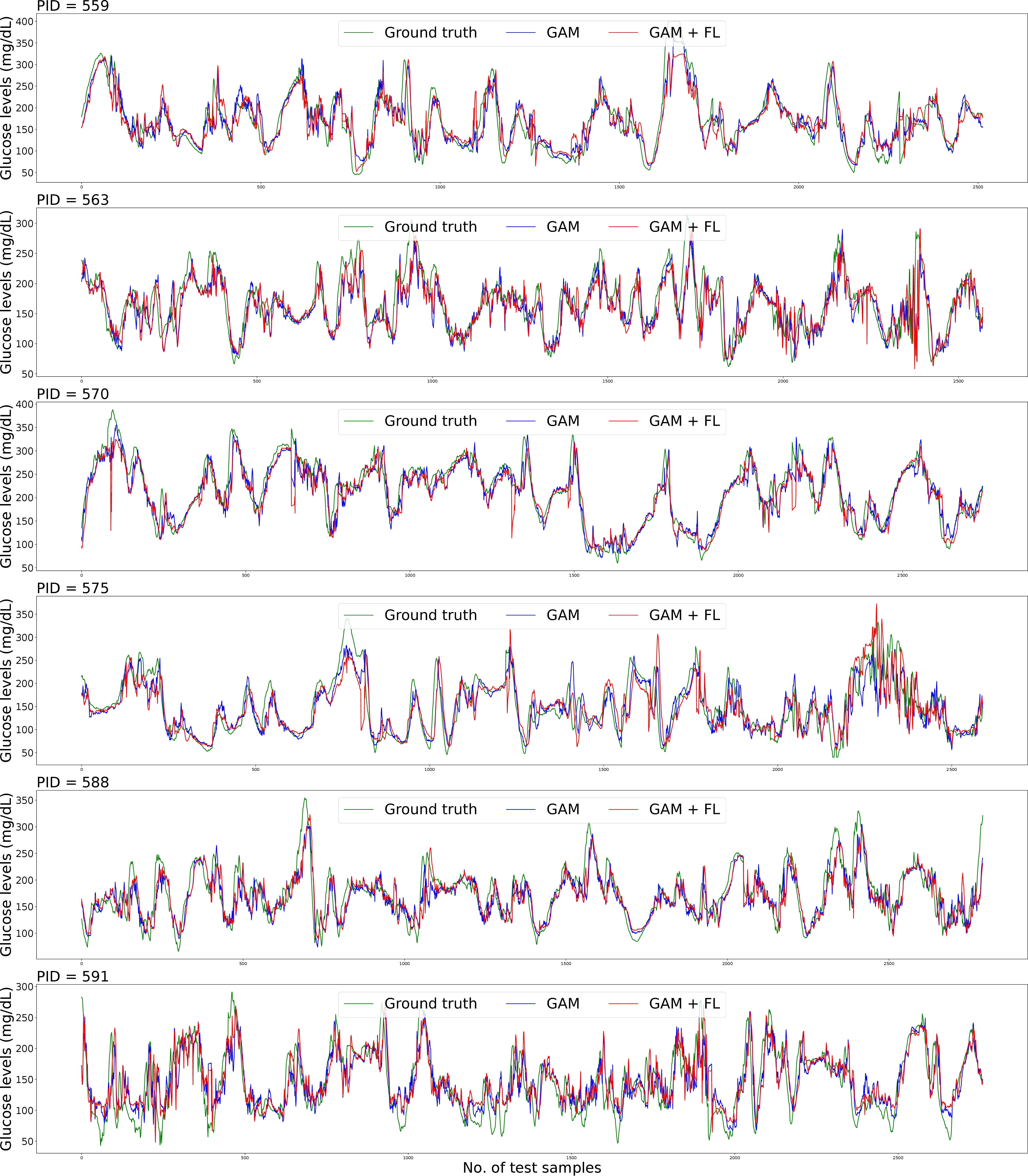}
	\caption{Example 1: visualization of predictions of GAM ($W=12$).}

	\label{fig:vis_pred_12_1}
\end{figure}

\begin{figure}[tb]
	\centering
	\includegraphics[width = 1\columnwidth]{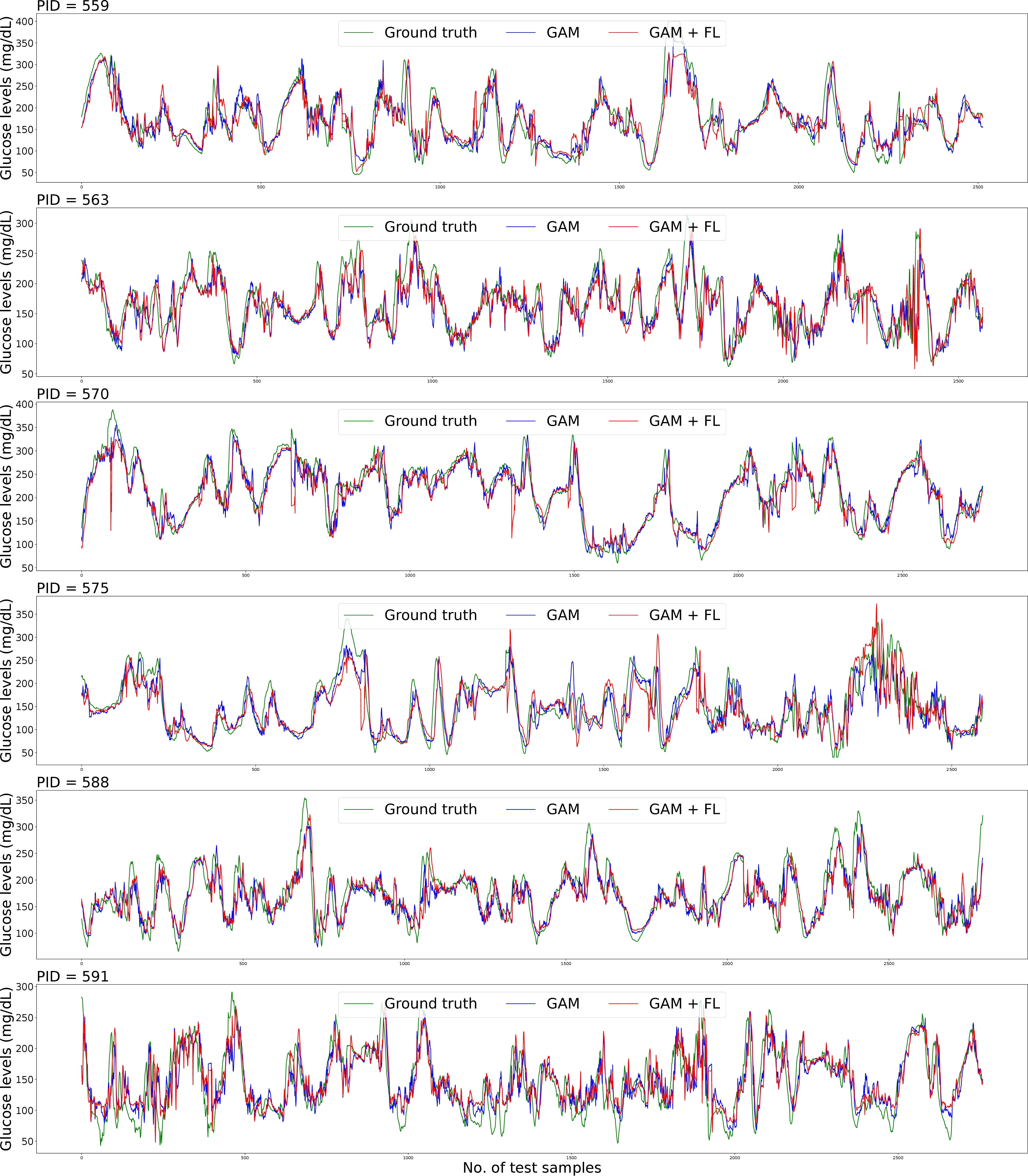}
	\caption{Example 2: visualization of predictions of GAM ($W=12$).}

	\label{fig:vis_pred_12_2}
\end{figure}

Then, we visualize the testing sample of all participants, as shown in Figures \ref{fig:vis_pred_6_1}-\ref{fig:vis_pred_12_2}.
The green curves are the ground truth, and the blue/red curves are the estimated BG levels by ``GAM''/``GAM+FL'' via sliding windows ($W=6$ or $W=12$).
When $W=6$ (see Figures \ref{fig:vis_pred_6_1}-\ref{fig:vis_pred_6_2}), there is no noticeable difference among ``GAM'', ``GAM+FL'' and ``Ground truth''.
When $W=12$ (see Figures \ref{fig:vis_pred_12_1}-\ref{fig:vis_pred_12_2}), there is also no visible difference between ``GAM'' and ``GAM+FL'', but there is a massive gap between ``GAM''/``GAM+FL'' and ``Ground Truth''.
Hence, even though there are distinct differences in Tables \ref{table:6_rmse_fl}-\ref{table:12_MAE_fl} when comparing ``GAM'' with ``GAM+FL'', the visual differences are minor.
The huge gap between ``GAM''/``GAM+FL'' and ``Ground Truth'' shows that the task gets much more challenging when $W=12$.

\section{Convergence and Training Time}
\label{sec:cong}
\begin{figure}[tb]
	\centering
	\includegraphics[width = 1.0\columnwidth]{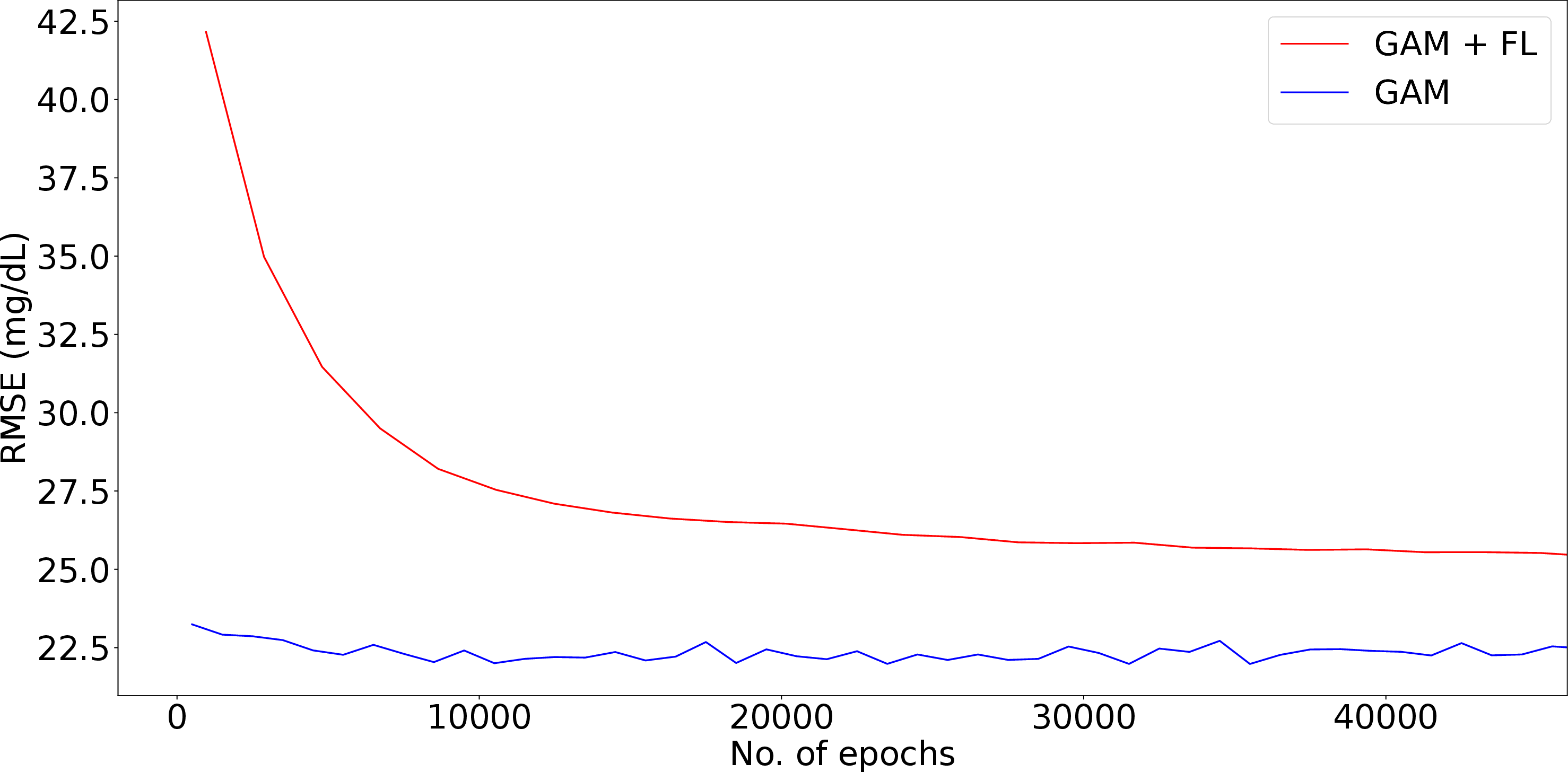}
	\caption{Training convergence of ``GAM'' and ``GAM + FL''.}
	\label{fig:cong_rmse}
\end{figure}

\begin{figure}[tb]
	\centering
	\includegraphics[width = 1.0\columnwidth]{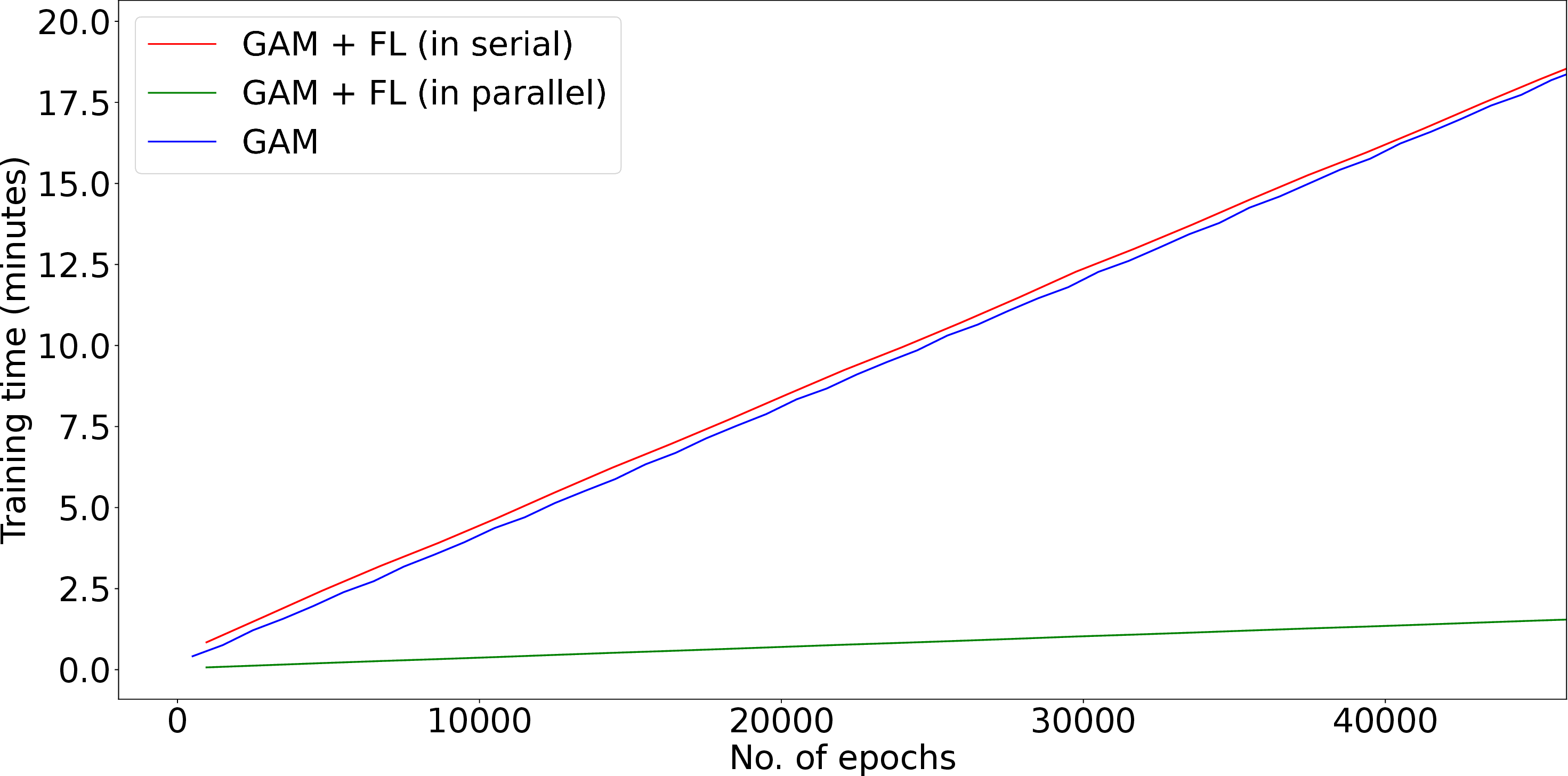}
	\caption{Training time of ``GAM'' and ``GAM + FL''.}
	\label{fig:cong_time}
\end{figure}

We compare the convergence and training time between ``GAM'' and ``GAM + FL'', where $W=6$, $H=256$, $M=1$, $L=1$ and $T=12$.
In Figure \ref{fig:cong_rmse}, in terms of the RMSE in the validation dataset, we can find that ``GAM'' converges immediately and maintains a better range compared with ``GAM + FL''.
When comparing the training time, ``GAM + FL (in serial)'' is slightly slower than ``GAM'' (see Figure \ref{fig:cong_time}). 
The clients of ``GAM + FL (in serial)'' run one by one, meaning that only when a client finishes training does another client start to train. 
``GAM + FL (in serial)'' spends more time collecting, averaging, and sending hyperparameters in the server.
In practice, clients run in parallel, and the fastest training time in theoretical is the green line in Figure \ref{fig:cong_time}, i.e., ``GAM + FL (in parallel)'', where all clients start to train at a specific time and finish training together.
Hence, even though the performance of ``GAM + FL'' is worse than ``GAM'' (see section \ref{sec:fl_res}), the training speed of ``GAM + FL'' might super fast.

\section{Conclusion and Future Work}
\label{sec:conclusion}

In this paper,
We have comprehensive data analysis, experiments and result analysis to show the selected 6 attributes are reliable and are positively affecting the prediction of the future BG levels.
We propose a novel graph attentive memory, called ``GAM''.
It is explainable, efficient and flexible, given that the attention weights of the graph can be utilized as attribute importance, and the attention weights and the graph structure can dynamically change according to the activated nodes.
We also have related experiments to show the stability and excellence of our GAM.
In the discussion, we compare ``GAM'' with other methods, finding the GAM is quite competitive.
We also introduce FL to ``GAM'', where FL enables leveraging population patterns without an invasion of the participants' privacy.
Relevant experiments show that the FL algorithm sacrifices some prediction accuracy.

The limitations of this work are: 1) only considering two datasets OhioT1DM'18 and OhioT1DM'20; 2) the GAM is not fully explainable, especially from the temporal perspective; 3) when compared with others, the results of other methods are directly introduced from the published papers, and it exists many unfair comparisons; 4) the two-step training is not elegant; 5) more advanced FL algorithms should be considered, as there are still obvious gaps between ``GAM'' and ``GAM+FL''.

Based on the limitations, in future work, we will consider larger datasets.
Then, we will also find more baselines from the latest top conferences, and run all the baselines by ourselves instead of introducing the results from the published papers.
In this case, all methods use the same data which is processed in the same manner, and the comparisons will be fairer because the performance differences are not caused by the differences in the data.

Meanwhile, in terms of the proposed model, we want to add new modules to make it explainable from a temporal view.
Furthermore, given that BGLP needs relatively long historical data (see section \ref{sec:hyper_res}), we may also take a more sophisticated memory mechanism into account in order to memorize long-term information.

On the other hand, in order to make our model more applicable, more sophisticated and advanced FL algorithms will be introduced or proposed, in order to make the training process asynchronous, personalized and more elegant, only containing one-step training rather than two-step training.
We will also try to deploy the FL in real mobile devices instead of only doing simulations.
In this case, the computation resources in mobile devices will be utilized, and asynchronous FL is indispensable.
This is because maybe sometimes users do not want their mobile devices to attend the training.
With the help of asynchronous FL, users can join the training flexibly.
Besides, there are no cold starting problems when using FL.
If a new user wants to join the FL training, the system can build a personal model immediately, and the model is internalized with the latest learnable parameters of the population model.
When the new user generates enough personal data after using CGM, the personalized model will be fine-tuned by the personal data.
If the FL is not only asynchronous but also personalized, there will be no fine-tuning process.
This is because, in personalized FL, there may be regularization items in the loss, making the personal model decoupled with the population model.
Then, the personalized model can keep the specialty, and the population model can remain general.

After achieving all the proposed points, our future model is not only explainable but also can be directly trained and leveraged in mobile devices.
The privacy of the participants is well protected.
The participants can use their mobile devices to train personalized and explainable models which concurrently leverage population patterns by FL.
Then, more participants can attend the training process flexibly and asynchronously.
Accordingly, when more participants join the training, the personalized model will be more accurate, attracting more potential participants.
The life quality of people with T1D will get better and better.


\bibliographystyle{plainnat}
\bibliography{example}

\end{document}